\documentclass{article}

\usepackage[final]{neurips_2025}

\usepackage[utf8]{inputenc} 
\usepackage[T1]{fontenc}    
\usepackage{hyperref}       
\usepackage{url}            
\usepackage{booktabs}       
\usepackage{amsfonts}       
\usepackage{nicefrac}       
\usepackage{microtype}      
\usepackage{xcolor}         

\usepackage[title]{appendix}
\usepackage{multibib}  
\newcites{appendix}{Appendix References}

\title{Large Language Models as Model Organisms for Human Associative Learning}

\author{%
  \textbf{Camila Kolling} \hspace{1.15cm} \textbf{Vy Ai Vo} \hspace{1.15cm} \textbf{Mariya Toneva}\\
  Max Planck Institute for Software Systems, Saarbrücken, Germany\\
  \texttt{\{ckolling, vyvo, mtoneva\}@mpi-sws.org}
}

\usepackage{booktabs}
\usepackage{multirow}
\usepackage{graphicx}
\usepackage{subcaption}
\usepackage{xcolor}
\usepackage{algorithm}
\usepackage{algpseudocode}
\usepackage{amsmath}
\usepackage{makecell}

 \usepackage{placeins}

\begin{document}

\maketitle



\begin{abstract}
Associative learning--forming links between co-occurring items--is fundamental to human cognition, reshaping internal representations in complex ways. 
Testing hypotheses on how representational changes occur in biological systems is challenging, but large language models (LLMs) offer a scalable alternative. Building on LLMs' in-context learning, we adapt a cognitive neuroscience associative learning paradigm and investigate how representations evolve across six models.
Our initial findings reveal a non-monotonic pattern consistent with the Non-Monotonic Plasticity Hypothesis, with moderately similar items differentiating after learning. Leveraging the controllability of LLMs, we further show that this differentiation is modulated by the overlap of associated items with the broader vocabulary--a factor we term vocabulary interference, capturing how new associations compete with prior knowledge.
We find that higher vocabulary interference amplifies differentiation, suggesting that representational change is influenced by both item similarity and global competition.
Our findings position LLMs not only as powerful tools for studying representational dynamics in human-like learning systems, but also as accessible and general computational models for generating new hypotheses about the principles underlying memory reorganization in the brain.



\end{abstract}

\section{Introduction}
\label{sec:intro}

\begingroup
\renewcommand\thefootnote{}\footnotetext{Code available at \href{https://github.com/bridge-ai-neuro/llm-associative-learning}{github.com/bridge-ai-neuro/llm-associative-learning}.}
\endgroup

\begin{figure}[!t]
    \centering
    \includegraphics[width=0.85\linewidth]{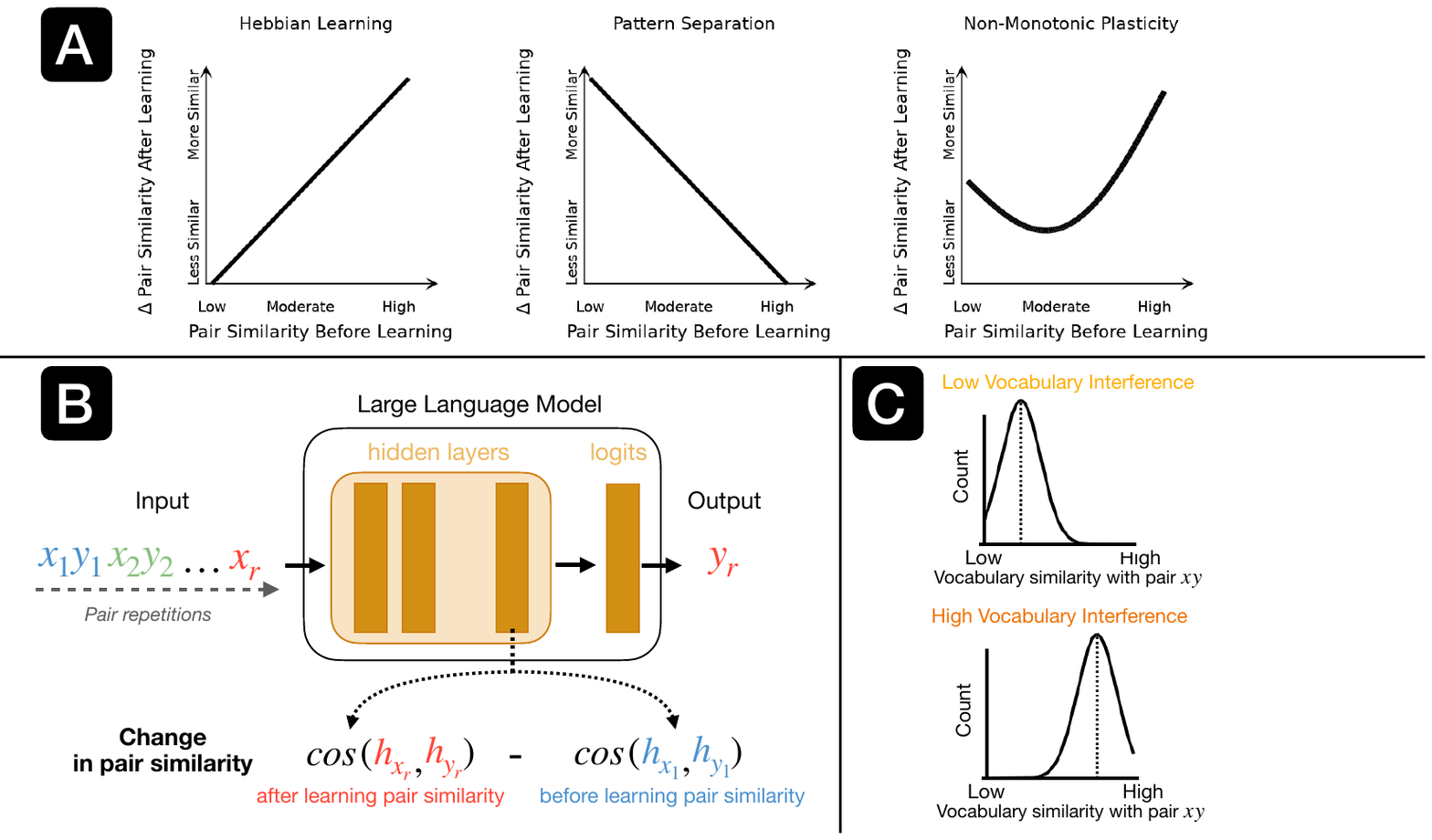}
    \caption{
    \textbf{(A)} Hypotheses about representational changes in humans. Competing theories propose different patterns of representational change as a function of pair similarity before learning: Hebbian learning predicts integration, pattern separation predicts differentiation, and the Non-Monotonic Plasticity Hypothesis (NMPH) predicts a U-shaped curve, with high differentiation at moderate similarity levels prior to learning.
    \textbf{(B)} Schematic of our adapted associative learning task for LLMs. Given repeated in-context presentations of a token pair $(x,y)$, the LLM learns to predict the associated token $y$. We measure representational change by computing the difference in cosine similarity between hidden representations of the pair before and after learning. This setup, inspired by a neuroscience paradigm~\cite{wammes2022increasing}, enables us to examine whether similar dynamics of representational restructuring emerge during in-context learning.
     \textbf{(C)} 
     Illustration of low and high vocabulary interference in the model's representational space. 
     In the low vocabulary interference case (top, yellow), the target token $y$ is dissimilar to most other tokens, resulting in less competition from alternative completions when paired with $x$. In the high interference case (bottom, orange), the pairing $xy$ is highly similar to many other possible token pairings, increasing competition and representational pressure to differentiate the learned association from potential distractors during learning.
    }
    \label{fig:approach_figure}
\end{figure}

Associative learning---the ability to form links between co-occurring items---is a fundamental mechanism that shapes how experiences are encoded, stored, and retrieved.
Its ubiquity across species and cognitive domains has made it a core component in theories of intelligence~\cite{wasserman1997s,drigas20208}.
As associations are learned, the brain’s internal representations of the associated items are altered---a reflection of the neural plasticity that strengthens some connections while weakening others~\cite{ritvo2019nonmonotonic,schapiro2017complementary,favila2020transforming}.
A central and ongoing question in cognitive neuroscience is how this self-supervised learning process reshapes representational structure, and why~\cite{wammes2022increasing,favila2016experience,chanales2021adaptive}. 
There are three main hypotheses for how associative learning alters representations in biological systems (see Figure~\ref{fig:approach_figure}A). 
The classical Hebbian learning rule, where repeatedly associating items strengthens connections between shared features, predicts more integrated representations across learned items~\citep{ritvo2019nonmonotonic}. 
However, alternative theories suggest the opposite. For example, the hippocampus often exhibits pattern separation, where rapidly learned memories reduce representational overlap to minimize interference and facilitate retrieval \citep{oreillyHippocampalConjunctiveEncoding1994, amerExtrahippocampalContributionsPattern2023, zotowBehavioralEvidencePattern2020, favila2016experience}. These opposing dynamics---integration versus differentiation---are both observed in human studies \citep{chanales2021adaptive, favila2016experience, schlichting2015learning, schapiro2012shaping}. To reconcile this, the Non-Monotonic Plasticity Hypothesis (NMPH) posits that representational change follows a U-shaped curve: highly similar or dissimilar items tend to integrate or remain stable, while moderately similar pairs differentiate \cite{ritvo2019nonmonotonic}.


Fully testing these hypotheses in biological systems is inherently difficult~\cite{wammes2022increasing, ritvo2024neural}.
A major challenge lies in precisely controlling the similarity between items before learning, a prerequisite for detecting non-monotonic representational change.
Moreover, the similarity level at which differentiation emerges can vary across tasks and stimuli, making it unclear in advance which mid-similarity range will reveal the effect.
Capturing this requires dense sampling across the similarity spectrum, further increasing experimental complexity.
Finally, human studies are constrained by cost, participant fatigue, and measurement noise, which limit the number of trials that can feasibly be conducted.
To address these challenges, we propose using large language models (LLMs) as model organisms for human associative learning.
LLMs exhibit complex cognitive behaviors~\cite{brown2020language, wei2022emergent,wei2022chain}, including in-context learning (ICL)~\cite{zhao2023context, jiang2024llms, dong2022survey}---rapidly forming associations without weight updates---making them promising for studying memory dynamics.
Unlike hand-crafted neural models designed to replicate specific representational dynamics~\cite{ritvo2024neural}, LLMs offer a scalable, natural testbed for uncovering emergent cognitive phenomena.

In this work, we investigate whether LLMs exhibit representational dynamics akin to human associative learning, and whether they can help disambiguate between competing hypotheses for representational change. We adapt a cognitive neuroscience associative learning paradigm to the ICL setting (Figure~\ref{fig:approach_figure}B), repeatedly presenting token pairs in-context to induce associations. By systematically controlling the similarity of token pairs before learning, we evaluate how representations evolve through learning across six open source, well-performing LLMs. Our initial findings support the NMPH: moderately similar pairs significantly differentiate after learning, mirroring human-like patterns of representational change.

We then leverage the controllability of LLMs to examine a factor that may further contribute to representational change, and is difficult to isolate in biological systems: the similarity between each paired item and the model's prior knowledge.
Because LLMs are pre-trained to encode co-occurrence statistics across the entire vocabulary--similar to how humans learn from experience--new associations introduced during ICL must compete with pre-existing patterns.
We refer to this competitive influence as \emph{vocabulary interference} (Figure~\ref{fig:approach_figure}C): the extent to which prior knowledge shapes the learning of new associations.
In such cases, learning the correct pairing may require greater changes in the model’s representations to distinguish it from competing associations.
This phenomenon has long been studied in neuroscience and psychology~\cite{raaijmakers1981search, caplan2014associations, stark2018s}, but empirical measurement in the brain is limited by the inability to access all competing representations.
By contrast, LLMs provide a tractable framework for quantifying this effect, as the entire distribution of token relationships is explicitly known.
We find that, while pair similarity remains a key determinant of representational change, vocabulary interference modulates this effect--greater interference leads to stronger differentiation.
These results position LLMs as valuable tools for probing associative learning principles, offering new insights into how both local and global associative structures influence representational change.

\section{Related work}
\label{sec:related_work}



\subsection{Representational change in human memory and neural models}


\textbf{Integration vs. differentiation in the brain.} 
Associative memory-related representational changes are primarily studied in the hippocampus, a region of the brain thought to be most influential in memory-driven behavior~\cite{squire1992memory,o1979precis}. Both integration and differentiation of memory representations have been shown to support distinct behavioral functions: differentiation reduces interference and enhances specific recall, while integration promotes generalization and inference across related experiences~\cite{wammes2022increasing, brunecIntegrationDifferentiationHippocampal2020}. These functional roles are thought to map onto distinct hippocampal subregions, e.g., integration with CA1 and differentiation with the dentate gyrus (DG), which shows sparse activity linked to orthogonalized representations~\cite{wammes2022increasing,yassa2011pattern}. How LLMs align with this integration–differentiation spectrum remains an open question.

\textbf{Non-monotonic plasticity in the brain.} The NMPH~\cite{ritvo2019nonmonotonic}
proposes that representational change depends non-linearly on pair similarity before learning: moderate similarity leads to differentiation, whereas low or high similarity leads to stability or integration.
Recently, Wammes et al.~\cite{wammes2022increasing} provided empirical results supporting this effect by parametrically manipulating the visual similarity of object pairs using CNN-derived~\cite{he2016deep,simonyan2014very} representations. 
Participants arranged images by perceived similarity, and the resulting pairwise distances correlated with model-based similarity estimates.
During fMRI, repeated exposure to these pairs revealed significant differentiation for mid-similarity pairs in the DG, but not for other parts of the hippocampus.

\textbf{Computational accounts}. To account for this variety of findings, \cite{ritvo2024neural}~proposed an unsupervised recurrent network model in which partial activation of competing memories during retrieval induces representational differentiation, a dynamic linked to retrieval-induced forgetting and inhibitory oscillations~\cite{norman2006inhibitory}.
While such models are an important step towards a computational account, they rely on hand-crafted inputs and necessitate simplified settings, limiting scalability and behavioral richness.
Our work complements prior computational efforts by investigating whether non-monotonic differentiation, previously observed in biological memory systems, emerges naturally in large-scale, general-purpose LLMs trained on real-world data--without an explicit separated memory system.

\subsection{Associative learning and in-context dynamics in LLMs}
LLMs are increasingly studied as systems capable of associative learning, rapidly forming token-level associations directly within the input context~\cite{brown2020language}.
Recent work shows that LLMs can form stable in-context associations that shape future predictions~\cite{jiang2024llms, zhao2023context}, exhibiting behaviors consistent with retrieval, interference, and generalization~\cite{akyurek2022learning, von2023transformers, geva2020transformer}.
These findings suggest that transformer-based architectures support implicit memory mechanisms across attention and MLP layers, despite the absence of explicit memory modules.
Several studies have also analyzed ICL as a form of fast memory encoding or Bayesian inference~\cite{burns2024associative, geva2020transformer}, and have shown that attention layers can support long-range retrieval, stability, and structured generalization~\cite{xie2021explanation,press2021train,garg2022can,olsson2022context,brown2020language,von2023transformers,jiang2024llms}.
ICL has been interpreted through the lens of both episodic memory, as models retrieve and reuse information based on context, and working memory, given that representations are updated dynamically across tokens without any parameter changes~\cite{zhao2023context,jiang2024llms,park2024competition, li2022large, burns2024associative,li2024linking,fountas2024human,dai2022can}.
We build on this line of research by shifting focus from behavioral outcomes to the internal representational dynamics underlying learning through repeated associative exposure.

\section{Methods}
\label{sec:method}


\subsection{Associative learning paradigm}
\label{subsec:learning_task}
Our associative learning paradigm is inspired by the experimental design of~\cite{wammes2022increasing}, who investigated how repeated exposure to stimulus pairs with different visual similarity leads to non-monotonic changes in human hippocampal representations.
We adapt this paradigm to LLMs using ICL~\cite{brown2020language, burns2024associative}, replacing visual stimuli with token pairs and modeling learning through repeated token co-occurrence.
This setup allows us to examine whether similar non-monotonic representational shifts occur in LLMs, and to what extent LLM behavior parallels hippocampal learning dynamics. 
We focus on ICL rather than fine-tuning, as LLMs are known to exhibit emergent associative abilities~\cite{wei2022emergent,brown2020language,dong2022survey}, making ICL a natural fit for studying association tasks.
It also provides a controlled and biologically plausible analogy to how humans acquire associations~\cite{zhao2023context,jiang2024llms}, while
enabling consistent comparisons across models of different sizes and architectures without introducing task-specific fine-tuning.

Formally, we present the token pair $(x,y)$ a total of $r-1$ times, followed by one final presentation of $x$ alone as a cue for predicting its paired token $y$.
Given the input sequence
\begin{equation}    
\mathbf{s} = [x_1,y_1,x_2,y_2,\dots,x_{r-1},y_{r-1},x_r],
\end{equation}
the model's goal is to generate a prediction of the associated paired token, $y$.
By default, we restrict the number of repetitions such that the total sequence length remains within each model’s $m$ maximum context length ($L^m_{\text{max}}$) or the limits imposed by available GPU memory ($L_\text{mem}\approx40k$ tokens), i.e., $L^m = \min(L^m_{\text{max}},\ L_{\text{mem}})$.
In our setup, the sequence length is $L_{s} \approx (2*r)-1$.


We predict that the LLM's representations of these tokens will change through the course of ICL, a phenomenon observed in prior studies analyzing ICL tasks~\cite{park2025iclr,yousefi2023decoding,dong2022survey}.
This prediction also aligns with findings from neuroscience, where repeated co-occurrence of stimuli is known to drive representational change in the hippocampus~\cite{ritvo2019nonmonotonic, schapiro2016statistical}.

\paragraph{(Pair) Representational change.} For a given model $m \in \mathcal{M}$, where $\mathcal{M}$ is the set of LLM models under study, we extract the hidden representations of a token $x$ at the last layer of the model, $\mathbf{h}^m_x$. We chose to examine the last hidden layer to more effectively control for representations that directly affect model behavior on the ICL task~\footnote{Preliminary results for the other layers are shown in Appendix~\ref{appendix:sec_other_layers}.}.
This choice also aligns with a sensory-information processing hierarchy in which the hippocampus sits at the top of the memory stream~\cite{o2012computational,lavenex2000hippocampal}.
(Pair) Representational change across ICL is then defined as the difference in cosine similarity between representations at repetition $r$ and the first occurrence of the pair:
\begin{equation}
\Delta S^m_r = cos(\mathbf{h}^m_{x_r}, \mathbf{h}^m_{y_r}) - cos(\mathbf{h}^m_{x_1}, \mathbf{h}^m_{y_1}),
\end{equation}
where the hidden representation is conditioned on the whole sequence up until that point, e.g., $\mathbf{h}^m_{x_r}=\mathbf{h}^m(x_r \mid x_1, y_1, \dots, x_{r-1}, y_{r-1})$. Note that our ICL paradigm means that the first occurrence of $y$ is always conditioned on $x$, $\mathbf{h}^m_{y_1}=\mathbf{h}^m(y_1 \mid x_1)$. 
This design mirrors human associative learning paradigms, where pairs are presented sequentially~\cite{wammes2022increasing}.
Throughout the work, we refer to the hidden states from the first occurrence $(\mathbf{h}_{x_1}, \mathbf{h}_{y_1})$ as the representations obtained before learning occurs.

\paragraph{Token similarity groups.} 
To examine whether LLMs exhibit representational dynamics consistent with those observed in humans, we sample token pairs across the similarity continuum.
More specifically, we sampled evenly along the cosine similarity axis, defining $17$ groups $g$ that fall within the interval 
$[0.1, 0.95)$. Each group is defined by a window $[\theta_{min}, \theta_{max})$ that spans $0.05$ cosine similarity.
The token pairs within each group are chosen such that their representational similarity before learning $cos(\mathbf{h}^m_{x_1}, \mathbf{h}^m_{y_1})$ lies within the group interval $[\theta_{\min}^g, \theta_{\max}^g)$. 

Details on our procedure to find these token pairs are given in the section below.
We find $12$ token pairs in each group to form a set for each model, $\mathcal{P}^m$. The token pair sets for each model are constructed independently due to differences in their vocabulary size and tokenization.


\subsection{Optimized search for pairs of tokens}
\label{subsec:pairmates_search}
To systematically find tokens whose pair similarity before learning falls within a given interval, we employ an efficient way for searching the large vocabulary space (between $10\text{k}^2$-$72\text{k}^2$ tokens, depending on the model).
Inspired by recent work on prompt and input optimization~\cite{zou2023universal, shin-etal-2020-autoprompt}, we follow a two-step approximation strategy to identify suitable pairs.
The Greedy Coordinate Gradient (GCG) algorithm~\cite{zou2023universal}, originally developed for optimizing sequences in adversarial settings (e.g., minimizing next-token likelihood), provides a framework for iteratively refining a sequence by making targeted, gradient-informed edits to individual tokens.
We repurpose the GCG method to minimize a loss defined over the cosine similarity of internal representations.

We start with a duplicate token pair $(x,x)$ with the goal of finding a pair $(x,y)$ that falls within the target cosine similarity range $[\theta_{\min}, \theta_{\max})$. 
We fix the first $x$ token, and iteratively replace the second token of this sequence by using gradient signals (without updating the model) to identify vocabulary items that would bring the pair’s cosine similarity closer to the target range.
We select replacements from the top-$k$ candidates that reduce the loss the most, repeating the process until the similarity falls within the desired range or a maximum number of steps is hit.
This approach efficiently guides token selection in a controlled, representation-aware way, enabling the construction of token pairs with precise similarity properties.
More information on this algorithm can be found in Appendix~\ref{appendix:methods}.

\subsection{Estimating vocabulary interference}
\label{subsec:vocab_methods}
To estimate how a given pair $(x,y)$ relates to the broader LLM vocabulary space, we fix $x$ and sample each alternative token $t$ from a representative subset of the vocabulary, $\tilde{\mathcal{V}}^{m} \subset \mathcal{V}^m$.
We then compute the similarity between the representation of the correctly associated token $y$ and each alternative token $t \in \tilde{\mathcal{V}}^{m}$, conditioned on $x$'s presentation in context.
This provides an estimate of how much $y$, when associated with $x$, competes with other pair completions in the vocabulary space, capturing the degree of the pair's \emph{vocabulary interference} in the model’s representational space.
Due to the computational cost of exhaustively computing all possible pairwise combinations, we randomly sample $1,000$ tokens from $\mathcal{V}^m$ to form the representative subset $\tilde{\mathcal{V}}^{m}$, resulting in $1$ million pairwise combinations.

Concretely, for each pair $(x,t)$ we extract its pair representation before learning, yielding the set 
$\mathcal{H}^{m}_{t} = \{ \mathbf{h}^{m}_{t_1} \mid \forall t \in \tilde{\mathcal{V}}^{m} \}$.
We then compare the representation of the pair $(x,y)$ before learning to each alternative pair, 
\begin{equation}
    \mathcal{S}_{y}^{\tilde{\mathcal{V}}^m} =  \{\cos(\mathbf{h}^{m}_{y_1},\mathbf{h}^{m}_{t_1}) \; \forall \mathbf{h}^{m}_{t_1} \in \mathcal{H}^m_t\}.
\end{equation}
We can interpret $\mathcal{S}_{y}^{\tilde{\mathcal{V}}^m}$ as a distribution showing how much interference the pair $(x,y)$ receives from all competing associations $(x,t)$ with $t \in \tilde{\mathcal{V}}^m$. We define the vocabulary interference score for each $(x,y)$ as the median of the set $\mathcal{S}_{y}^{\tilde{\mathcal{V}}^m}$.

All of the above has been described for a single token pair $(x,y)$. The analysis shown in Figure~\ref{fig:reprchange_pairsim_vs_vocabsim_initialpairmates} depicts results for token pairs drawn from the original stimulus set described above, $\mathcal{P}^m$, optimized solely for token pair similarity before learning.
We then extend the original set of $(x,y)$ pairs, from $\mathcal{P}^m$, to uniformly sample from the joint distribution of before-learning pair similarity and vocabulary interference (Figure~\ref{fig:reprchange_pairsim_vs_vocabsim_controlledsampling}). That is, we use the $\sim1$ million token pairs from our sub-sampled vocabulary $\tilde{\mathcal{V}}^m$ to yield a larger set $\mathcal{Q}^m$. We aimed to find at least $10$ pairs per similarity group $g$ and vocabulary interference group (details in Appendix~\ref{appendix:methods}).


\subsection{Experimental setup}
\label{subsec:experimental_setup}

We analyze six recent open-source base LLMs: Llama2-7b, Llama3.1-8b, Llama3.2-1b, Llama3.2-3b, Gemma2-9b, and Mistral-7b~\cite{touvron2023llama,grattafiori2024llama,team2024gemma,jiang2023mistral7b}.
These models were selected for their recency, open availability, and relatively small size within their respective families, providing a balance between computational efficiency and architectural representativeness.
All experiments were performed on internal compute clusters, using two NVIDIA H100 PCIe GPUs with $\approx 80$GB GPU memory per device. 
The computation of the experiments took a total of $\approx 15$ days.

\section{Results}
\label{sec:results}

\subsection{LLMs exhibit structured, multi-phase learning dynamics}
\label{subsec:accuracy_pairsim}

\begin{figure*}[!t]
    \centering
    \begin{subfigure}{0.48\textwidth}
        \centering
        \includegraphics[width=\linewidth]{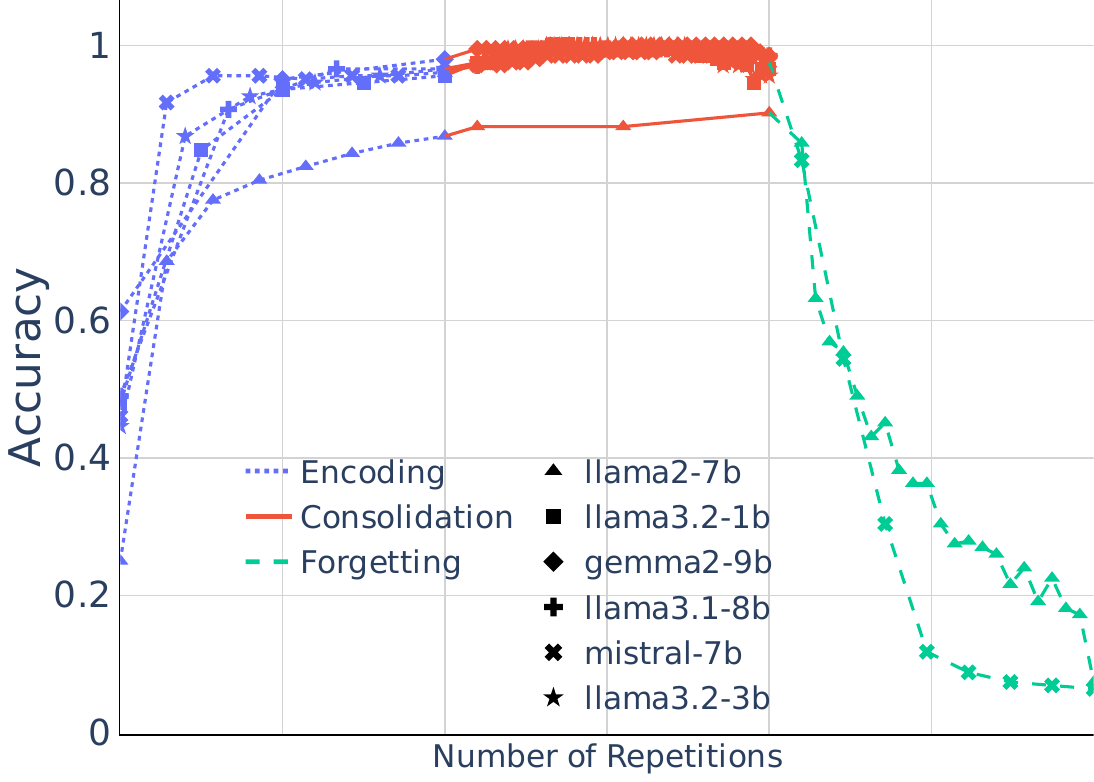}
        \caption{Accuracy across phases of learning.}
        \label{fig:acc}
    \end{subfigure}
    \hfill
    \begin{subfigure}{0.48\textwidth}
        \centering
        \includegraphics[width=\linewidth]{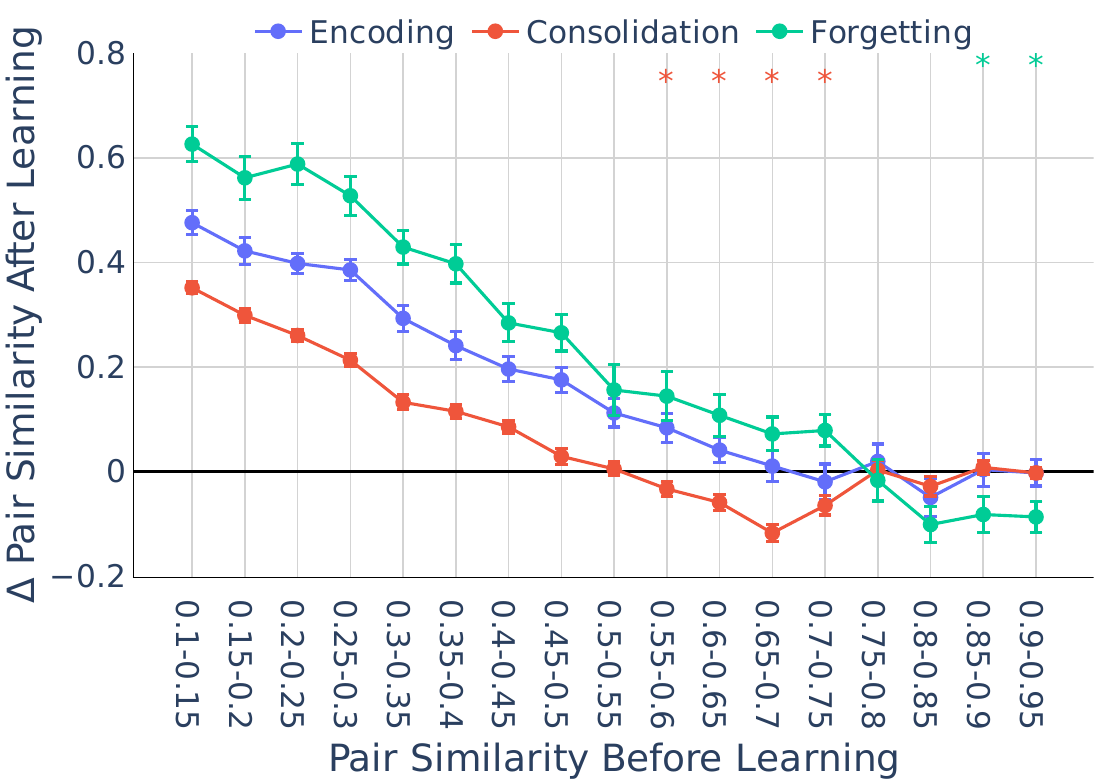}
        \caption{Representational change due to learning.}
        \label{fig:repr}
    \end{subfigure}
    \caption{
        Accuracy and representational changes during learning.
        (a) Models generally show three phases of learning: encoding, where accuracy steeply increases; consolidation, where accuracy stabilizes;
        and forgetting, where accuracy declines.
        To compare across models with different phase lengths, the x-axis is normalized: repetitions within each phase are linearly scaled to fixed intervals ($0-1$ for encoding, $1-2$ for consolidation, $2-3$ for forgetting), allowing phase-aligned trends to be visualized despite variability in learning dynamics.
        (b) The U-shaped differentiation pattern, characteristic of the Non-Monotonic Plasticity Hypothesis, is observed only during consolidation (red).
        Asterisks ($\ast$) indicate groups that remain significant after Benjamini--Yekutieli correction for multiple comparisons across similarity groups and phases ($p<0.05$).
    }
    \label{fig:analysis}
\end{figure*}

As expected, LLMs are able to complete the in-context associative learning task with high accuracy (between $90-100$\%), though the number of repetitions required to reach peak accuracy varies across models.
Figure~\ref{fig:acc} shows how overall prediction accuracy evolves as a function of the number of repetitions.
We identified three distinct phases of learning--Encoding, Consolidation, and Forgetting--and we observed that their duration varied across models.
To enable direct comparison across models, we normalized the number of repetitions in each phase by aligning phase boundaries: repetitions within each phase were linearly rescaled to fixed intervals ($0-1$ for Encoding, $1-2$ for Consolidation, and $2-3$ for Forgetting). 
This normalization preserves each model’s internal dynamics while making phase-aligned trends directly comparable across models.
The accuracy curves per model are provided in Appendix~\ref{appendix:subsec_acc_dyn_per_model}.

\begin{itemize}
    \item \textit{Encoding phase (blue):} This phase corresponds to the initial stage of learning, defined by a steep increase in accuracy as the model is repeatedly exposed to the token pair. We define this phase as the period during which accuracy continues to rise by more than $3$\% between consecutive repetitions, until the model reaches at least $97$\% of its peak performance.

    \item \textit{Consolidation phase (red):} This phase reflects a stable performance regime, where the model has largely acquired the association and maintains high accuracy over repetitions. Accuracy remains within $\pm 3$\% of the peak, indicating that learning has plateaued and performance is stabilized.

    \item \textit{Forgetting phase (green):} Surprisingly, in some models, accuracy begins to decline even though the number of repetitions remains within the model's maximum context window ($L_s < L^{m}_\text{max}$). We define the forgetting phase as the point where accuracy drops by more than $3$\% relative to the average of the two prior repetitions, marking the emergence of performance degradation.
\end{itemize}

While all models exhibited the Encoding and Consolidation phases, only two models (Llama2-7b and Mistral-7b) showed a forgetting phase.
For Llama2-7b, forgetting begins relatively early ($r=40$), whereas for Mistral-7b it emerges much later ($r=3,000$).
We speculate that the delayed forgetting in Mistral-7b may be related to its use of a sliding window attention (SWA).
For Llama2-7b, we present preliminary analyses in Appendix~\ref{appendix:subsec_forgetting_phase}, but the underlying cause of early forgetting is not yet well understood.
More broadly, it remains unclear how to predict if, and when, forgetting will occur. We leave this question to future work.
Overall, these results demonstrate that LLMs can effectively acquire associations and maintain them for a sustained period before eventual degradation.

\subsection{Moderately similar pairs significantly differentiate during consolidation}
\label{subsec:reprchange_pairsim}

We next investigate how the representations of successfully associated token pairs evolve during learning, specifically focusing on identifying when integration or differentiation occurs.
Figure~\ref{fig:repr} shows how the representational similarity between token pairs changes as a function of their similarity before learning across different phases of learning. We aggregate representational change values $\Delta S$ by collapsing across models and token pairs within each similarity group $g$ and learning phase, and report the mean and standard error of the resulting values.
To test for differentiation, we performed one-sided paired $t$-tests for each similarity group and learning phase, testing whether pair similarity after learning was significantly lower than pairsimilarity before learning.
To account for multiple comparisons across the $17$ similarity groups and $3$ learning phases, we applied the Benjamini–Yekutieli (BY) procedure to control the false discovery rate under dependency among tests.
Groups that remain significant after BY correction ($p<0.05$) are marked with asterisks.

During the \emph{Encoding} phase, no significant differentiation is observed across groups that were highly similar before learning.
Instead, models show a consistent increase in pairwise similarity for low- to mid-similarity pairs (between $0.1$–$0.6$), reflecting early-stage representational integration: repeated co-occurrence leads these tokens to move closer together in representation space, supporting initial association formation. In contrast, mid- to high-similarity pairs ($0.65$–$0.95$) exhibit little to no representational change at this stage.

During the \emph{Consolidation} phase, a striking effect emerges for pairs that were moderately similar before learning ($0.55-0.75$): these groups exhibit a significant decrease in pairwise similarity during this phase of learning.
This produces a clear U-shaped pattern in representational change--consistent with predictions from the NMPH~\cite{ritvo2019nonmonotonic, wammes2022increasing}.
Notably, this effect coincides with the stabilization of model performance, suggesting that LLMs undergo structured reorganization of internal representations to maintain high task accuracy.
Otherwise, we find that lower similarity pairs still exhibit integration, although to a lesser extent than during Encoding.
Higher similarity groups remain largely unchanged, suggesting that their representational similarity is relatively stable across the first two learning phases.

During the \emph{Forgetting} phase, the previously observed non-monotonic pattern disappears, and mid-similarity pairs no longer exhibit significant differentiation.
Surprisingly, this is the only phase in which groups that were highly similar before learning show a notable change in their representational structure, displaying clear signs of differentiation relative to their before-learning similarity.
Low-similarity pairs, by contrast, undergo even stronger integration than during the Encoding phase.
This results in a mild, approximately linear trend in representational change as a function of similarity before learning---resembling the general trend of Encoding, but with greater integration at low similarity and stronger differentiation at high similarity.
This trend indicates a loss of structured representational updates, aligning with the observed decline in accuracy. Further results of the evolution of these changes are presented in Appendix~\ref{appendix:subsec_repr_dyn_per_model}.

Taken together, our findings show that LLMs exhibit structured representational dynamics consistent with the NMPH.
Interestingly, this non-monotonic pattern is present only during the Consolidation phase, when behavioral performance is stably high, but absent during the Encoding and Forgetting---phases marked by behavioral instability and less structured representational change.
Importantly, unlike prior computational models explicitly designed to produce U-shaped dynamics~\cite{ritvo2024neural}, the LLMs that exhibit this non-monotonic effect are general-purpose, pretrained models that were not architecturally constrained or fine-tuned to exhibit such behavior.



\subsection{Pair similarity drives representation change, modulated by vocabulary interference}
\label{subsec:pairsim_vs_vocabsim}


During the \emph{Consolidation} phase---when models exhibited stable maintenance of learned associations---we observed a non-monotonic pattern of representational change as a function of pairwise similarity before learning: low-similarity pairs (up to $\approx 0.5$) integrated, mid-similarity pairs ($0.55$–$0.75$) differentiated, and high-similarity pairs ($> 0.75$) showed little to no representational change, aligning best with the NMPH.
Building on this analysis, we next examine an additional factor that may contribute to this pattern and is difficult to isolate in biological systems: the similarity between each paired item and the model’s prior knowledge.
Because LLMs are pre-trained to encode co-occurrence statistics across the entire vocabulary, as humans are thought to do through learning, new associations introduced during ICL must compete with pre-existing patterns. We expect that this competition---what we refer to as vocabulary interference---influences representational change: greater interference (i.e., higher similarity to other items in the vocabulary space) should impose stronger pressure for differentiation to support successful learning.

We thus extend the representational similarity change analysis from Section~\ref{subsec:reprchange_pairsim} by systematically examining these changes across different levels of vocabulary interference. Specifically, using our original token pairs $(x,y)$ in $\mathcal{P}^m$, we estimate their vocabulary interference with respect to alternative tokens in the set $\tilde{\mathcal{V}}^m$ (see Section~\ref{subsec:vocab_methods}).
To facilitate comparison across conditions, we categorize pairs into three equally sized groups based on their vocabulary interference scores: \emph{Low}, \emph{Mid}, and \emph{High} (see Appendix~\ref{appendix:subsec_vocab_sim} for details).

\begin{figure*}[!t]
    \centering
    \begin{subfigure}{0.48\textwidth}
        \centering
        \includegraphics[width=\linewidth]{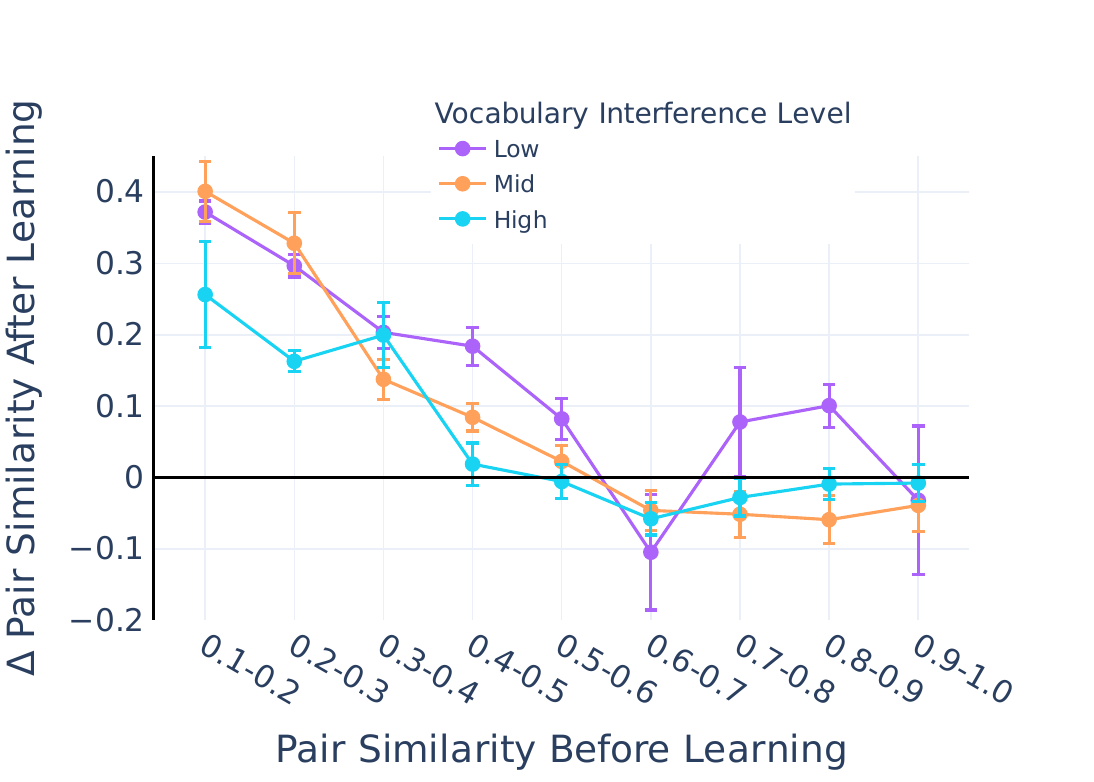}
        \caption{Control for pair similarity.}
        \label{fig:reprchange_pairsim_vs_vocabsim_initialpairmates}
    \end{subfigure}
    \begin{subfigure}{0.48\textwidth}
        \centering
        \includegraphics[width=\linewidth]{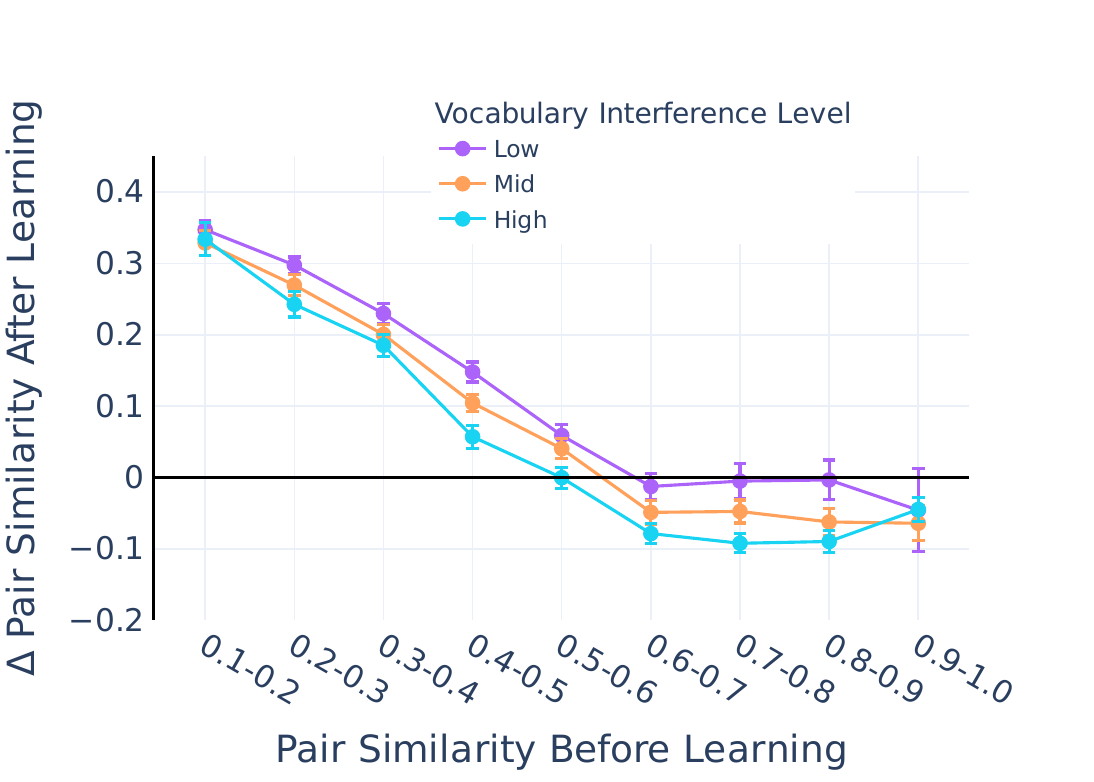}
        \caption{Control for pair similarity \& vocab. interference.}
        \label{fig:reprchange_pairsim_vs_vocabsim_controlledsampling}
    \end{subfigure}
    \hfill
    \caption{
        Effect of vocabulary interference on representational change across different pair similarity groups.
        (a) Results for our original token pairs, sampled uniformly with respect to pair similarity before learning (x-axis). We observe a consistent integration trend for low-similarity pairs and a shift toward differentiation for mid-similarity pairs ($0.6$–$0.7$). High-similarity pairs show more heterogeneous behavior, where low interference tends to promote integration, while higher interference tends to yield differentiation.
        (b) Results for extended token pairs, sampled uniformly over pair similarity before learning (x-axis) and vocabulary interference level (colored lines). 
        Higher vocabulary interference consistently leads to more differentiation, especially for mid- and high-similarity groups.
        These results suggest that while pairwise similarity is a key driver of differentiation, vocabulary interference amplifies this effect.
    }
    \label{fig:reprchange_pairsim_vs_vocabsim}
\end{figure*}

Figure~\ref{fig:reprchange_pairsim_vs_vocabsim_initialpairmates} shows how vocabulary interference modulates representational change, with the pattern of this effect varying depending on the pairs' similarity before learning.
For low-similarity pairs (up to $0.5-0.6$), we observe consistent integration at all levels of vocabulary interference.
For mid-similarity pairs ($0.6-0.7$), differentiation emerges as the primary driver across interference levels.
High-similarity pairs (above $0.7$), however, display high variability and heterogeneous effects: lower interference tends to promote integration, whereas higher interference tends to yield differentiation.
This heterogeneity may help explain the apparent U-shaped pattern in our earlier analysis: while low- and mid-similarity pairs exhibit seemingly consistent behavior across interference levels, the variability among high-similarity pairs can mask these opposing trends when averaged, leading to an apparent lack of representational change.

\paragraph{Sampling the full spectrum of vocabulary interference.}
These findings suggest an interaction between pairwise similarity and vocabulary interference, particularly in the high-similarity regime.
To more directly test this interaction, we next control for both factors simultaneously by examining token pairs that span the full joint distribution of pairwise similarity and vocabulary interference (see Section \ref{subsec:vocab_methods} and Appendix~\ref{appendix:methods} for details).
To do this, we form an extended set of $(x,y)$ tokens pairs, $\mathcal{Q}^m$, by sampling additional token pairs uniformly across vocabulary interference values. Our approach ensured a minimum of $10$ representative pairs per model, similarity group and vocabulary interference level. 

Figure~\ref{fig:reprchange_pairsim_vs_vocabsim_controlledsampling} shows the results under this controlled sampling regime, where both pair similarity and vocabulary interference are explicitly balanced.
As before, we observe a consistent pattern of integration for low-similarity pairs (up to $0.5-0.6$).
Yet now we observe a robust effect of vocabulary interference: the curves for higher interference levels lie below those for lower interference, indicating reduced, but still present, integration.
For mid-similarity pairs, we observe differentiation across all levels of vocabulary interference, and the effect is stronger under higher interference.
For high-similarity pairs, we observe a distinct trend: under moderate or high interference, these pairs clearly differentiate, while under low interference, their representations remain relatively stable.

Importantly, across all similarity levels, we observe that higher vocabulary interference is consistently associated with reduced pairwise similarity after learning.
One possible explanation for this pattern is that increased interference introduces greater competitive pressure: to reliably associate with each other, paired tokens must distinguish themselves from many similar distractors in the vocabulary.
This competition drives the model to reshape representations not only to encode the intended association, but also to preserve distinctiveness within the broader context of the model’s prior knowledge.

\paragraph{Representational change through the lens of vocabulary interference.}
The idea that increased interference introduces competitive pressure provides a useful lens for interpreting the distinct behaviors observed across different pair similarity regimes.
For instance, one possible interpretation of the robust integration observed among low-similarity pairs, regardless of vocabulary interference level, is that their representational distance before learning provides greater flexibility for alignment.
Because these pairs begin far apart in the representation space, the model can bring them closer (i.e., integrate) without risking excessive overlap that would compromise their individual distinguishability.
In this regime, vocabulary interference may impose relatively weak constraints, since the updated representations remain unlikely to be confounded with each other---thus, integration can proceed even under high interference.  
We speculate that this relative freedom from competition allows the model to prioritize pairwise association without necessitating broader adjustments across the vocabulary space.

At the other extreme, high-similarity pairs begin very close in representational space.
Under high vocabulary interference, the model must reshape these representations to prevent confusion with nearby distractors---yet increasing their similarity further could risk entanglement.
As a result, differentiation becomes the most likely direction of change, consistent with the strong divergence we observe under high vocabulary interference.
In contrast, when vocabulary interference is low, these pairs are already well isolated from the rest of the vocabulary, reducing the pressure for representational differentiation.

Mid-similarity pairs lie in a ``sensitive zone'' where both factors---pairwise similarity and vocabulary interference---interact most dynamically.
They are similar enough to each other that further integration might risk overlap to the point of risking their distinction, yet not similar enough to be clearly associated.
Consequently, differentiation appears to be the primary orientation of change for mid-similarity pairs, intensifying with greater vocabulary interference to preserve separability.
This suggests that mid-similarity pairs are especially vulnerable to representational reorganization, regardless of the specific interference level.

Therefore, the observed U-shaped curve in the previous analysis (Section~\ref{subsec:reprchange_pairsim}) may be partially explained by a nuanced interaction between pairwise similarity and vocabulary interference.
In the high-similarity range, pairs fragment into opposing behaviors across levels of vocabulary interference, so that averaging over these heterogeneous effects can mask systematic representational change and create an illusion of stability.

\section{Discussion}
\label{sec:discussion}



In this paper, we investigate whether LLMs exhibit representational changes during associative learning that mirror those observed in humans, and whether they help disambiguate between competing hypotheses about how such changes unfold.
Controlling for within pair similarity, we found a non-monotonic pattern of representational change, consistent with the NMPH.
This pattern is observed when models stabilize their learning, in what we name the \emph{Consolidation} phase. The fact that LLMs naturally give rise to these dynamics---without any task-specific optimization, and under conditions aligned with how humans learn associations---suggests that they may serve as emergent, flexible model organisms for studying memory reorganization in the brain.

We then leverage the controllability of LLMs to investigate how the vocabulary interference--defined as the interaction between token pair similarity and their similarity to the broader vocabulary--affects representational changes.
By introducing this second dimension of analysis, we show that representational dynamics cannot be fully understood in terms of pairwise similarity alone.
Instead, representational change reflects a joint influence of pairwise similarity and global contextual competition within the model’s prior knowledge.
This interaction is especially evident at the extremes: low-similarity pairs integrate consistently across all interference levels, suggesting greater flexibility due to low risk of confusion.
High-similarity pairs, by contrast, are already near each other in representational space and face stronger constraints: when vocabulary interference is high, differentiation is the only viable way to maintain separability, whereas under low interference, they remain relatively insulated from external competition, reducing the pressure for further differentiation.
Mid-similarity pairs appear to lie at a critical boundary--similar enough to risk confusion, yet not similar enough to form a strong association--making them particularly susceptible to interference-induced differentiation.
This sensitivity highlights how small shifts in competitive context can alter the direction of representational change.

Our results show that, while pairwise similarity is a key determinant of representational change, vocabulary interference modulates this effect.
This interaction between pair association strength and global contextual interference reveals richer representational dynamic than previously assumed, and may help reconcile diverging findings in the neuroscience literature, where such vocabulary-level interference remains difficult to assess due to limited access to global representational structure.
Critically, this kind of fine-grained, systematic manipulation is difficult to achieve in human studies, where both pairwise similarity and global interference are hard to quantify and control.
LLMs thus serve as powerful computational model organisms for testing hypotheses about memory dynamics, offering a level of scale and experimental control that is rarely achievable in biological systems.

\paragraph{Limitations and future work.}
Although LLMs differ mechanistically from human brains, they provide a valuable model system for generating and testing hypotheses that are otherwise challenging to examine in biological systems.
Nonetheless, they are not direct stand-ins for humans, and empirical validation in human studies remains essential.

Our operationalization of vocabulary interference also has limitations. By design, it estimates representational competition from the broader vocabulary space, but this approximation may not fully capture the dynamics of interference in human memory, where similarity is shaped by experience, attention, and context. Furthermore, our measure relies on sampled subsets of tokens for tractability, which may underrepresent the true structure of competition across the full vocabulary.

Methodologically, our analysis focused on the final hidden layer of each model, with only preliminary exploration of earlier layers. Future work could systematically track representational change across layers, providing an extended analysis of how interference and differentiation emerge throughout the model hierarchy.
Additionally, to ensure coverage across a wide range of similarity values, we used token pairs defined by geometric properties rather than naturalistic semantics.
A preliminary analysis on WordNet stimuli is provided in Appendix~\ref{appendix:sec_analysis_wordnet}, with further investigation left for future work.

Finally, this work focuses on hypotheses tested in mature adult brains, leaving open the question of how these processes emerge during development.
Promising future directions include exploring curriculum-learning setups that more closely mirror human developmental trajectories, analyzing attention patterns to identify circuit-level mechanisms involved in resolving interference, and examining how representational dynamics evolve during fine-tuning and longer-term learning.

\clearpage

\section*{Acknowledgments}
\label{sec:acknowledgements}

This work was partially funded by the German Research Foundation (DFG) - DFG Research Unit FOR 5368 and by the Max Planck Institute for Software Systems graduate center.
We thank Omer Moussa and Mathis Pink for providing valuable feedback on earlier versions of this work.

\bibliographystyle{plain}
\bibliography{neurips_2025}

\clearpage

\newpage
\section*{NeurIPS Paper Checklist}
 
\begin{enumerate}

\item {\bf Claims}
    \item[] Question: Do the main claims made in the abstract and introduction accurately reflect the paper's contributions and scope?
    \item[] Answer: \answerYes{} 
    \item[] Justification: In the abstract and at the end of the introduction.

\item {\bf Limitations}
    \item[] Question: Does the paper discuss the limitations of the work performed by the authors?
    \item[] Answer: \answerYes{} 
    \item[] Justification: We have discussed the limitations of our work in Section~\ref{sec:discussion}.

\item {\bf Theory assumptions and proofs}
    \item[] Question: For each theoretical result, does the paper provide the full set of assumptions and a complete (and correct) proof?
    \item[] Answer: \answerNA{} 
    \item[] Justification: Our paper does not involve any proofs or assumptions.

    \item {\bf Experimental result reproducibility}
    \item[] Question: Does the paper fully disclose all the information needed to reproduce the main experimental results of the paper to the extent that it affects the main claims and/or conclusions of the paper (regardless of whether the code and data are provided or not)?
    \item[] Answer: \answerYes{} 
    \item[] Justification: As described in Section~\ref{sec:method} and throughout the Appendix, we have provided detailed descriptions and analyses of the experimental setups for all our investigations.

\item {\bf Open access to data and code}
    \item[] Question: Does the paper provide open access to the data and code, with sufficient instructions to faithfully reproduce the main experimental results, as described in supplemental material?
    \item[] Answer: \answerYes{} 
    \item[] Justification: We have attached with the submission the code necessary to reproduce our main results and upon acceptance we will publicly release it.

\item {\bf Experimental setting/details}
    \item[] Question: Does the paper specify all the training and test details (e.g., data splits, hyperparameters, how they were chosen, type of optimizer, etc.) necessary to understand the results?
    \item[] Answer: \answerYes{} 
    \item[] Justification: As detailed in Section~\ref{sec:method}, we have thoroughly described the experimental setups for all our experiments. Additional details have been provided in the Appendix.

\item {\bf Experiment statistical significance}
    \item[] Question: Does the paper report error bars suitably and correctly defined or other appropriate information about the statistical significance of the experiments?
    \item[] Answer: \answerYes{} 
    \item[] Justification: We have included error bars and statistical tests in Section~\ref{sec:results} and their corresponding explanations.

\item {\bf Experiments compute resources}
    \item[] Question: For each experiment, does the paper provide sufficient information on the computer resources (type of compute workers, memory, time of execution) needed to reproduce the experiments?
    \item[] Answer: \answerYes{} 
    \item[] Justification: As detailed in Section~\ref{sec:method}, we outline the specific model compute resources provided. 
    
\item {\bf Code of ethics}
    \item[] Question: Does the research conducted in the paper conform, in every respect, with the NeurIPS Code of Ethics \url{https://neurips.cc/public/EthicsGuidelines}?
    \item[] Answer: \answerYes{} 
    \item[] Justification: We are convinced that we comply with NeurIPS Code of Ethics.

\item {\bf Broader impacts}
    \item[] Question: Does the paper discuss both potential positive societal impacts and negative societal impacts of the work performed?
    \item[] Answer: \answerYes{} 
    \item[] Justification: We have addressed the broader impacts of our work in Section~\ref{sec:discussion}. Additionally, as our research is primarily an empirical exploration and poses no additional social risks, we have not included a discussion on potential harmfulness.
    
\item {\bf Safeguards}
    \item[] Question: Does the paper describe safeguards that have been put in place for responsible release of data or models that have a high risk for misuse (e.g., pretrained language models, image generators, or scraped datasets)?
    \item[] Answer: \answerNA{} 
    \item[] Justification: The paper poses no such risks.

\item {\bf Licenses for existing assets}
    \item[] Question: Are the creators or original owners of assets (e.g., code, data, models), used in the paper, properly credited and are the license and terms of use explicitly mentioned and properly respected?
    \item[] Answer: \answerYes{} 
    \item[] Justification: We use LLM models and they are properly credited in Section~\ref{sec:method}.

\item {\bf New assets}
    \item[] Question: Are new assets introduced in the paper well documented and is the documentation provided alongside the assets?
    \item[] Answer: \answerNA{} 
    \item[] Justification: The paper does not release new assets.

\item {\bf Crowdsourcing and research with human subjects}
    \item[] Question: For crowdsourcing experiments and research with human subjects, does the paper include the full text of instructions given to participants and screenshots, if applicable, as well as details about compensation (if any)? 
    \item[] Answer: \answerNA{} 
    \item[] Justification: The paper does not involve crowdsourcing nor research with human subjects.

\item {\bf Institutional review board (IRB) approvals or equivalent for research with human subjects}
    \item[] Question: Does the paper describe potential risks incurred by study participants, whether such risks were disclosed to the subjects, and whether Institutional Review Board (IRB) approvals (or an equivalent approval/review based on the requirements of your country or institution) were obtained?
    \item[] Answer: \answerNA{} 
    \item[] Justification: The paper does not involve crowdsourcing nor research with human subjects.

\item {\bf Declaration of LLM usage}
    \item[] Question: Does the paper describe the usage of LLMs if it is an important, original, or non-standard component of the core methods in this research? Note that if the LLM is used only for writing, editing, or formatting purposes and does not impact the core methodology, scientific rigorousness, or originality of the research, declaration is not required.
    \item[] Answer: \answerYes{} 
    \item[] Justification: We describe the use of LLM models in Section~\ref{sec:method} and in our Appendix.

\end{enumerate}

\clearpage

\begin{appendices}


\section{Methods}
\label{appendix:methods}

\subsection{Models}
\label{appendix:subsec_models}

All models used in our study are listed in Table~\ref{tab:appendix_models_details}.
We employ the base versions (i.e., without fine-tuning), since our prompts do not include any instructions in its format, as described in Section~\ref{appendix:subsec_prompt}.

\begin{table}[H]
\begin{minipage}\textwidth
    \caption{Details on models used in our study, including maximum context length.}
    \vspace{1em}
    \centering
    \begin{tabular}{ccccc}
        Architecture & Version & Size & Context Length \\
        \toprule
        \multirow{4}{*}{Llama} & 2 & 7b & 4k \\
                               & 3.1 & 8b & 132k \\
                               & 3.2 & 1b & 132k \\
                               & 3.2 & 3b & 132k \\
        \multirow{1}{*}{Gemma} & 2 & 9b & 8k \\
        \multirow{1}{*}{Mistral} & 0.1 & 7b & 4k~\footnote{Original context window of 4k, but extendable to 16k with a sliding window attention (SWA) mechanism.} \\ 
    \end{tabular}
    \label{tab:appendix_models_details}
\end{minipage}
\end{table}


\subsection{Prompt}
\label{appendix:subsec_prompt}

We format our prompts by presenting the token pairs $(x,y)$ as direct concatenations without any separator, punctuation, or instructional context.
For instance, if the pair is ($A$, $B$), the prompt would contain $AB$ (for $r=1$) without a space or symbol between them.
This minimal setup ensures that the model relies purely on co-occurrence patterns to form associations, rather than leveraging syntactic or structural cues.
All models under study include a beginning-of-sequence (BOS) token, which we consistently use as the first token in every prompt.
To avoid degenerate token pairs, we restrict the vocabulary space by excluding stop words, punctuation, and numerals.

\subsection{Vocabulary sampling}
\label{appendix:subsec_vocabsampling}

As detailed in Section~\ref{subsec:vocab_methods}, for each model, we randomly sampled $1,000$ tokens from $\mathcal{V}$ to form the representative subset $\tilde{\mathcal{V}}$, resulting in approximately $1$M pairwise token combinations.

\begin{figure}[H]
    \centering

    \begin{subfigure}{0.48\textwidth}
        \centering
        \includegraphics[width=\linewidth,height=4cm]{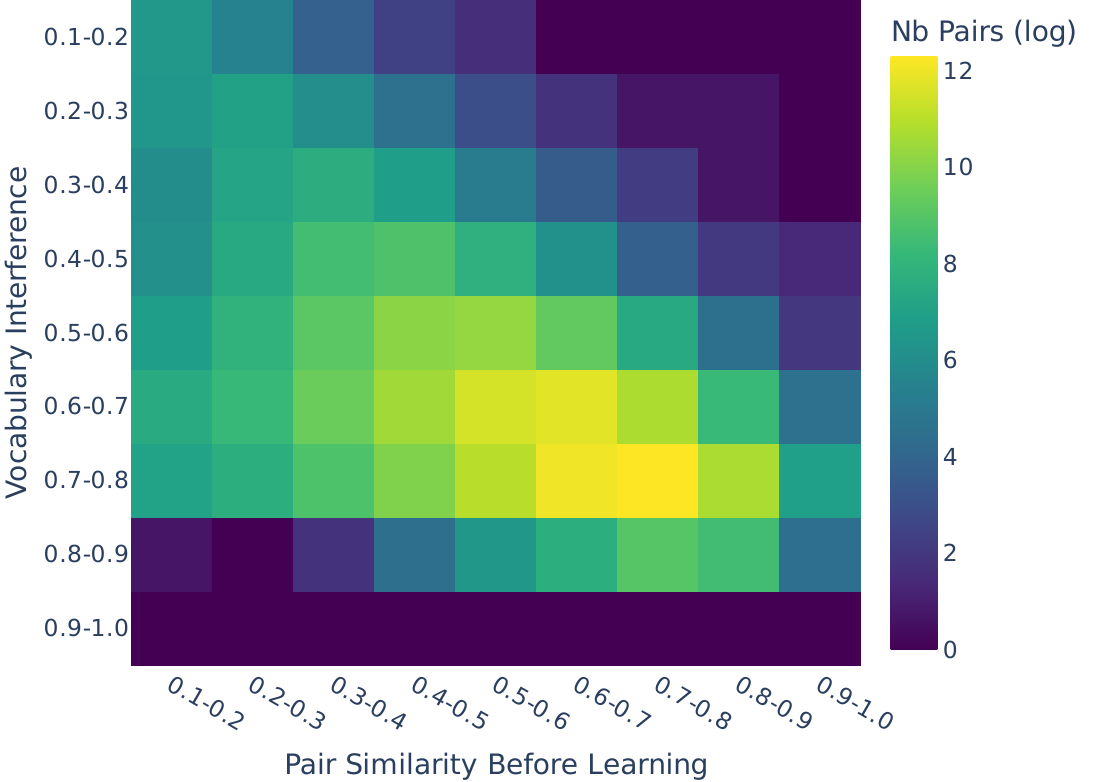}
        \caption{Llama2-7b}
    \end{subfigure}
    \hfill
    \begin{subfigure}{0.48\textwidth}
        \centering
        \includegraphics[width=\linewidth,height=4cm]{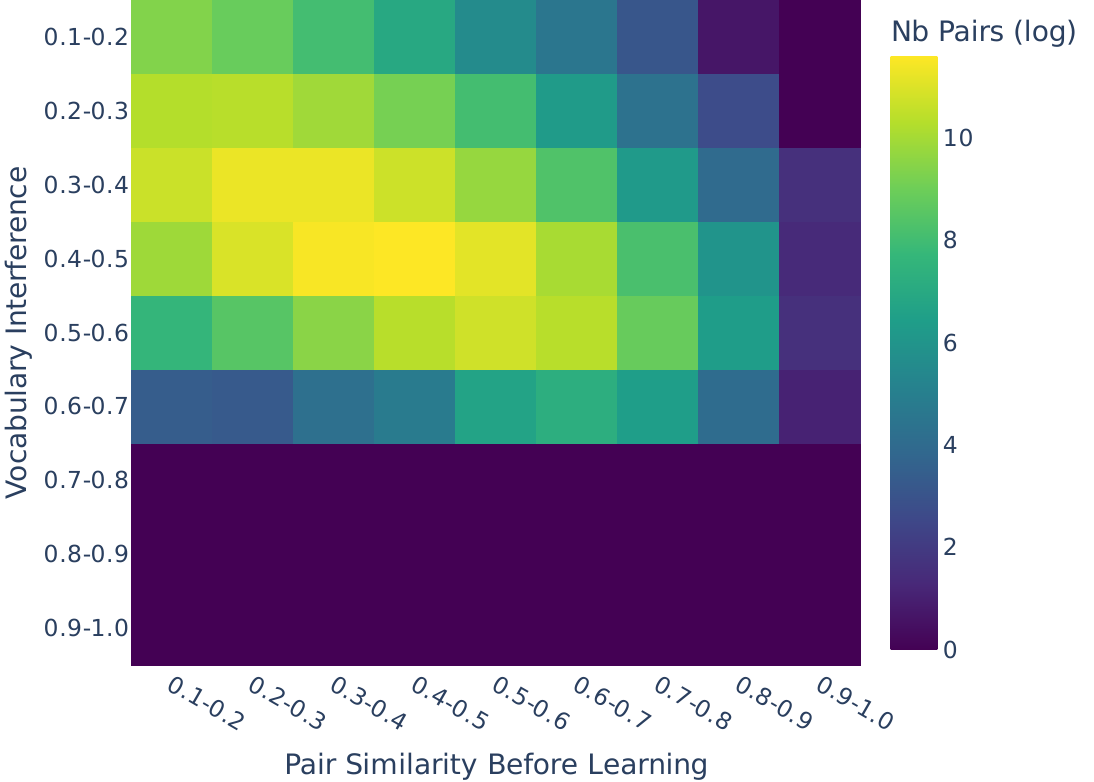}
        \caption{Mistral-7b}
    \end{subfigure}
    
    \begin{subfigure}{0.48\textwidth}
        \centering
        \includegraphics[width=\linewidth,height=4cm]{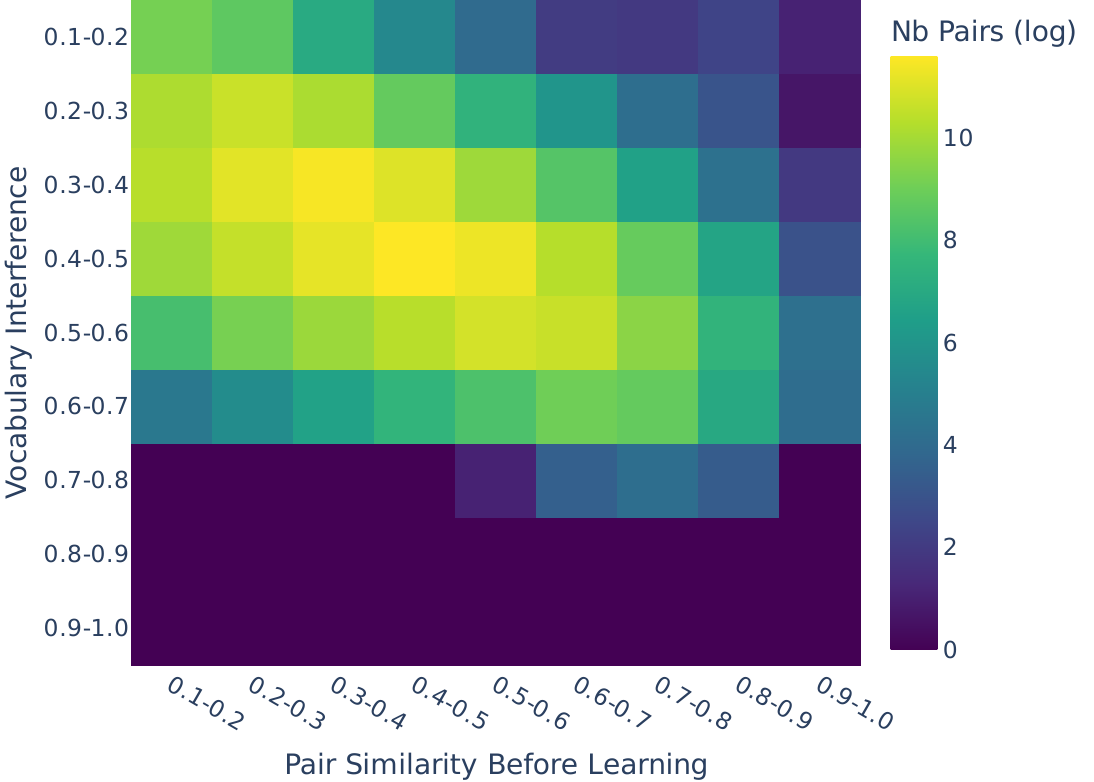}
        \caption{Gemma2-9b}
    \end{subfigure}
    \hfill
    \begin{subfigure}{0.48\textwidth}
        \centering
        \includegraphics[width=\linewidth,height=4cm]{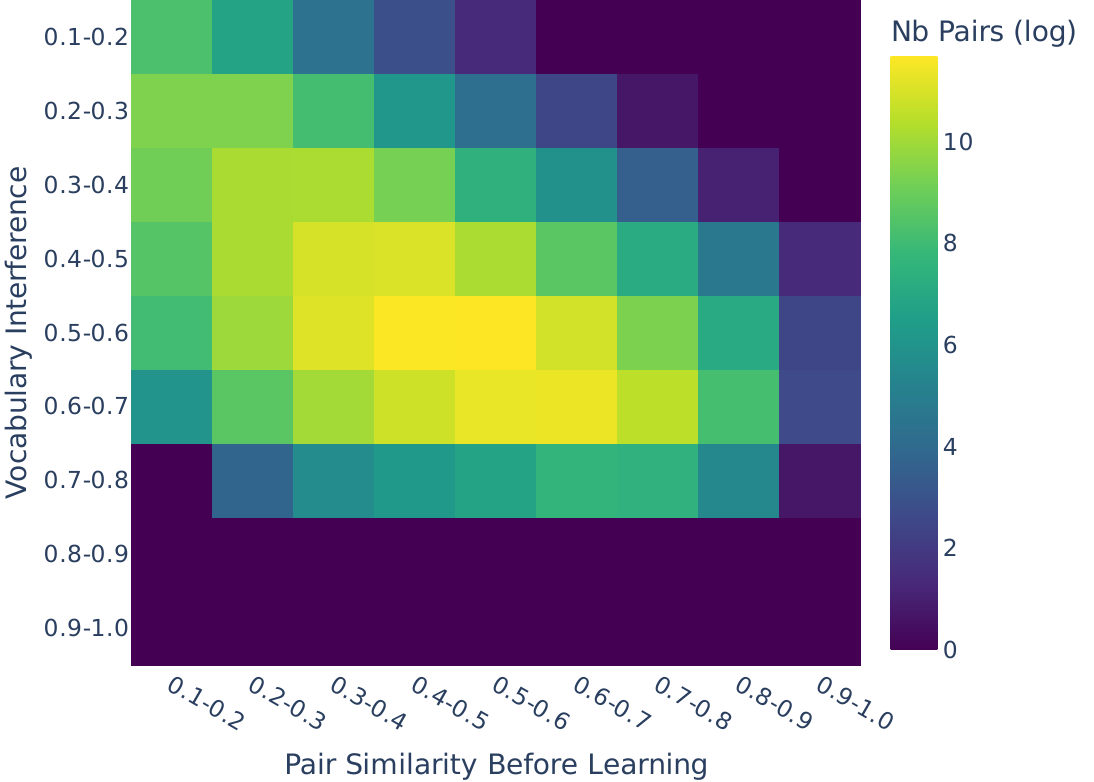}
        \caption{Llama3.1-8b}
    \end{subfigure}

    \begin{subfigure}{0.48\textwidth}
        \centering
        \includegraphics[width=\linewidth,height=4cm]{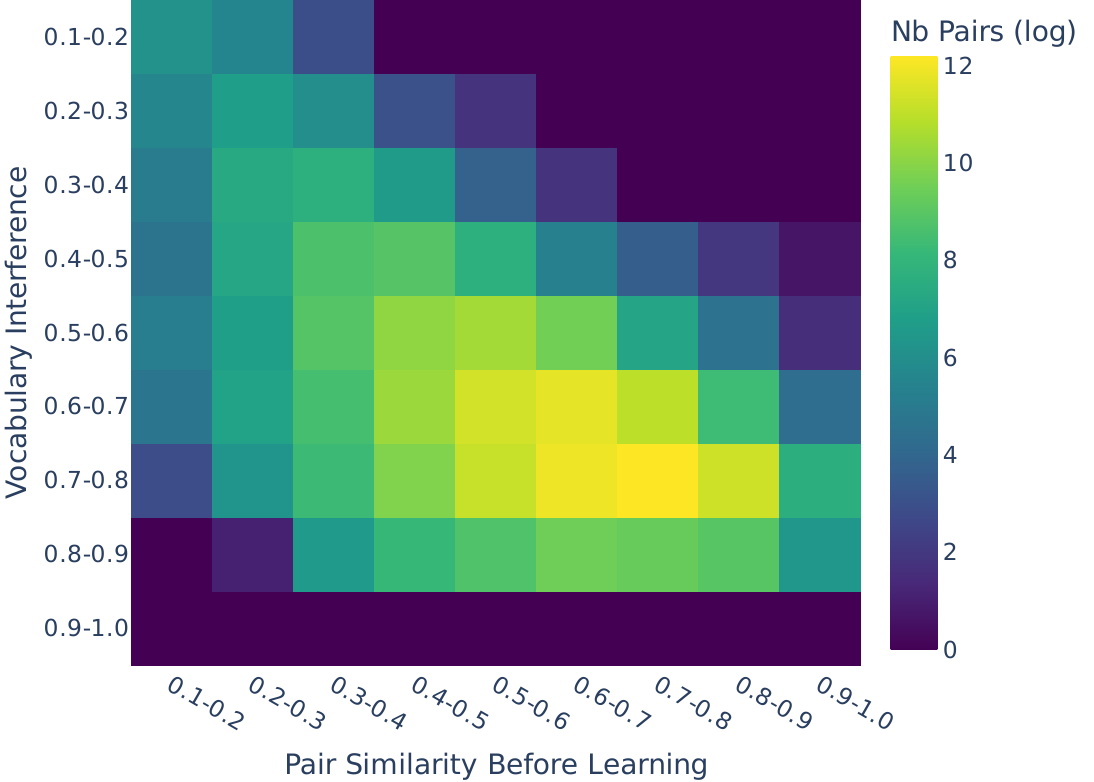}
        \caption{Llama3.2-1b}
    \end{subfigure}
    \hfill
    \begin{subfigure}{0.48\textwidth}
        \centering
        \includegraphics[width=\linewidth,height=4cm]{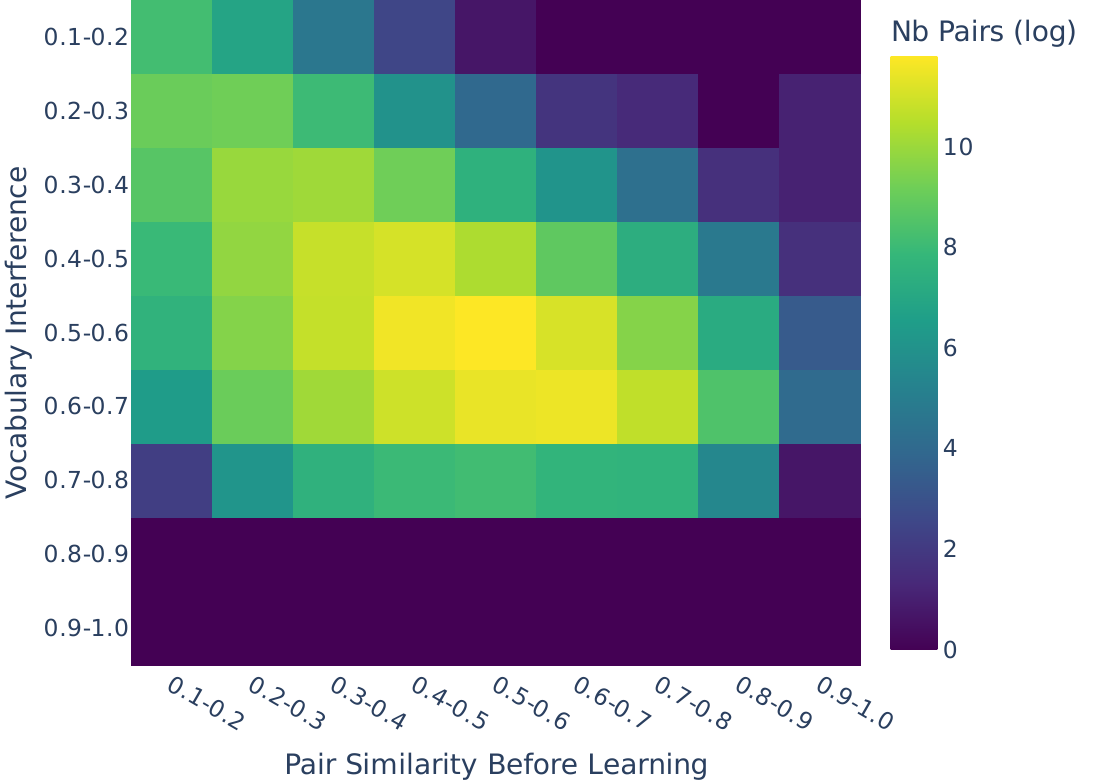}
        \caption{Llama3.2-3b}
    \end{subfigure}

    \begin{subfigure}{\textwidth}
        \centering
        \includegraphics[width=0.48\linewidth,height=4cm]{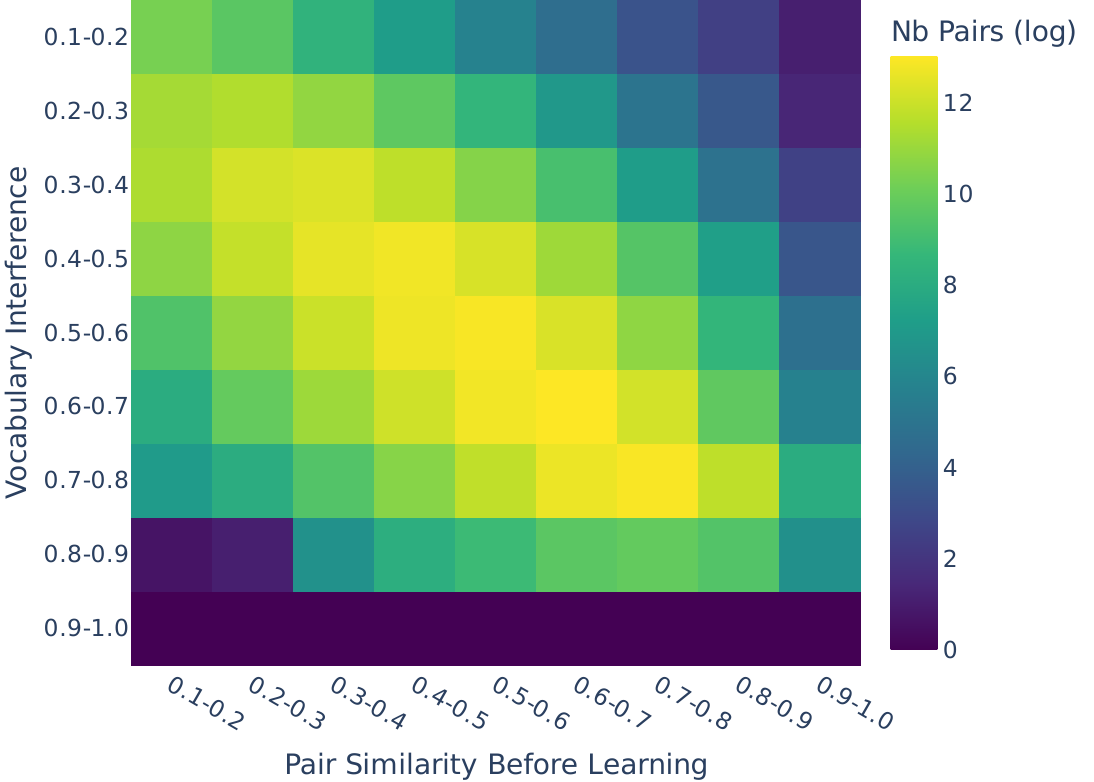}
        \caption{All models}
    \end{subfigure}
    \caption{
        Log-scale heatmap showing the joint distribution of token pairs across pairwise similarity before learning (x-axis) and vocabulary interference (y-axis) in the representative vocabulary subset $\tilde{\mathcal{V}}$. 
        Subplots (a–f) correspond to individual models; subplot (g) aggregates results across all models.
    }
    \label{fig:appendix_heatmap_pairs_pairsim_vs_vocabsim}
\end{figure}

Figure~\ref{fig:appendix_heatmap_pairs_pairsim_vs_vocabsim} presents heatmaps showing how the $1$ million sampled token pairs are distributed across pairwise similarity before learning (x-axis) and vocabulary interference (y-axis), for each individual model (a–f) and across all models combined (g).
The color scale indicates the log-transformed number of token pairs in each bin.
These distributions reflect the natural data availability prior to applying uniform sampling of $10$ items per bin. 
Overall, the density of sampled pairs tends to concentrate in the low-to-mid similarity and interference ranges, with some variation across models.

\begin{figure}[H]
    \centering

    \begin{subfigure}{0.48\textwidth}
        \centering
        \includegraphics[width=\linewidth,height=4cm]{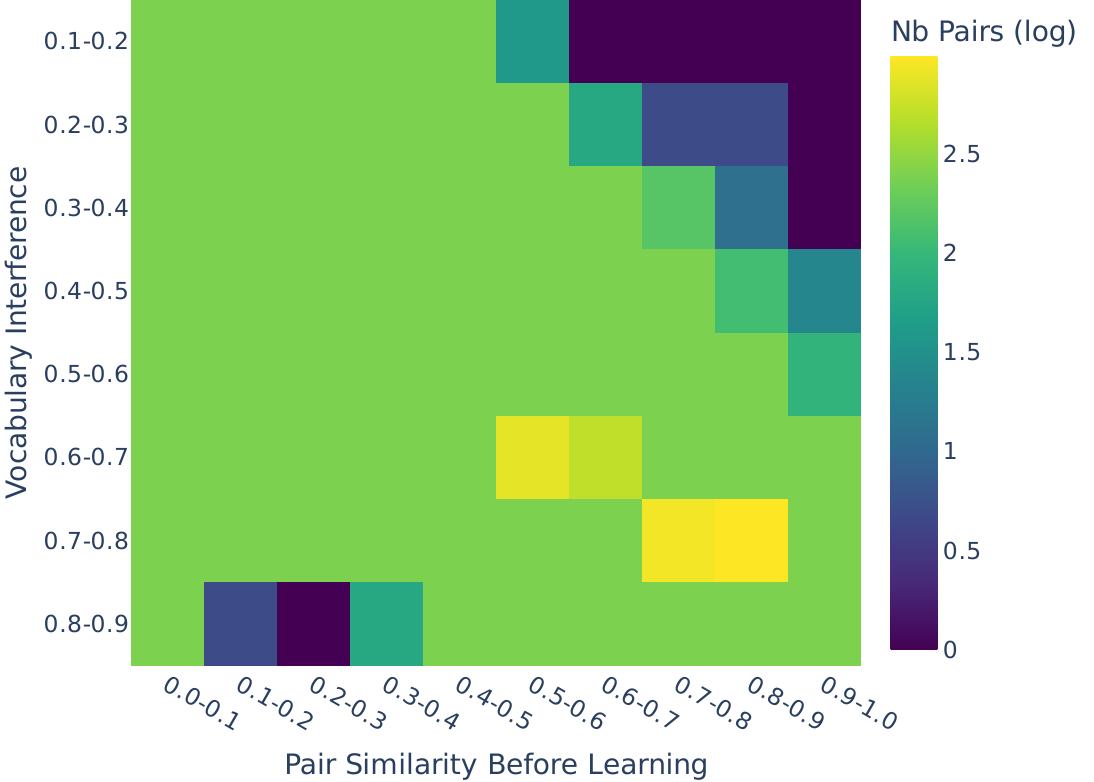}
        \caption{Llama2-7b.}
    \end{subfigure}
    \hfill
    \begin{subfigure}{0.48\textwidth}
        \centering
        \includegraphics[width=\linewidth,height=4cm]{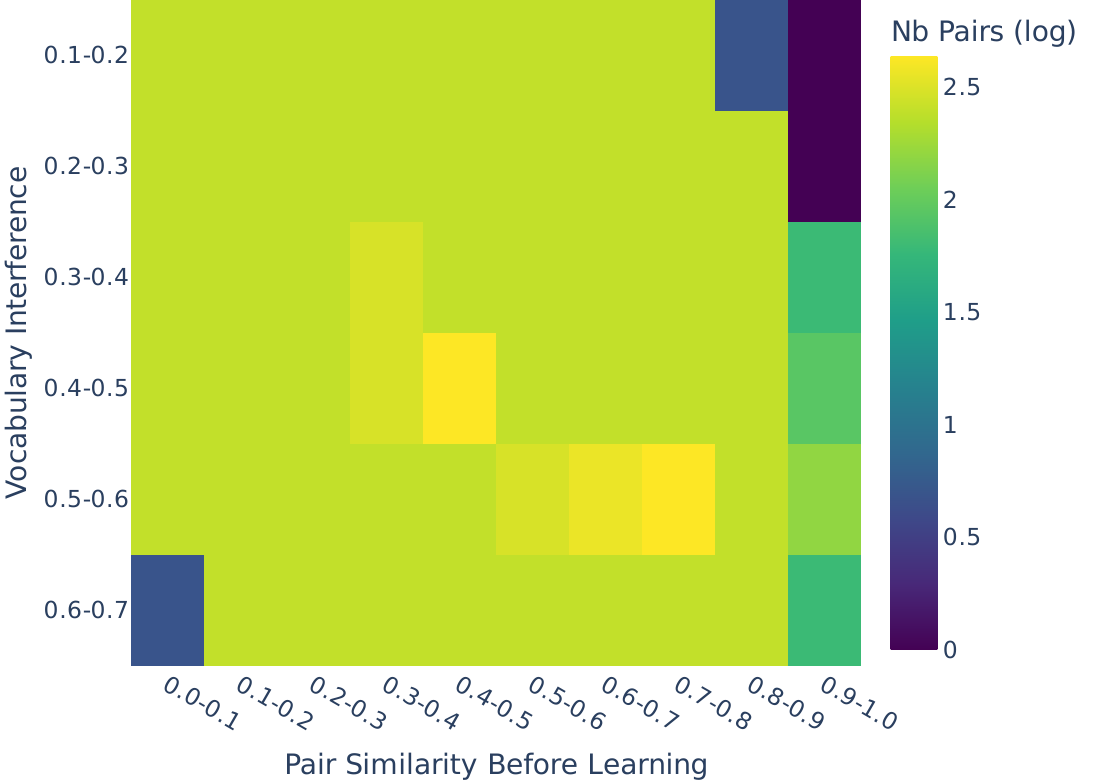}
        \caption{Mistral-7b.}
    \end{subfigure}
    
    \begin{subfigure}{0.48\textwidth}
        \centering
        \includegraphics[width=\linewidth,height=4cm]{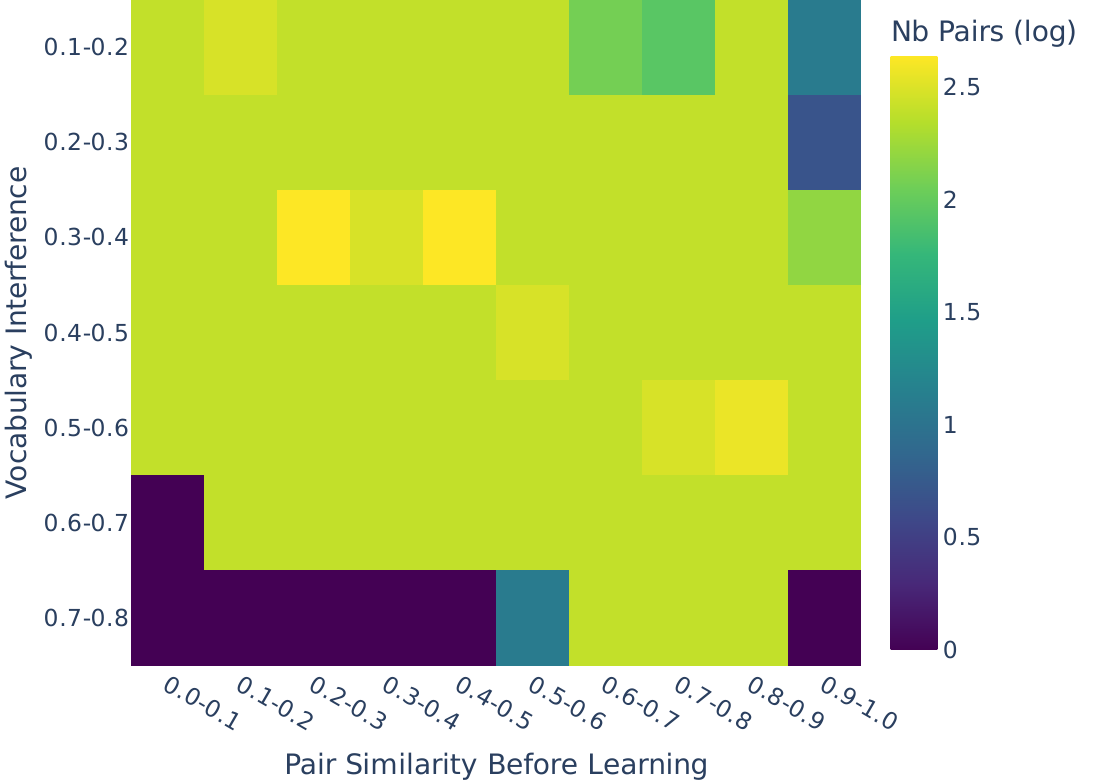}
        \caption{Gemma2-9b}
    \end{subfigure}
    \hfill
    \begin{subfigure}{0.48\textwidth}
        \centering
        \includegraphics[width=\linewidth,height=4cm]{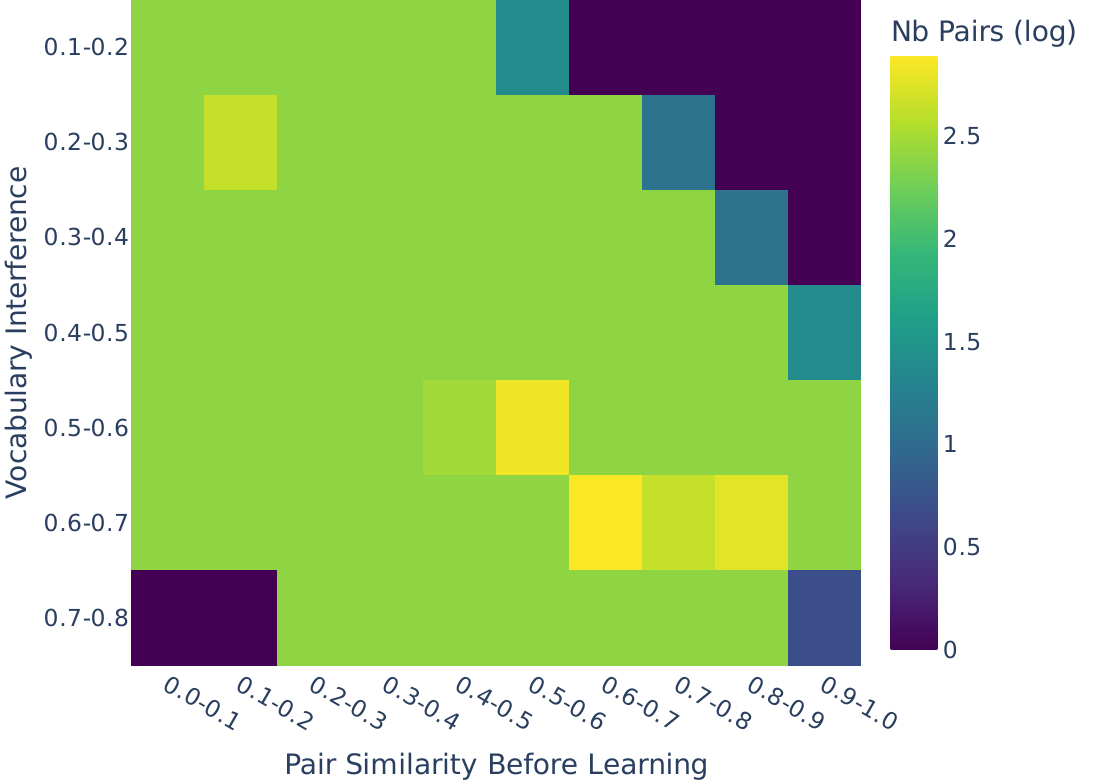}
        \caption{Llama3.1-8b.}
    \end{subfigure}

    \begin{subfigure}{0.48\textwidth}
        \centering
        \includegraphics[width=\linewidth,height=4cm]{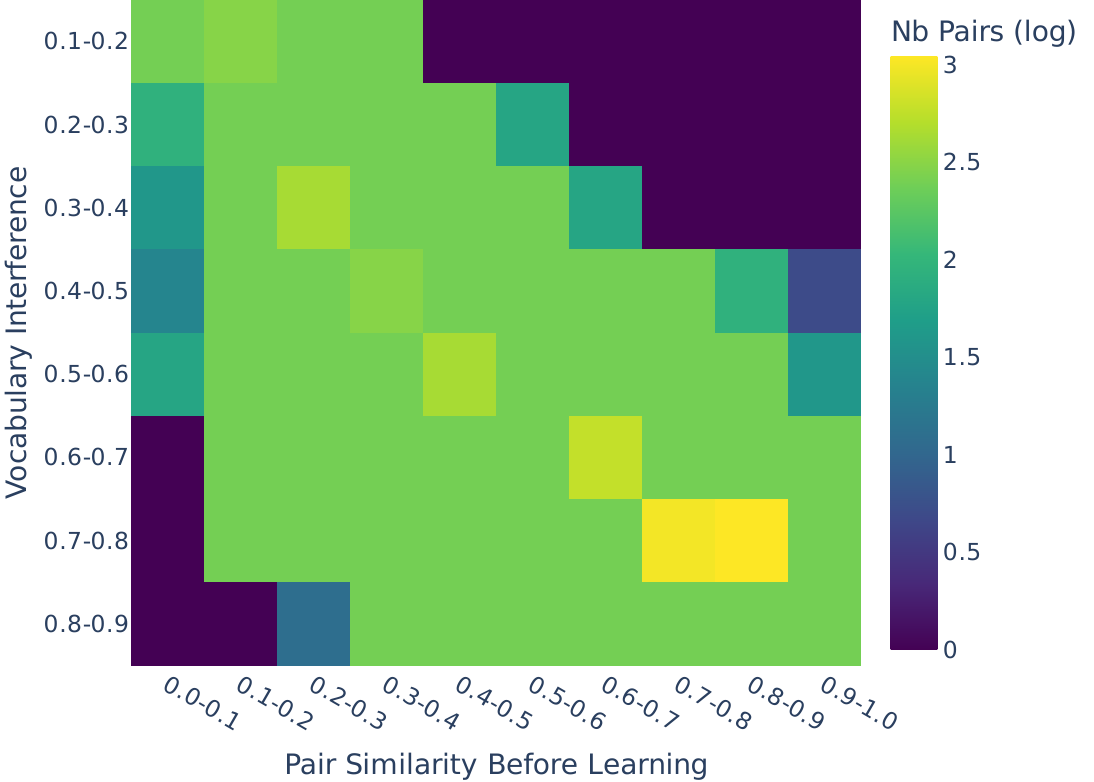}
        \caption{Llama3.2-1b.}
    \end{subfigure}
    \hfill
    \begin{subfigure}{0.48\textwidth}
        \centering
        \includegraphics[width=\linewidth,height=4cm]{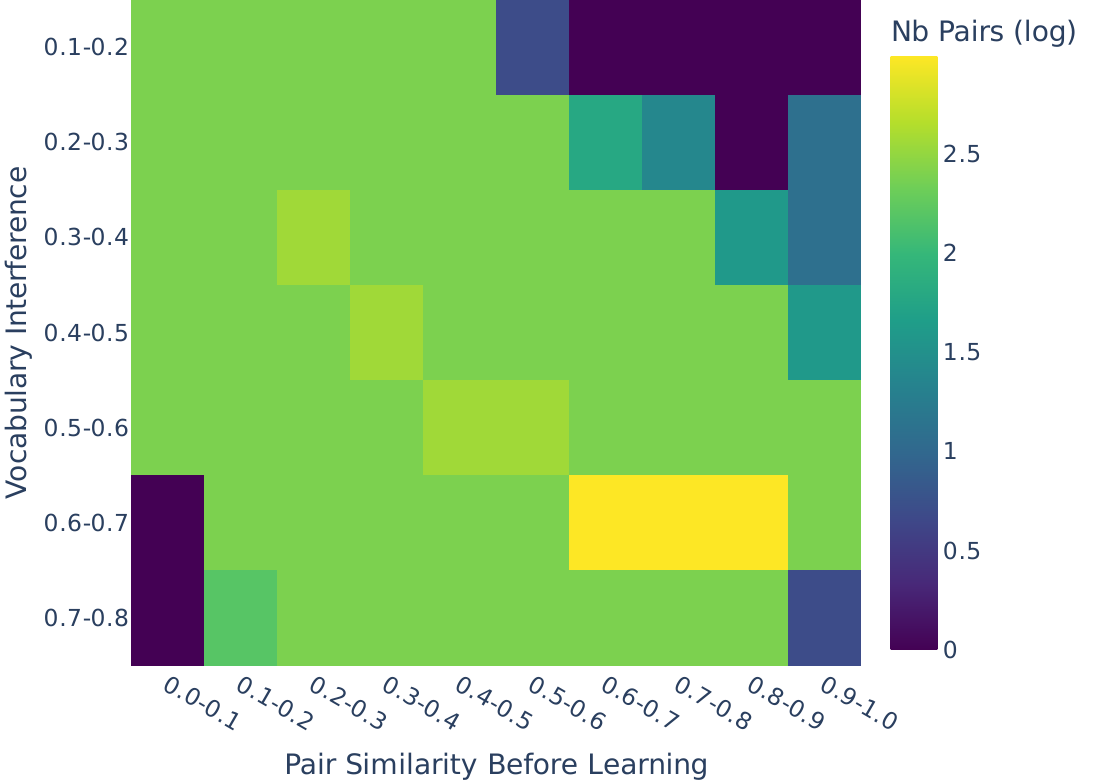}
        \caption{Llama3.2-3b.}
    \end{subfigure}

    \begin{subfigure}{\textwidth}
        \centering
        \includegraphics[width=0.48\linewidth,height=4cm]{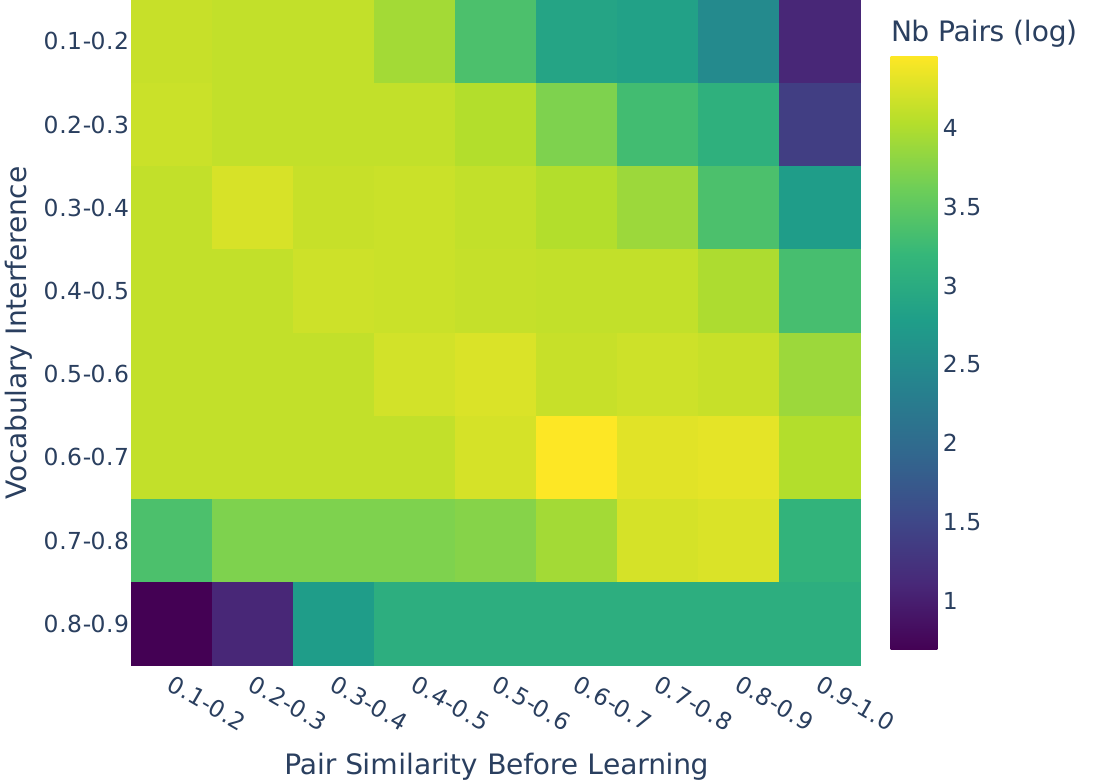}
        \caption{Combining all models.}
    \end{subfigure}
    \caption{
        Log-scale heatmap showing the joint distribution of token pairs across pairwise similarity before learning (x-axis) and vocabulary interference (y-axis) after uniformly sampling of $10$ items per pairwise similarity $\times$ vocabulary interference bin.
        Subplots (a–f) correspond to individual models; subplot (g) aggregates results across all models.
    }
    \label{fig:appendix_heatmap_pairs_pairsim_vs_vocabsim_uniform}
\end{figure}

To ensure balanced coverage across the pairwise similarity and vocabulary interference space, we applied a uniform sampling strategy to construct the set $\mathcal{Q}_m$ for each model. 
All token pairs were first assigned to bins according to their pair similarity and vocabulary interference values.
We then counted how many token pairs already existed in each bin from the original set $\mathcal{P}_m$, and filtered out any duplicates to avoid reusing token pairs.
The final set $\mathcal{Q}_m$ was created by combining the original and newly sampled pairs, resulting in an approximately uniform distribution of token pairs across the similarity–interference grid.
Figure~\ref{fig:appendix_heatmap_pairs_pairsim_vs_vocabsim_uniform} illustrates the resulting distributions after this sampling procedure.
Subfigures (a–f) show the heatmaps for each model individually, while (g) displays the combined heatmap representing the aggregated distribution across all models. 
Each cell indicates the (log-transformed) number of token pairs in the corresponding pair similarity $\times$ vocabulary interference bin.
As intended, the distributions are largely uniform, with minor deviations due to constraints in available data for certain bins.
\clearpage


\clearpage
\subsection{Defining levels of vocabulary interference}
\label{appendix:subsec_vocab_sim}

\begin{figure}[H]
    \centering
    \begin{subfigure}{0.55\textwidth}
        \centering
        \includegraphics[width=\linewidth]{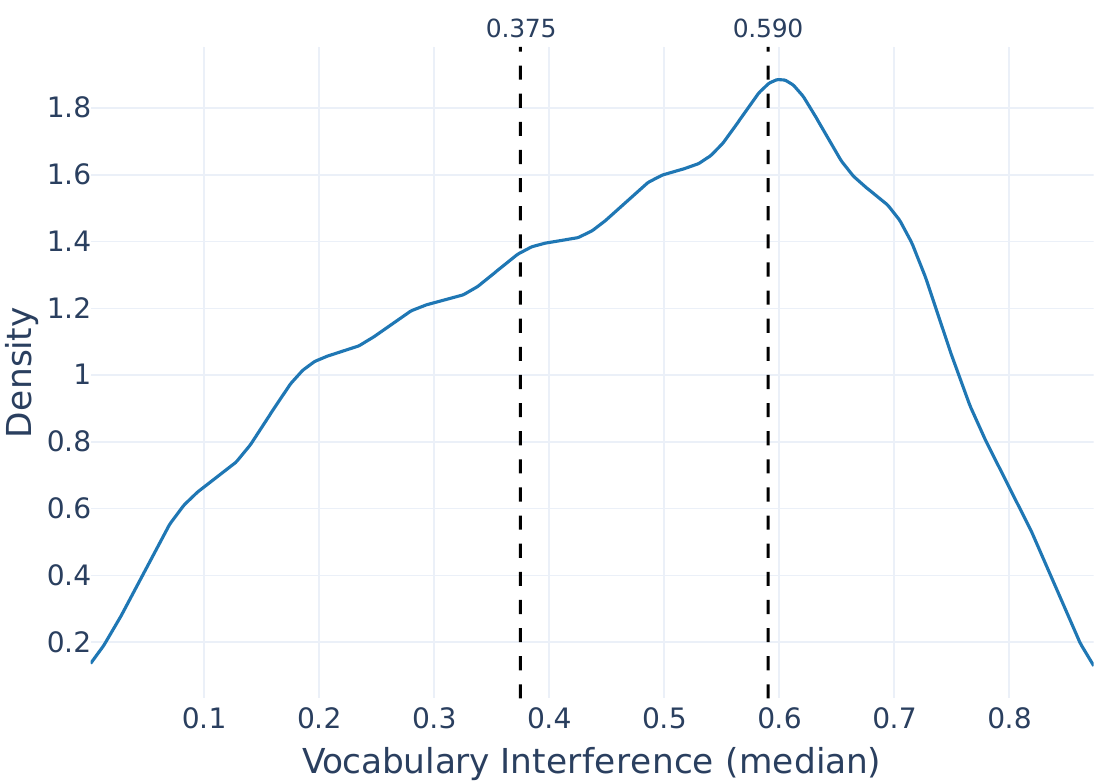}
        \caption{Vocabulary interference distribution.}
        \label{fig:appendix_vocabsim_distr_with_quantiles}
    \end{subfigure}
    \hfill
    \begin{subfigure}{0.40\textwidth}
        \centering
        \includegraphics[width=\linewidth]{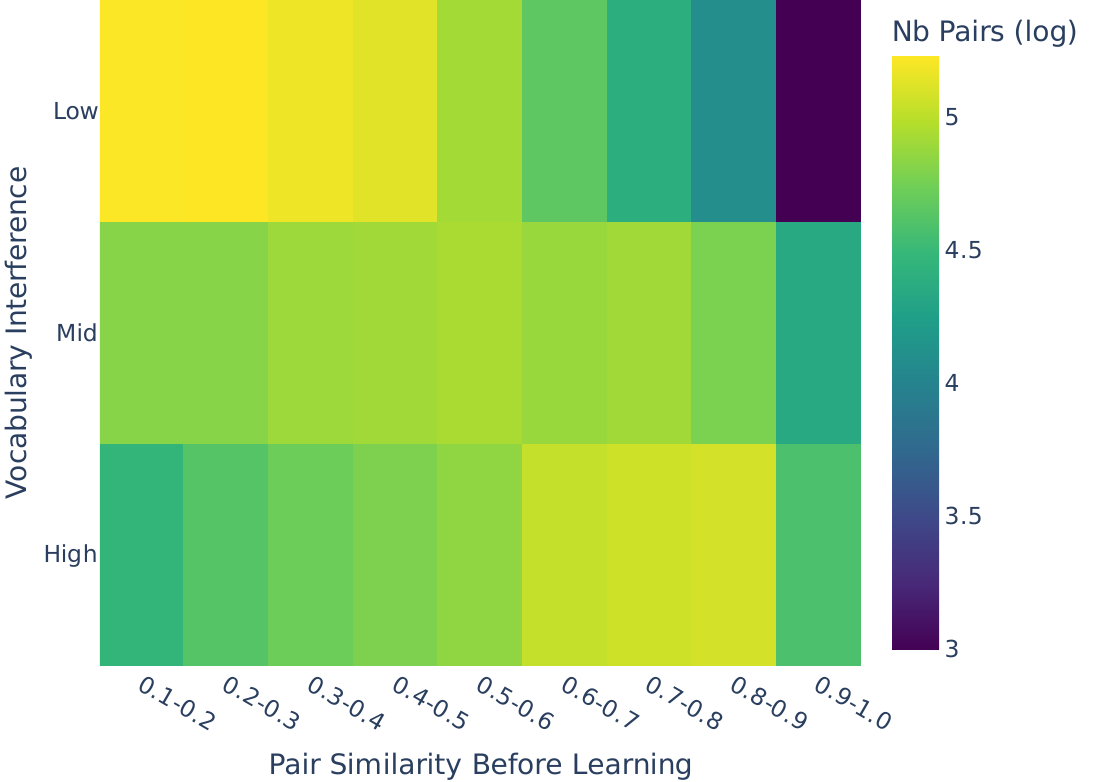}
        \caption{Number of token pairs (log scale).}
        \label{fig:appendix_paircount_vocabsim_vs_pairsim}
    \end{subfigure}

    \caption{
        (a) Distribution of vocabulary interference (median) values. Vertical lines show the thresholds used to equally split this distribution into Low, Mid and High similarity levels.
        (b) Heatmap showing the number of token pairs (log scale) as a function of pairwise similarity before learning (x-axis) and vocabulary interference level (y-axis) after uniformly sampling of $10$ items per pairwise similarity $\times$ vocabulary interference bin.
    }
    
\end{figure}

Figure~\ref{fig:appendix_vocabsim_distr_with_quantiles} shows a kernel density estimate (KDE) of the vocabulary interference scores (median values) computed across all evaluated token pairs.
To define the Low, Mid, and High interference categories, we divided the distribution into three quantiles, with the resulting quantile thresholds indicated by the vertical dashed lines. Figure~\ref{fig:appendix_paircount_vocabsim_vs_pairsim} reports the number of token pairs (log scale) used in our analysis, stratified by both vocabulary interference level and pair similarity prior to learning.

\subsection{Modification to Greedy Coordinate Descent (GCG) algorithm}
\label{appendix:subsec_cgc}

We repurpose the GCG~\cite{zou2023universal} method to minimize a loss defined over the cosine similarity of internal activations.
Specifically, we randomly sample a token $x$ and construct a starting input sequence $\mathbf{s} = [x_1, y_1]$, where $x_1=y_1$, i.e., the same token is used as starting point in both positions.
We then measure their pair similarity, $S^m_1=cos(\mathbf{h}^m_{x_1}, \mathbf{h}^m_{y_1})$.
We fix $x_1$, and our goal is to iteratively replace $y_1$ until the pair similarity
converges to the target interval $[\theta_{\min}^g, \theta_{\max}^g)$.
To find a suitable replacement, we define a loss function for each group to target the midpoint of the interval, $\mathcal{L}_g= (\frac{\theta_{\max}^g-\theta_{\min}^g}{2} - S^m_{1})^2$.
We then compute its gradient with respect to the one-hot encoding of $y_1$.
This produces a vector indicating how sensitive the loss is to each token in the vocabulary, which we then use to guide the search for a more suitable substitution, without updating the model’s weights.

Next, we identify the top-$k$ ($k=256$) tokens associated with the steepest decrease in loss (i.e. the most negative gradients).
These candidates serve as a rough approximation of the most promising substitutions, obtained via a first-order Taylor expansion.
We randomly shuffle this top-$k$ subset and evaluate each candidate sequentially by constructing a modified input sequence and computing the loss for each candidate pair.
The candidate yielding the smallest loss (i.e., closest cosine similarity to the target range) is selected as the updated token for $y_1$ on this iteration.
This procedure is repeated for a fixed number of iterations ($it=100$) or until the similarity score falls within the target interval.
If convergence is not achieved within the allotted iterations, the process restarts from a newly sampled initial token pair.

\clearpage 
\section{Supplementary analyses of the main paper results}
\label{appendix:sec_analysis_per_model}

\subsection{Example of token pairs}
\label{appendix:synthetic_token_pair_examples}

\begin{table}[h!]
\centering
\begin{tabular}{llll}
    \textbf{Group} & \textbf{Pair 1} & \textbf{Pair 2} & \textbf{Pair 3} \\
    \toprule
    0.1--0.15 & (Liter, CLARE) & (artifactId, gew) & (emat, SOUR) \\
    0.15--0.2 & (Ste, UITableView) & (Pers, pmatrix) & (ries, pragma) \\
    0.2--0.25 & (it, Autres) & (bt, Autres) & (Bad, tf) \\
    0.25--0.3 & (VD, Autres) & (Vertical, ierte) & (coordinate, gesch) \\
    0.3--0.35 & (elf, ScrollView) & (Tr, named) & (Else, newcommand) \\
    0.35--0.4 & (DER, stackexchange) & (ific, ently) & (von, trightarrow) \\
    0.4--0.45 & (uk, ThreadPool) & (vez, ISBN) & (under, rov) \\
    0.45--0.5 & (illet, cially) & (icio, atr) & (ptop, Wikimedia) \\
    0.5--0.55 & (bootstrap, rach) & (utes, Vorlage) & (iveau, tersuch) \\
    0.55--0.6 & (mittel, umbn) & (Series, notify) & (Problem, emptyset) \\
    0.6--0.65 & (fte, zott) & (Length, TRUE) & (elve, PDF) \\
    0.65--0.7 & (nings, setAttribute) & (isen, issenschaft) & (ouv, schluss) \\
    0.7--0.75 & (ru, occup) & (result, utzt) & (aka, rola) \\
    0.75--0.8 & (cock, eland) & (hib, heast) & (prepare, Once) \\
    0.8--0.85 & (relation, emptyset) & (reen, bmatrix) & (uliar, ienn) \\
    0.85--0.9 & (Italie, urre) & (cement, cement) & (onna, onna) \\
    0.9--0.95 & (aped, aped) & (loster, loster) & (lict, lict) \\
    \bottomrule
    \end{tabular}
    \caption{Examples of token pairs for Llama2-7b.}
\label{tab:synthetic_pairs_llama2-7b}
\end{table}

\begin{table}[h!]
    \centering
    \begin{tabular}{llll}
    \textbf{Group} & \textbf{Pair 1} & \textbf{Pair 2} & \textbf{Pair 3} \\
    \toprule
    0.1--0.15 & (anity, OptionsMenu) & (attributes, Bitte) & (Commission, OptionsMenu) \\
    0.15--0.2 & (Transform, LEncoder) & (orgetown, DataProvider) & (download, SFML) \\
    0.2--0.25 & (types, DefaultCloseOperation) & (Defense, Autor) & (Cookies, Magn) \\
    0.25--0.3 & (VISION, Nut) & (ansi, Very) & (plants, addAll) \\
    0.3--0.35 & (COM, Refer) & (UFFIX, getResource) & (ModelError, fol) \\
    0.35--0.4 & (ifers, findViewById) & (Boundary, Foot) & (Quant, developers) \\
    0.4--0.45 & (arena, rak) & (Reuters, inflate) & (replacement, Detail) \\
    0.45--0.5 & (container, flu) & (webElementX, Sal) & (yet, multipart) \\
    0.5--0.55 & (Mission, Axis) & (down, remark) & (Rails, pictureBox) \\
    0.55--0.6 & (tier, messages) & (Mart, bold) & (analytics, Vis) \\
    0.6--0.65 & (grunt, pro) & (led, closest) & (matrix, stackpath) \\
    0.65--0.7 & (bye, byte) & (zeros, asctime) & (ending, protect) \\
    0.7--0.75 & (icated, ensure) & (afka, ref) & (flowers, caption) \\
    0.75--0.8 & (classed, classed) & (ERM, NASA) & (icana, mui) \\
    0.8--0.85 & (aggio, derive) & (tura, fillna) & (OnClick, endregion) \\
    0.85--0.9 & (ylko, echo) & (entario, cite) & (Avoid, inf) \\
    0.9--0.95 & (hotel, hotel) & (igrate, cite) & (recio, ulado) \\
    \bottomrule
    \end{tabular}
    \caption{Examples of token pairs for Llama3.1-8b.}
    \label{tab:synthetic_pairs_llama3.1-8b}
\end{table}

\begin{table}[h!]
    \centering
    \begin{tabular}{llll}
    \textbf{Group} & \textbf{Pair 1} & \textbf{Pair 2} & \textbf{Pair 3} \\
    \toprule
    0.1--0.15 & (Flash, ph) & (fake, stantiateViewController) & (Package, rid) \\
    0.15--0.2 & (taken, addPreferredGap) & (lambda, stantiateViewController) & (Nintendo, Fre) \\
    0.2--0.25 & (Chooser, meye) & (ENDED, queueReusable) & (Phil, NegativeButton) \\
    0.25--0.3 & (vehicles, uden) & (ideon, DllImport) & (Inverse, Wel) \\
    0.3--0.35 & (pressure, VertexAttrib) & (ForeignKey, gesch) & (those, CLLocation) \\
    0.35--0.4 & (Deferred, textAlign) & (sharp, SetBranchAddress) & (DEST, British) \\
    0.4--0.45 & (Across, Cas) & (Career, ud) & (Andre, Cod) \\
    0.45--0.5 & (oldemort, NumberFormatException) & (Binary, IMITIVE) & (indexes, BitConverter) \\
    0.5--0.55 & (represented, toContain) & (Chinese, THIS) & (jk, flatMap) \\
    0.55--0.6 & (Bon, Bon) & (Already, dbc) & (iston, Sub) \\
    0.6--0.65 & (Exam, let) & (Projectile, iores) & (odie, objectManager) \\
    0.65--0.7 & (Americans, illes) & (diff, stderr) & (Consulta, ificantly) \\
    0.7--0.75 & (LinkedIn, Prime) & (doctrine, iale) & (Space, vore) \\
    0.75--0.8 & (Get, Get) & (bomb, bomb) & (stripe, rightarrow) \\
    0.8--0.85 & (identifier, identifier) & (roller, roller) & (dimension, Intialized) \\
    0.85--0.9 & (Width, Width) & (Kitchen, Kitchen) & (landers, landers) \\
    0.9--0.95 & (Iterator, Iterator) & (balanced, balanced) & (pricing, pricing) \\
    \bottomrule
    \end{tabular}
    \caption{Examples of token pairs for Llama3.2-1b.}
    \label{tab:synthetic_pairs_llama3.2-1b}
\end{table}

\begin{table}[h!]
    \centering
    \begin{tabular}{llll}
    
    \textbf{Group} & \textbf{Pair 1} & \textbf{Pair 2} & \textbf{Pair 3} \\
    \toprule
    0.1--0.15 & (Great, getKey) & (Warnings, toEqual) & (Direct, Ser) \\
    0.15--0.2 & (Water, Ref) & (Pop, CREATE) & (sole, NET) \\
    0.2--0.25 & (children, ischen) & (hentic, redirect) & (TEMP, operatorname) \\
    0.25--0.3 & (username, por) & (Permission, LECT) & (submit, riev) \\
    0.3--0.35 & (JSON, optim) & (objects, junit) & (cat, contr) \\
    0.35--0.4 & (good, resolve) & (sam, partition) & (glas, forEach) \\
    0.4--0.45 & (Bank, Thank) & (append, stri) & (Interval, Mic) \\
    0.45--0.5 & (One, One) & (CEPT, Accept) & (ervices, Reference) \\
    0.5--0.55 & (Must, Must) & (Temp, Temp) & (foo, hbar) \\
    0.55--0.6 & (testing, testing) & (Input, Selector) & (PATH, THE) \\
    0.6--0.65 & (size, size) & (friend, mary) & (LowerCase, pathy) \\
    0.65--0.7 & (ebook, ebook) & (itzer, sender) & (LIB, LIB) \\
    0.7--0.75 & (century, century) & (ted, ted) & (JSON, ensuremath) \\
    0.75--0.8 & (Azure, Azure) & (aws, aws) & (ton, ton) \\
    0.8--0.85 & (getInt, getInt) & (uff le, ible) & (jem, jem) \\
    0.85--0.9 & (backup, backup) & (strlen, strlen) & (Width, Width) \\
    0.9--0.95 & (NonNull, NonNull) & (jed, jed) & (urd, urd) \\
    \bottomrule
    \end{tabular}
    \caption{Examples of token pairs for Mistral-7b.}
    \label{tab:synthetic_pairs_mistral-7b}
\end{table}

\begin{table}[h!]
    \centering
    \begin{tabular}{llll}
        \textbf{Group} & \textbf{Pair 1} & \textbf{Pair 2} & \textbf{Pair 3} \\
        \toprule
        0.1--0.15 & (Wednesday, Ngh) & (right, NSMutable) & (Pear, FILES) \\
        0.15--0.2 & (particle, Nej) & (easy, unc) & (Trader, multip) \\
        0.2--0.25 & (berry, Incre) & (Drawer, requ) & (OPTIONS, Fil) \\
        0.25--0.3 & (Life, bef) & (Reward, coc) & (Pages, Jer) \\
        0.3--0.35 & (your, enc) & (Thickness, atab) & (widgets, ilerek) \\
        0.35--0.4 & (borrow, empre) & (Restart, hatt) & (Crypto, orm) \\
        0.4--0.45 & (bul, dney) & (Creates, olumn) & (bout, OrNull) \\
        0.45--0.5 & (Hola, vert) & (ops, olare) & (Companies, SuppressWarnings) \\
        0.5--0.55 & (Fall, comm) & (Transient, bold) & (affected, big) \\
        0.55--0.6 & (Informe, sys) & (high, big) & (anna, badge) \\
        0.6--0.65 & (thought, Tags) & (Experiment, operator) & (sure, isset) \\
        0.65--0.7 & (yellow, center) & (empty, tiny) & (creen, rather) \\
        0.7--0.75 & (answers, prompt) & (Seats, Seats) & (Ryan, Ryan) \\
        0.75--0.8 & (sets, sets) & (ordinal, ordinal) & (conte, conte) \\
        0.8--0.85 & (telegram, telegram) & (Israeli, Israeli) & (fontsize, fontsize) \\
        0.85--0.9 & (country, country) & (pokemon, pokemon) & (gmail, gmail) \\
        0.9--0.95 & (PlainText, PlainText) & (bbc, bbc) & (jquery, jquery) \\
        \bottomrule
    \end{tabular}
    \caption{Examples of token pairs for Llama3.2-3b.}
    \label{tab:human_pairs_llama3.2-3b}
\end{table}

\begin{table}[h!]
    \centering
    \begin{tabular}{llll}
        \textbf{Group} & \textbf{Pair 1} & \textbf{Pair 2} & \textbf{Pair 3} \\
        \toprule
        0.1--0.15 & (logged, Zapraszamy) & (pharmacy, SuspendLayout) & (Closing, CODES) \\
        0.15--0.2 & (al, population) & (Readers, charging) & (brink, setObjectName) \\
        0.2--0.25 & (Prev, Findings) & (Knowledge, Whip) & (Detalle, WriteTagHelper) \\
        0.25--0.3 & (Colin, Lauren) & (favor, Aunt) & (Wikimedia, mechanical) \\
        0.3--0.35 & (few, trust) & (networking, StructEnd) & (luxe, Williams) \\
        0.35--0.4 & (Attribute, Angels) & (Hotline, Bought) & (ScrollView, archiviato) \\
        0.4--0.45 & (Mila, conductor) & (cian, river) & (zhen, Really) \\
        0.45--0.5 & (Holly, Nicole) & (Runners, Anybody) & (politics, Shown) \\
        0.5--0.55 & (Rejection, Accreditation) & (association, assembler) & (voiture, Document) \\
        0.55--0.6 & (developing, specifically) & (nvidia, codegen) & (pressing, scribers) \\
        0.6--0.65 & (assistance, expliquer) & (METHOD, CARD) & (MouseMove, cooperation) \\
        0.65--0.7 & (covariance, collection) & (markup, Upgrade) & (monger, metrist) \\
        0.7--0.75 & (COMPLEX, TRUST) & (ExecuteReader, MemoryWarning) & (dima, dima) \\
        0.75--0.8 & (listBox, errorMessage) & (Commit, Commit) & (Flesh, Flesh) \\
        0.8--0.85 & (ModelAndView, Element) & (componentWill, componentWill) & (navigateTo, navigateTo) \\
        0.85--0.9 & (tapete, felpa) & (ByteString, Interface) & (mapreduce, mapreduce) \\
        0.9--0.95 & (getSelection, getSelection) & (Salmo, Salmo) & (ido, ido) \\
        \bottomrule
    \end{tabular}
    \caption{Examples of token pairs for Gemma2-9b.}
\label{tab:human_pairs_gemma2-9b}
\end{table}

\clearpage

\subsection{Accuracy dynamics per model}
\label{appendix:subsec_acc_dyn_per_model}

\begin{figure*}[!h]
    \centering
    \begin{subfigure}{0.48\textwidth}
        \centering
        \includegraphics[width=\linewidth]{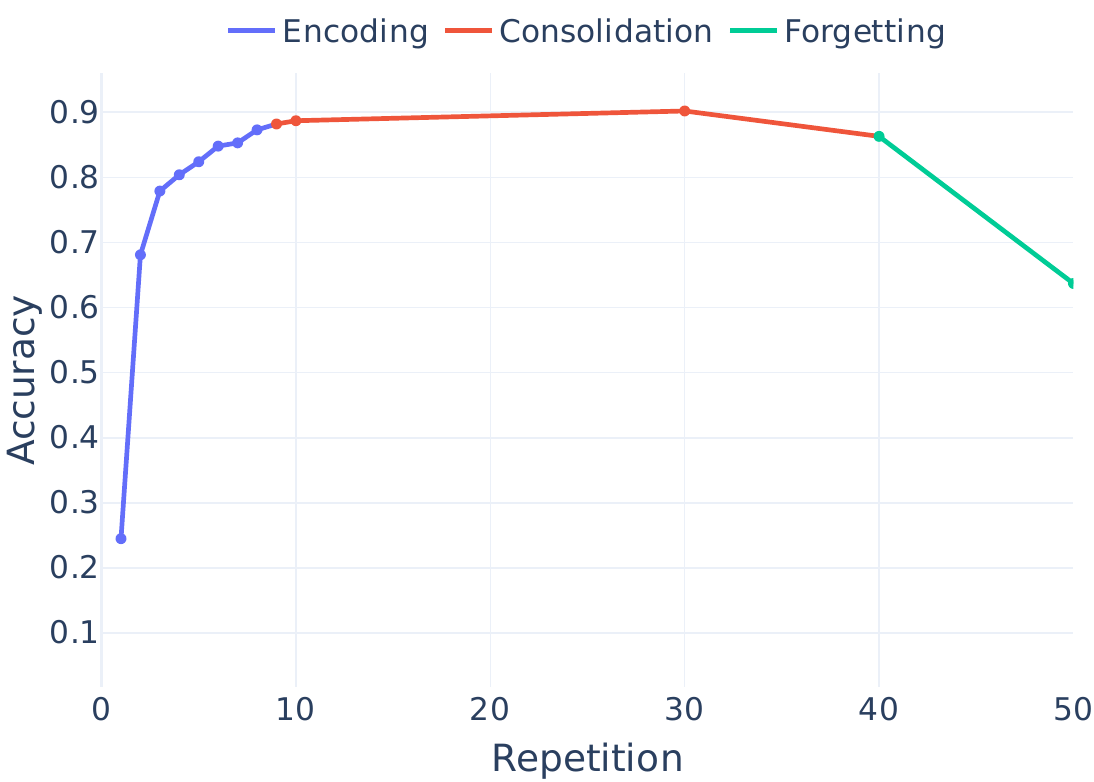}
        \caption{Llama2-7b.}
    \end{subfigure}
    \hfill
    \begin{subfigure}{0.48\textwidth}
        \centering
        \includegraphics[width=\linewidth]{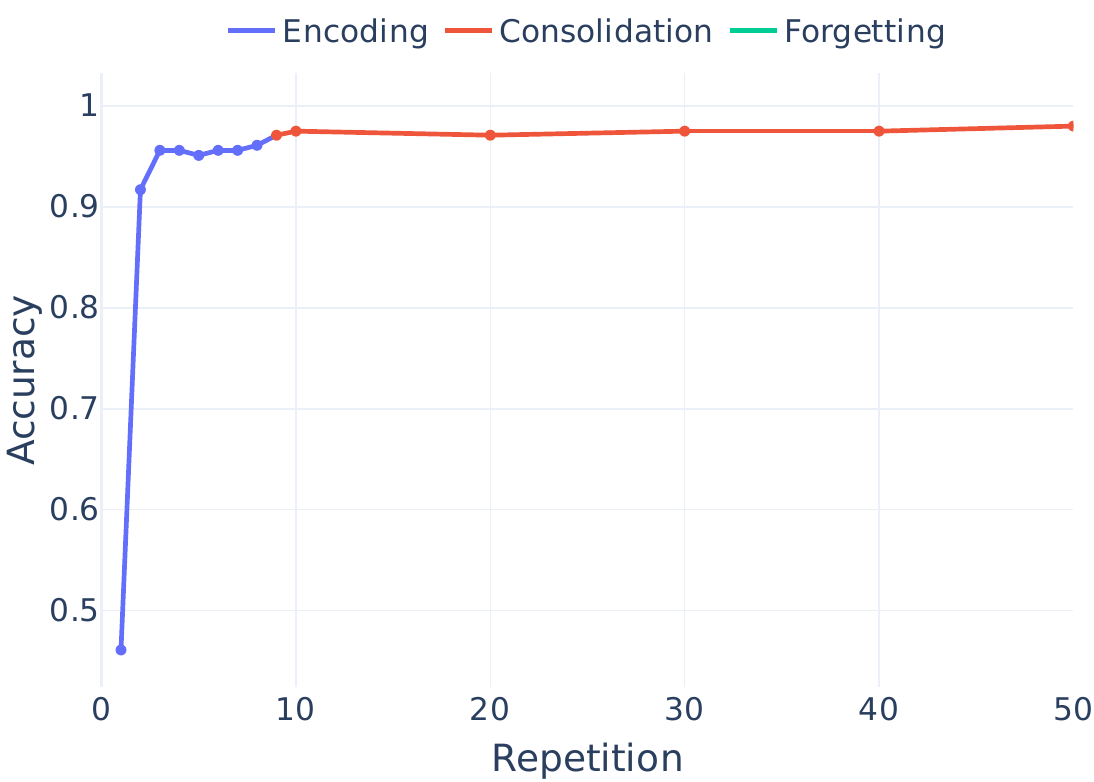}
        \caption{Mistral-7.}
    \end{subfigure}

    \begin{subfigure}{0.48\textwidth}
        \centering
        \includegraphics[width=\linewidth]{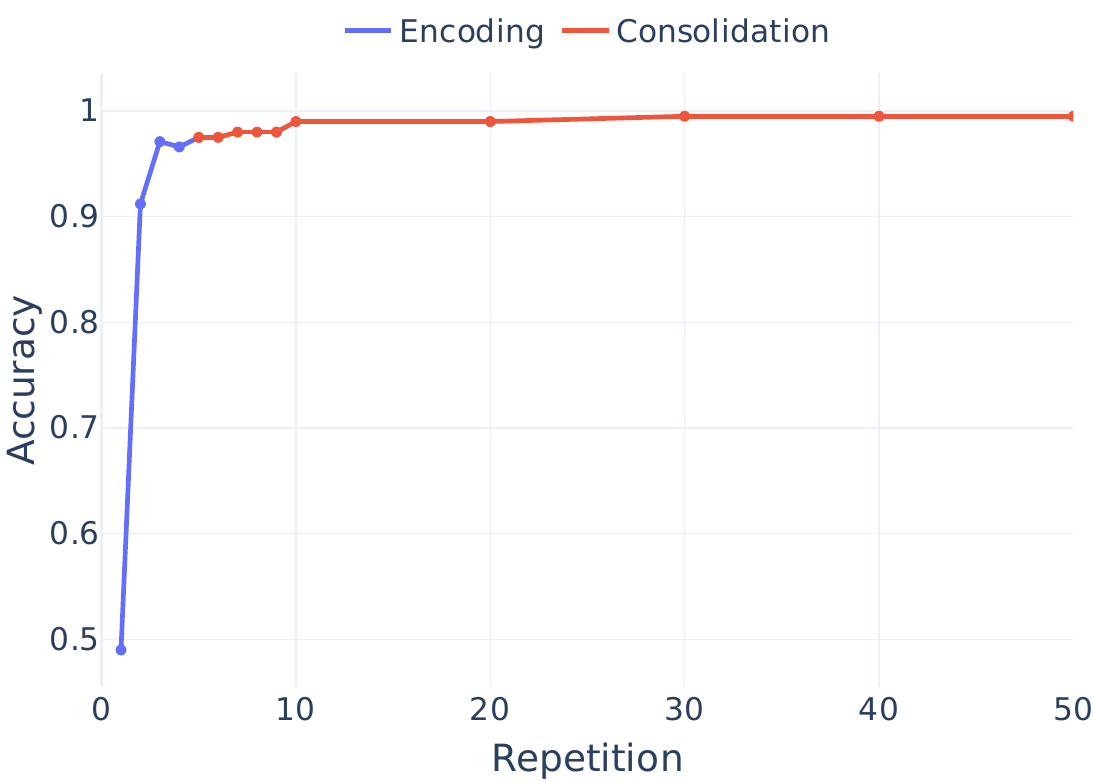}
        \caption{Llama3.1-8b.}
    \end{subfigure}
    \hfill
    \begin{subfigure}{0.48\textwidth}
        \centering
        \includegraphics[width=\linewidth]{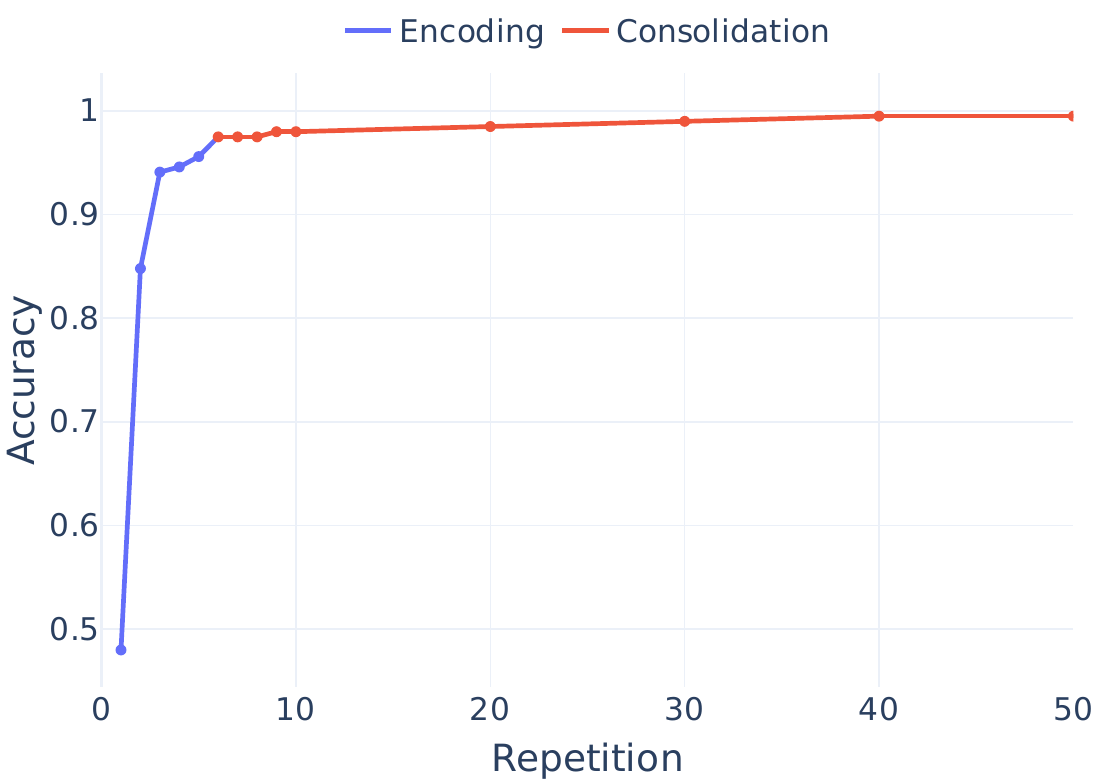}
        \caption{Llama3.2-1b.}
    \end{subfigure}

    \begin{subfigure}{0.48\textwidth}
        \centering
        \includegraphics[width=\linewidth]{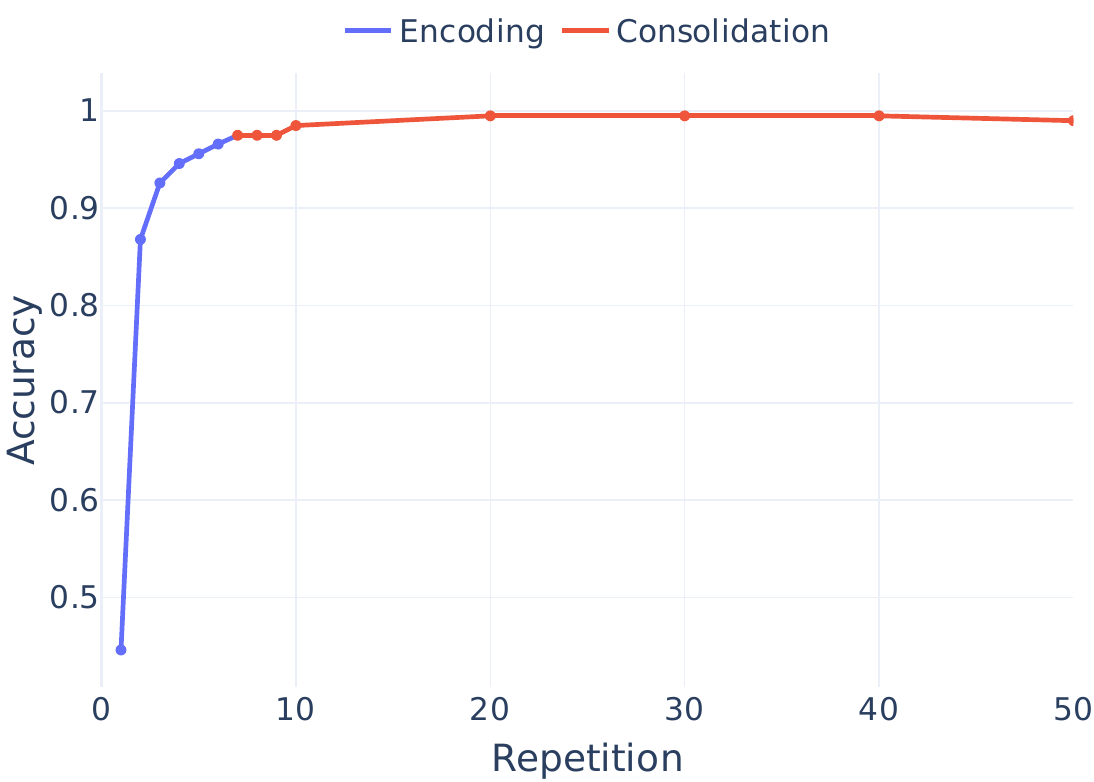}
        \caption{Llama3.2-3b.}
    \end{subfigure}
    \hfill
    \begin{subfigure}{0.48\textwidth}
        \centering
        \includegraphics[width=\linewidth]{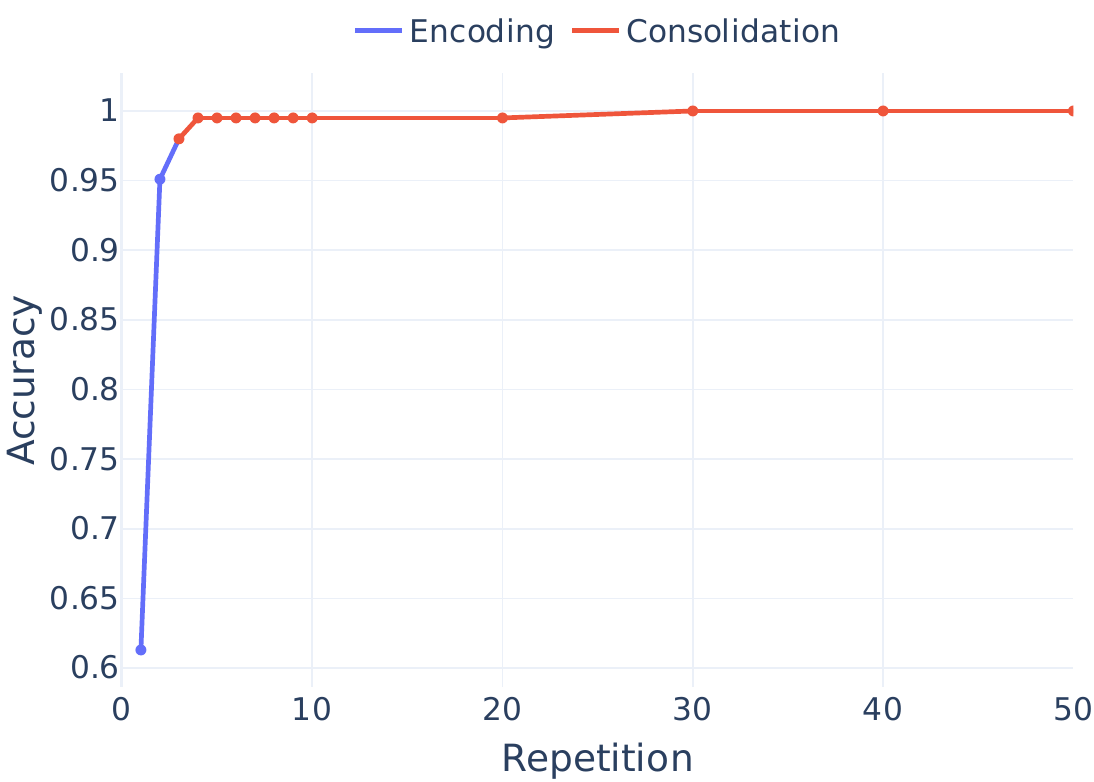}
        \caption{Gemma2-9b.}
    \end{subfigure}
    \caption{
    Accuracy over repetitions for all model, shown up to $50$ repetitions.
    }
    \label{fig:acc_dyn_per_model_by_rep_upto100}
\end{figure*}

In the main text (Figure~\ref{fig:acc}), we visualize how model accuracy evolves across different training stages by segmenting the process into three distinct phases: Encoding, Consolidation, and Forgetting. 
Since models may undergo a different number of repetitions in each phase, we normalized the $x$-axis by mapping each phase to a fixed interval.
This temporal alignment enables meaningful comparison of performance trajectories across models on a shared timeline. In Table~\ref{tab:appendix_phase_rep_per_model}, we provide details on when the learning phase transitions occur, and show the performance of each model at those transitions. Figure~\ref{fig:acc_dyn_per_model_by_rep_upto100} shows the accuracy dynamics across repetitions for all models, up to $50$ repetitions.
We can observe that each model has a slight different learning dynamic.

\begin{table}[H]
    \centering
    \caption{
    Model performance (i.e., accuracy on the associative task) across learning phases. For each model, we report the accuracy and repetition: at the end of the Encoding $\rightarrow$ Consolidation phase, at the maximum accuracy achieved during Consolidation, and at the end of Consolidation $\rightarrow$ Forgetting phase when applicable.
    }
    \vspace{1em}
    \begin{tabular}{lccc}
        Model & Encoding $\rightarrow$ Consolidation & Max. Accuracy & Consolidation $\rightarrow$ Forgetting \\
        \toprule
        Gemma2-9b & 0.98 ($r=3$) & 1.0 
        ($r=30$) & - \\
        Llama3.2-1b & 0.96 ($r=5$) & 1.0 
        ($r=150$) & - \\
        Llama3.2-3b & 0.97 ($r=6$) & 1.0 
        ($r=150$) & - \\
        Llama3.1-8b & 0.97 ($r=4$) & 1.0
        ($r=100$) & - \\
        Llama2-7b & 0.87 ($r=8$) & 0.9
        ($r=30$) & 0.86 ($r=40$) \\
        Mistral-7b & 0.96 ($r=8$) & 1.0
        ($r=600$) & 0.83 ($r=3$k) \\
        \bottomrule
    \end{tabular}
    \label{tab:appendix_phase_rep_per_model}
\end{table}

\clearpage

\subsection{Representation dynamics per model}
\label{appendix:subsec_repr_dyn_per_model}

\begin{figure*}[!h]
    \centering
    \begin{subfigure}{0.48\textwidth}
        \centering
        \includegraphics[width=\linewidth]{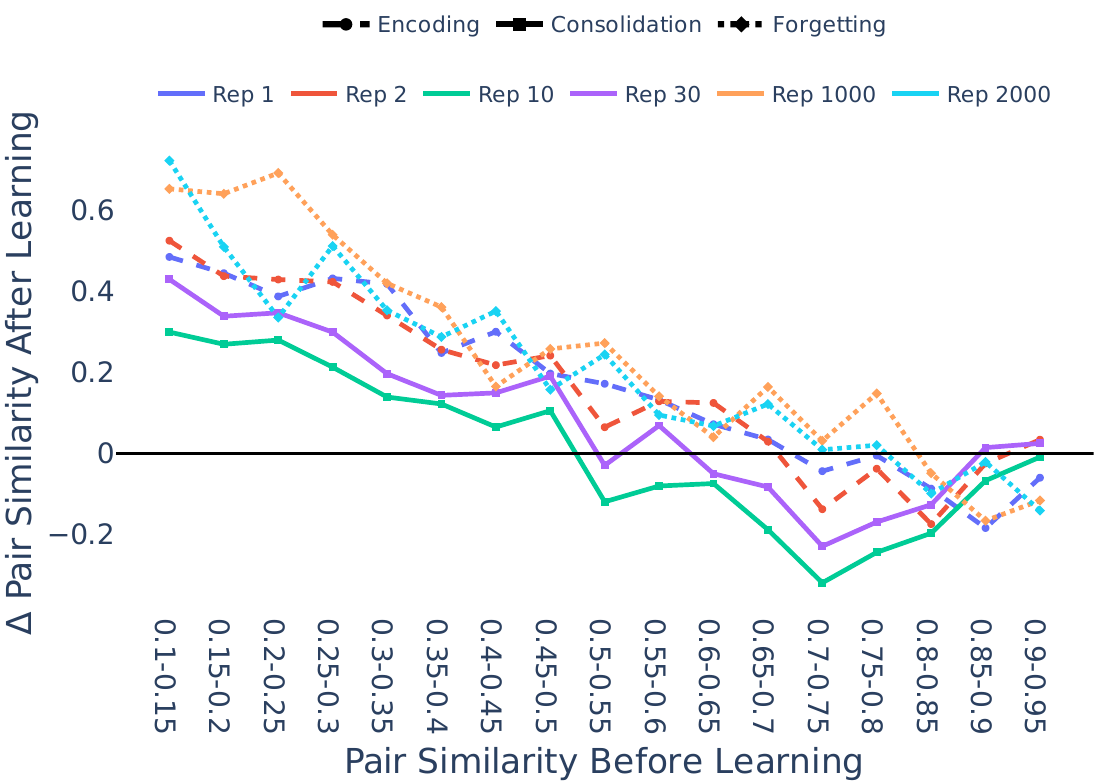}
        \caption{Llama2-7b.}
    \end{subfigure}
    \hfill
    \begin{subfigure}{0.48\textwidth}
        \centering
        \includegraphics[width=\linewidth]{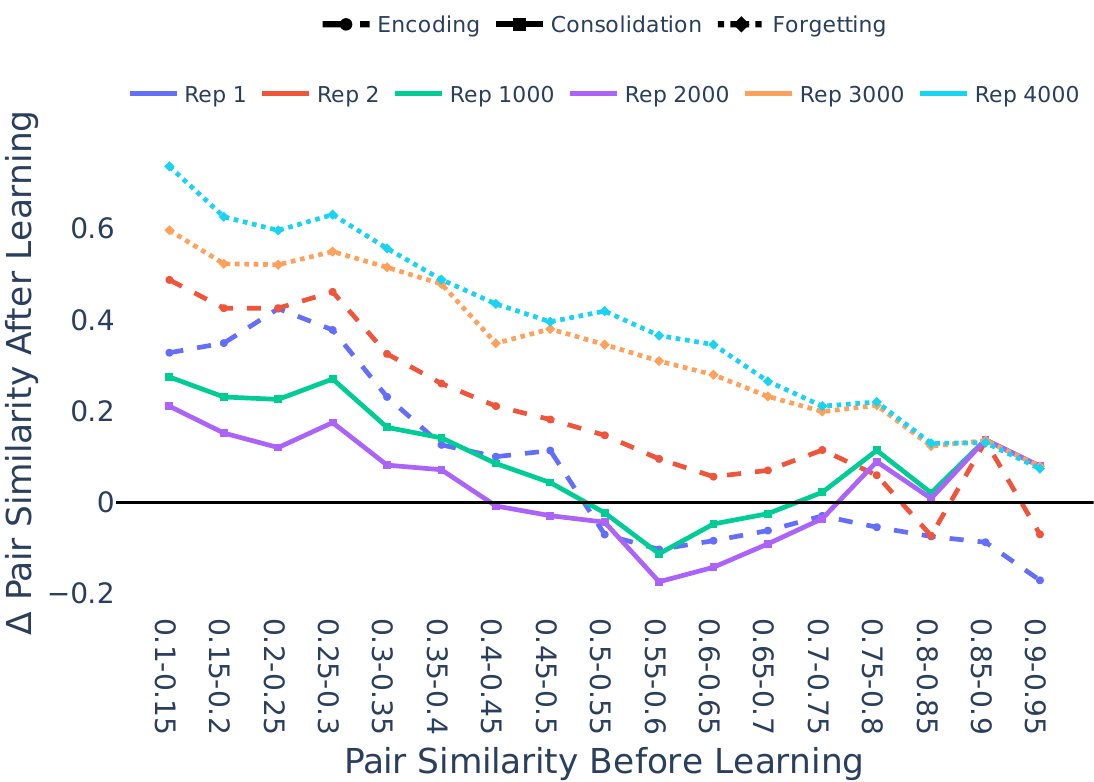}
        \caption{Mistral-7.}
    \end{subfigure}

    \begin{subfigure}{0.48\textwidth}
        \centering
        \includegraphics[width=\linewidth]{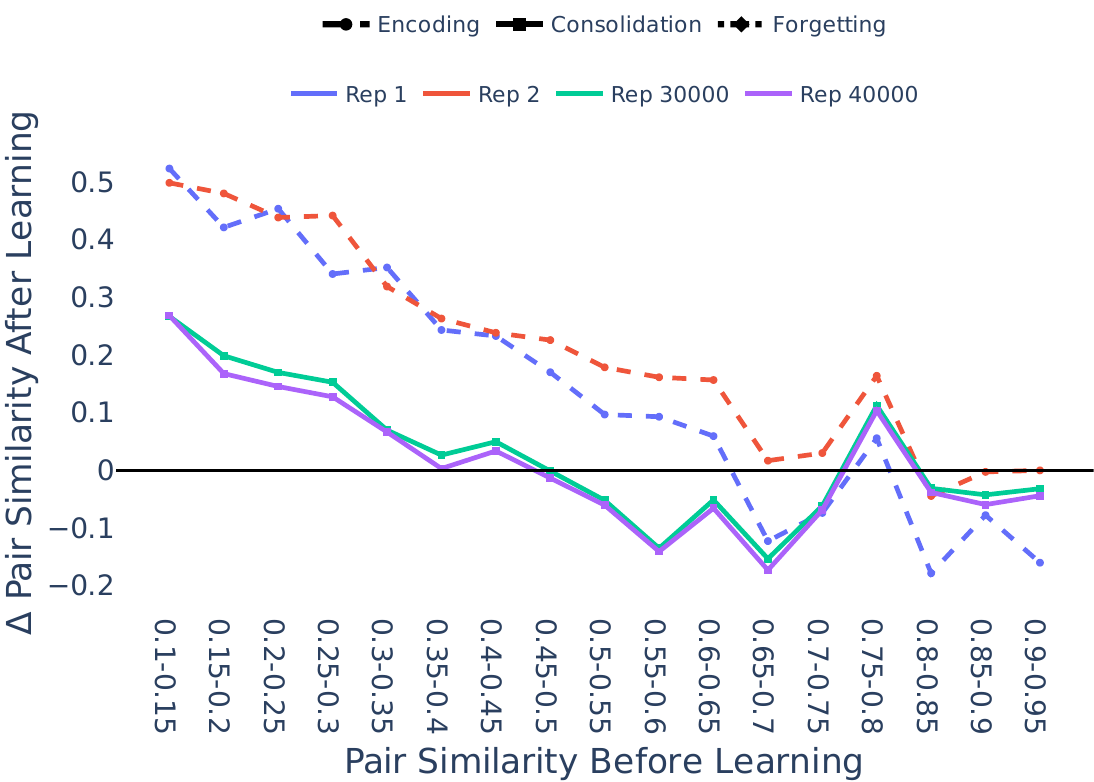}
        \caption{Llama3.1-8b.}
    \end{subfigure}
    \hfill
    \begin{subfigure}{0.48\textwidth}
        \centering
        \includegraphics[width=\linewidth]{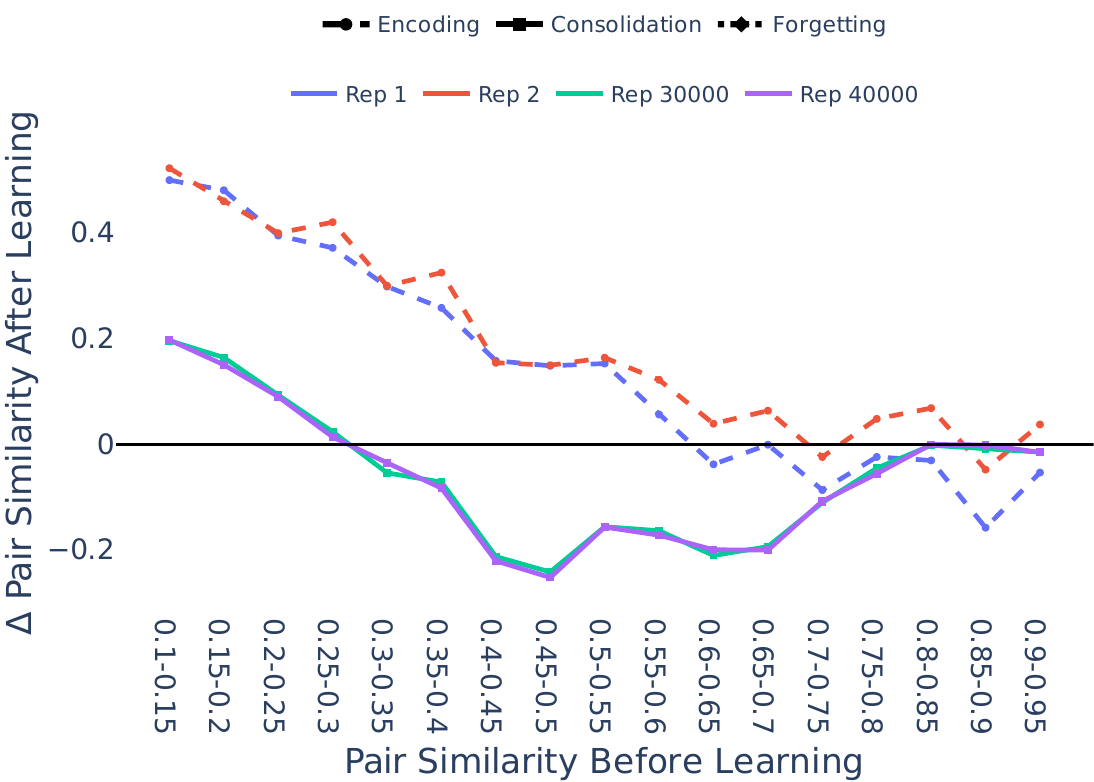}
        \caption{Llama3.2-3b.}
    \end{subfigure}

    \begin{subfigure}{0.48\textwidth}
        \centering
        \includegraphics[width=\linewidth]{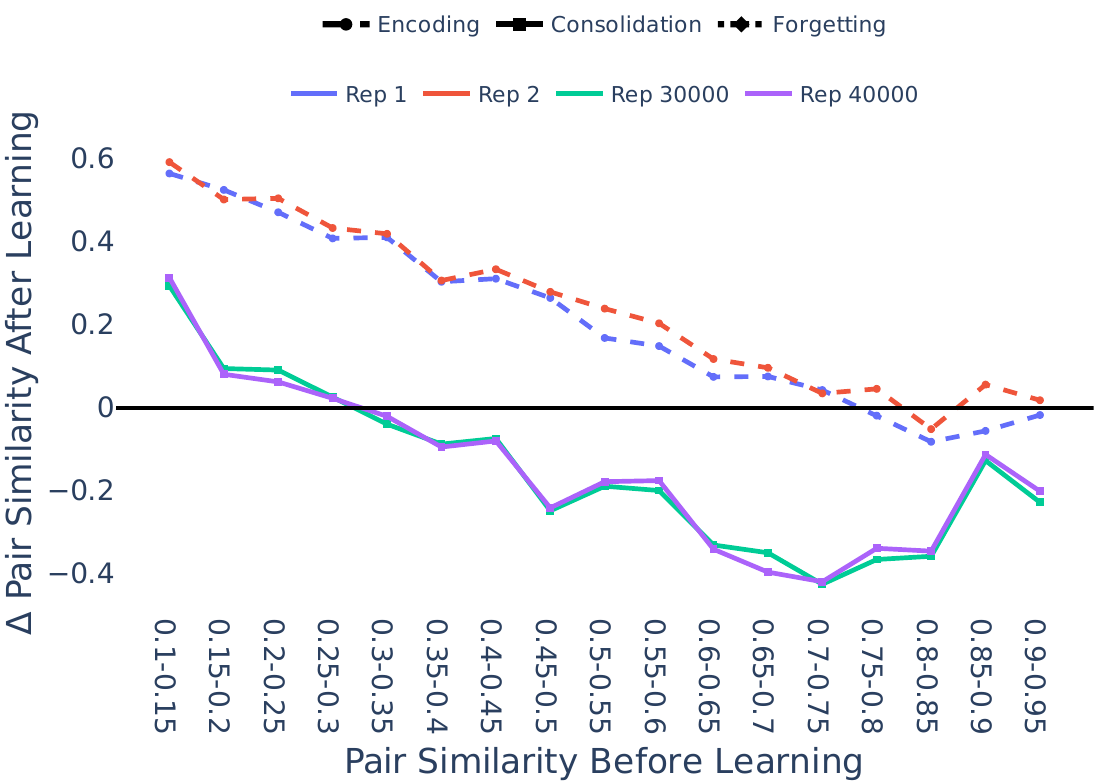}
        \caption{Llama3.2-1b.}
    \end{subfigure}
    \hfill
    \begin{subfigure}{0.48\textwidth}
        \centering
        \includegraphics[width=\linewidth]{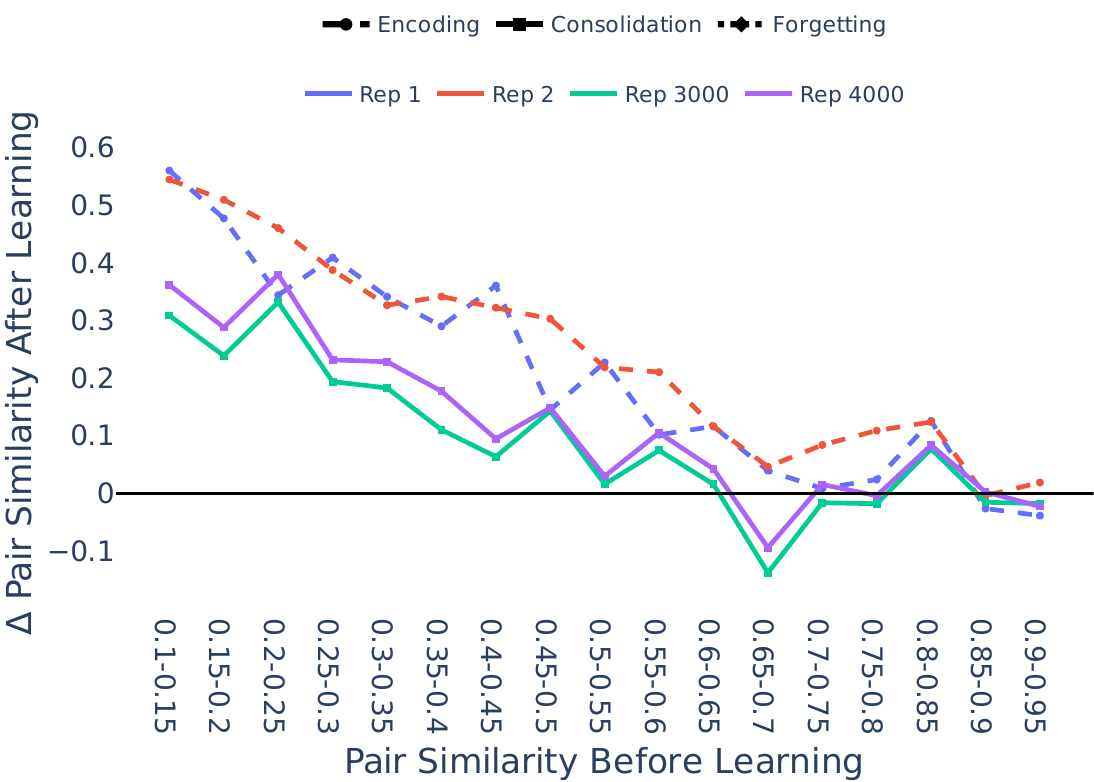}
        \caption{Gemma2-9b.}
    \end{subfigure}
    \caption{
    Representational changes across repetitions and their corresponding learning phase (one plot per model).
    To reduce an overly dense visualization, we display a subset of repetitions: for models with a forgetting phase, $2$ repetitions per phase were selected; for models without a forgetting phase, $3$ repetitions per phase were included.
    Across all models, we observe a non-monotonic trend aligned with NMPH during the consolidation phase.
    }
    \label{fig:repr_dyn_per_model_by_rep}
\end{figure*}

In the main text (Figure~\ref{fig:repr}), we present normalized trajectories of representational change across learning phases, allowing comparison across models.
Here, we provide the corresponding per-model plots (Figure~\ref{fig:repr_dyn_per_model_by_rep}), showing representational change across repetitions.
Across models, we observe a consistent non-monotonic trend during the Consolidation (straight line) phase.

\clearpage
\subsection{Potential factors in forgetting phase}
\label{appendix:subsec_forgetting_phase}

In the main text (Section~\ref{subsec:accuracy_pairsim}), we observed that two models--Llama2-7b and Mistral-7b--showed a forgetting phase, characterized by a drop in accuracy greater than $3\%$ relative to the average of the two preceding repetitions.
This behavior indicates the start of performance degradation. We speculate that the delayed forgetting observed in Mistral-7b may be influenced by its use of a sliding window attention (SWA) mechanism.

\begin{figure}[h!]
    \centering
    \begin{subfigure}{0.48\textwidth}
        \centering
        \includegraphics[width=\linewidth]{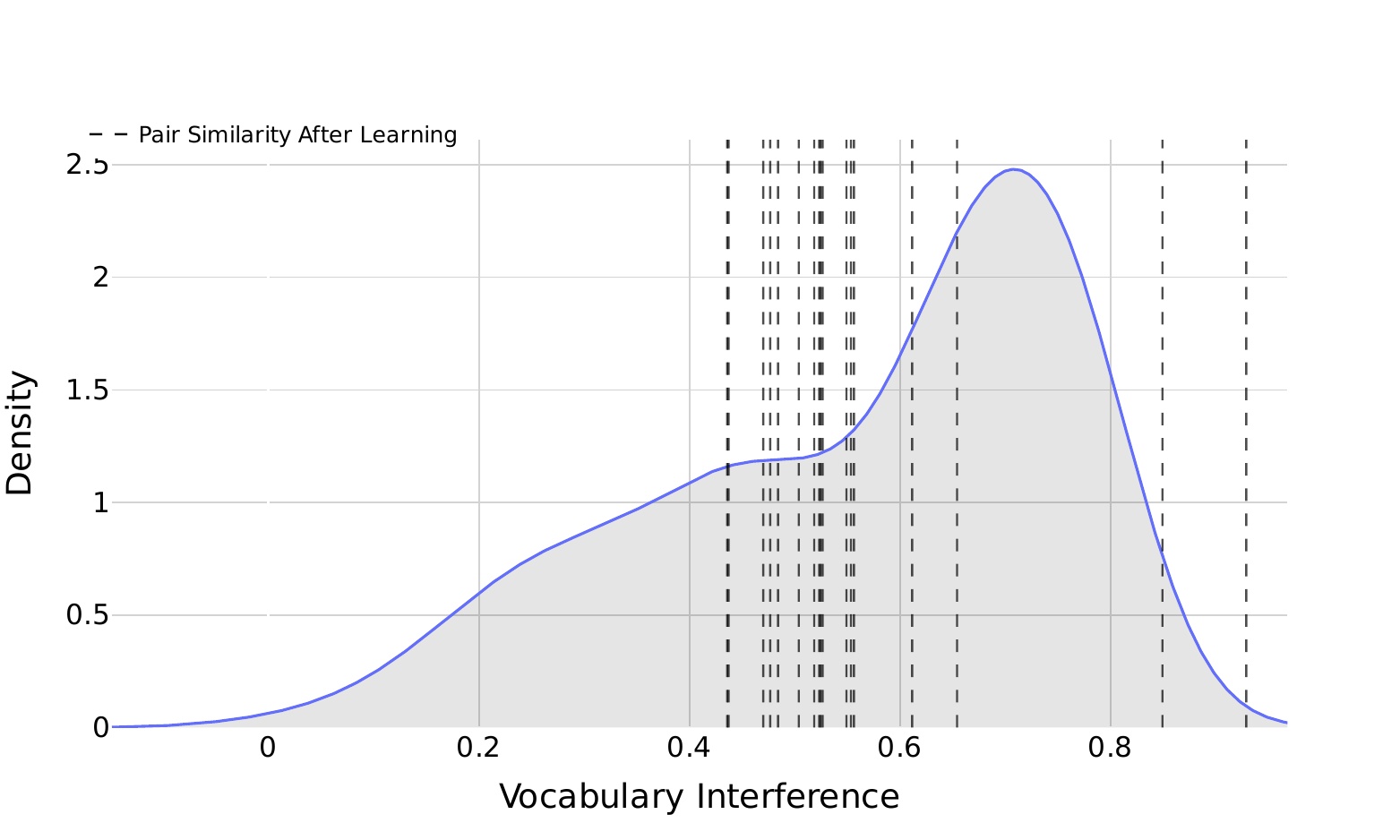}
        \caption{End of Consolidation}
        \label{fig:appendix_llama27b_vocabdistr_a}
    \end{subfigure}
    \hfill
    \begin{subfigure}{0.48\textwidth}
        \centering
        \includegraphics[width=\linewidth]{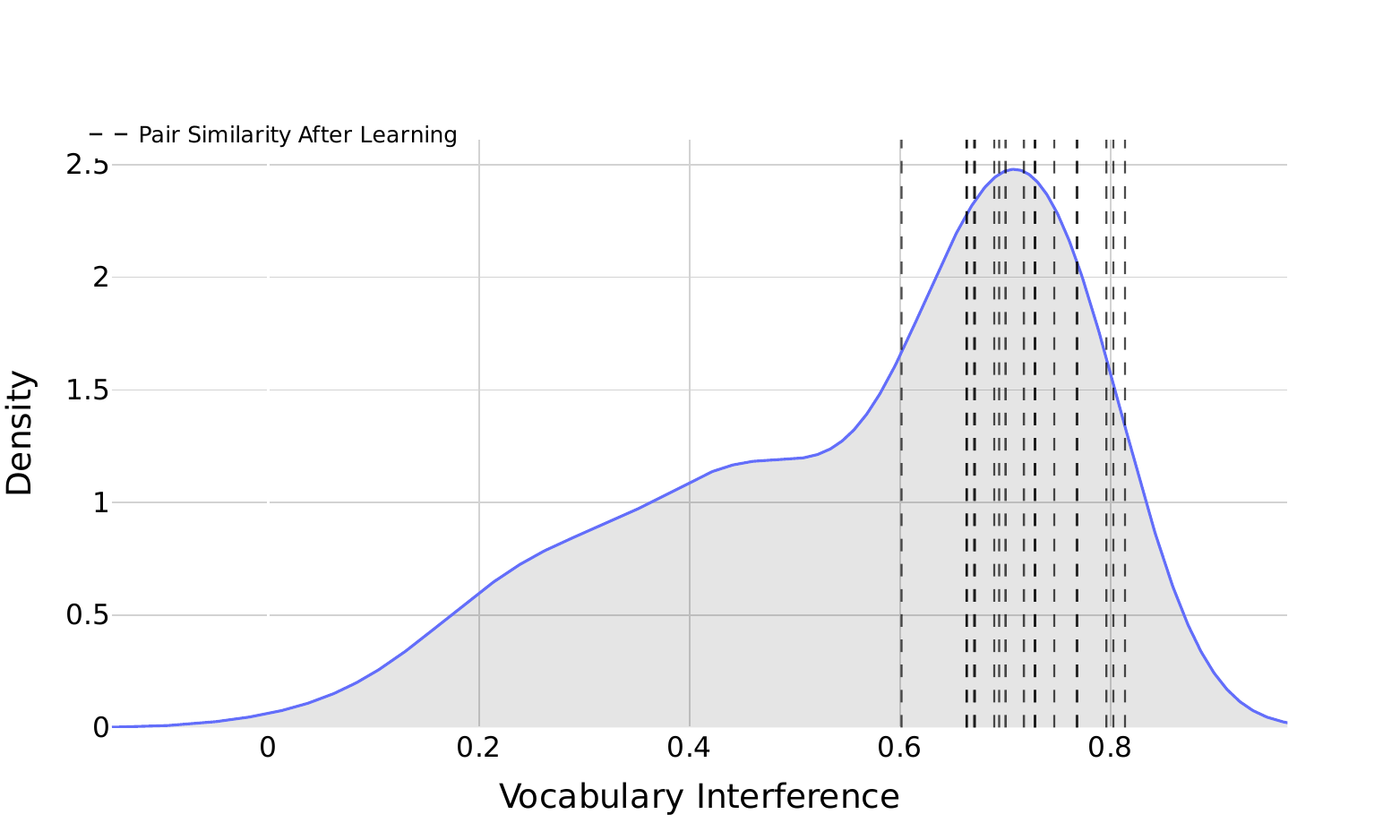}
        \caption{Beginning of Forgetting}
        \label{fig:appendix_llama27b_vocabdistr_b}
    \end{subfigure}
    \caption{
    Vocabulary interference distribution for Llama2-7b at (a) the end of the Consolidation phase ($r=30$), and (b) the start of the Forgetting phase ($r=40$).
    Vertical dashed lines indicate the average pair similarity after learning for each group.
    During Forgetting, a noticeable shift in pair similarity toward the peak of the interference distribution suggests increased competition, potentially contributing to the observed decline in performance.
    }
    \label{fig:appendix_vocabsim_distr_llama27}
\end{figure}

We performed an initial analysis of a possible---though speculative---factor that may have influenced the forgetting phase observed in the Llama2-7b model.
Figure~\ref{fig:appendix_vocabsim_distr_llama27} shows the distribution of vocabulary interference, where vertical lines show the average pair similarity after learning, per group.
The subfigures show the distribution for the last repetition of the Consolidation ($r=30$) and the first repetition of the Forgetting phases ($r=40$), respectively.
Notably, during Forgetting, token pairs shift toward the peak of the interference distribution. This suggests that Forgetting occurs when there is increased competition from similar vocabulary items, which could be impairing the model’s ability to maintain accurate associations.
This interpretation remains speculative, and future work could further investigate the causes of forgetting and their relationship to interference.

\subsection{Supplementary analyses of representational dynamics}



\begin{figure}[H]
    \centering
    \includegraphics[width=0.8\linewidth]{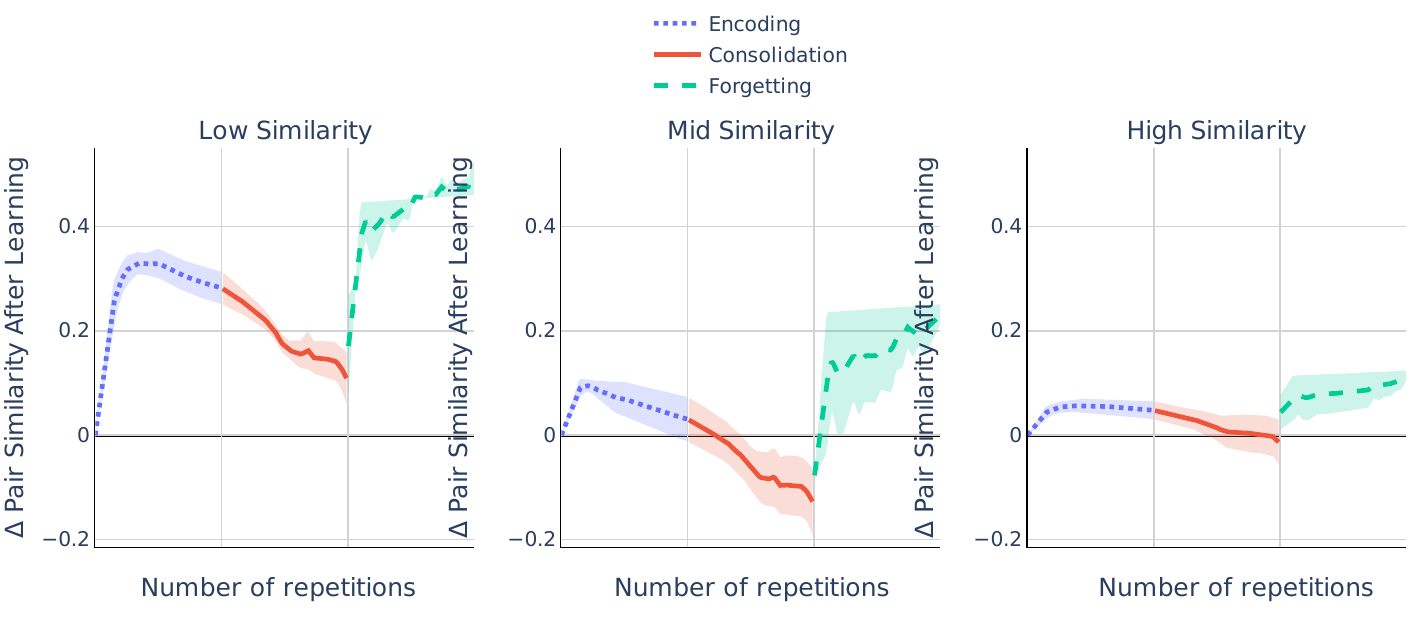}
    \caption{
    Representational change across learning phases (Encoding, Consolidation, Forgetting) for different pairwise similarity categories.
    Mid-similarity pairs were selected based on the groups that showed significant differentiation in our t-test analysis (Figure~\ref{fig:repr}). All groups with lower similarity scores were aggregated into the low-similarity category, and those with higher scores into the high-similarity category. Data is averaged across models. Shaded areas represent the standard error across models.
    }
    \label{fig:appendix_reprchange_percatsim_acrossphases}
\end{figure}

Figure~\ref{fig:appendix_reprchange_percatsim_acrossphases} shows the trajectory of representational change across learning phases separately for low, moderate, and high similarity groups.
The mid-similarity group includes only those pairs that exhibited significant differentiation in the t-test analysis from Section~\ref{subsec:reprchange_pairsim}.
Low- and high-similarity categories were defined by aggregating the remaining pairs based on their similarity scores.
The results reveal distinct dynamics across similarity regimes, although the overall shape of the changes remains consistent across similarity groups.
Low-similarity pairs exhibit a sharp increase in representational similarity during the initial repetitions of the Encoding phase, followed by a gradual decline throughout Consolidation.
In contrast, mid-similarity pairs show a more modest increase during Encoding but undergo a significant decrease during Consolidation, ultimately exhibiting strong differentiation.
High-similarity pairs remain relatively stable, with only a slight increase during Encoding and a minor reduction during Consolidation.
These trends are broadly consistent across models.

\subsection{Analysis for extended set}
\label{appendix:sec_analysis_extended_set}

We extended the main analysis to search for $100$ token pairs per similarity group, for both Llama2-7b and Llama3.2-1b.
The results reveal consistent patterns with those shown in Figure~\ref{fig:analysis} of the main paper.

\begin{figure*}[!h]
    \centering
    \begin{subfigure}{0.48\textwidth}
        \centering
        \includegraphics[width=\linewidth]{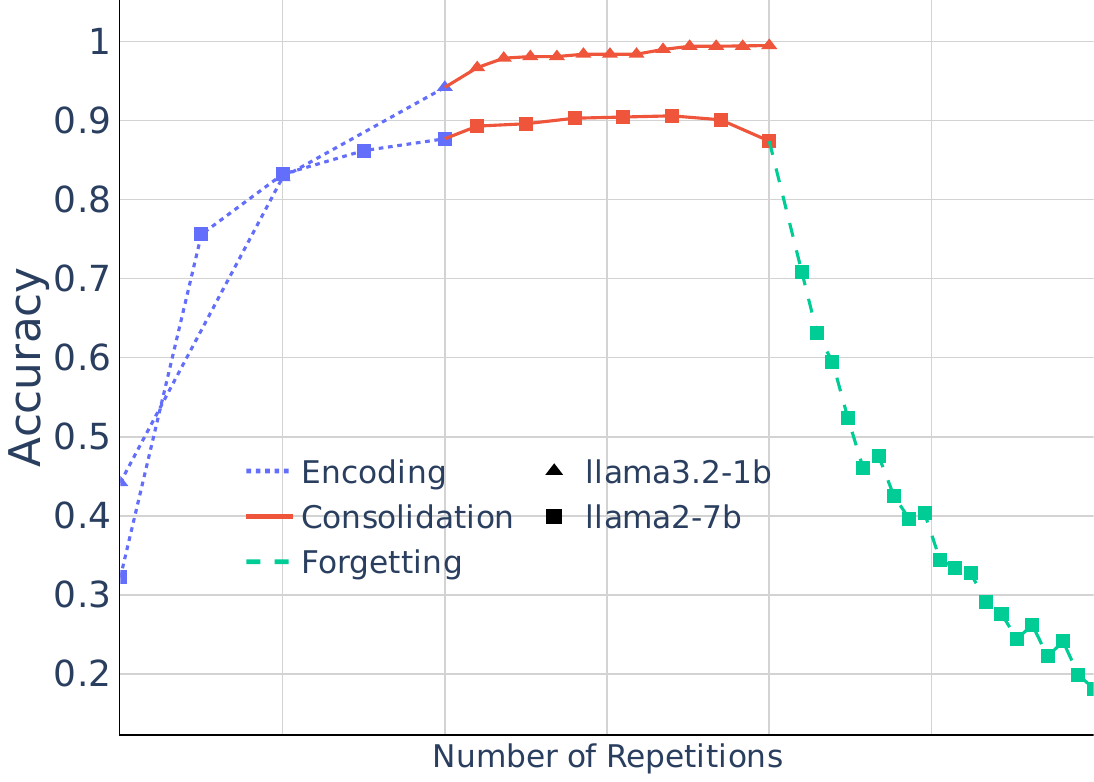}
        \caption{Accuracy across phases of learning.}
        \label{fig:acc_100extendedp}
    \end{subfigure}
    \hfill
    \begin{subfigure}{0.48\textwidth}
        \centering
        \includegraphics[width=\linewidth]{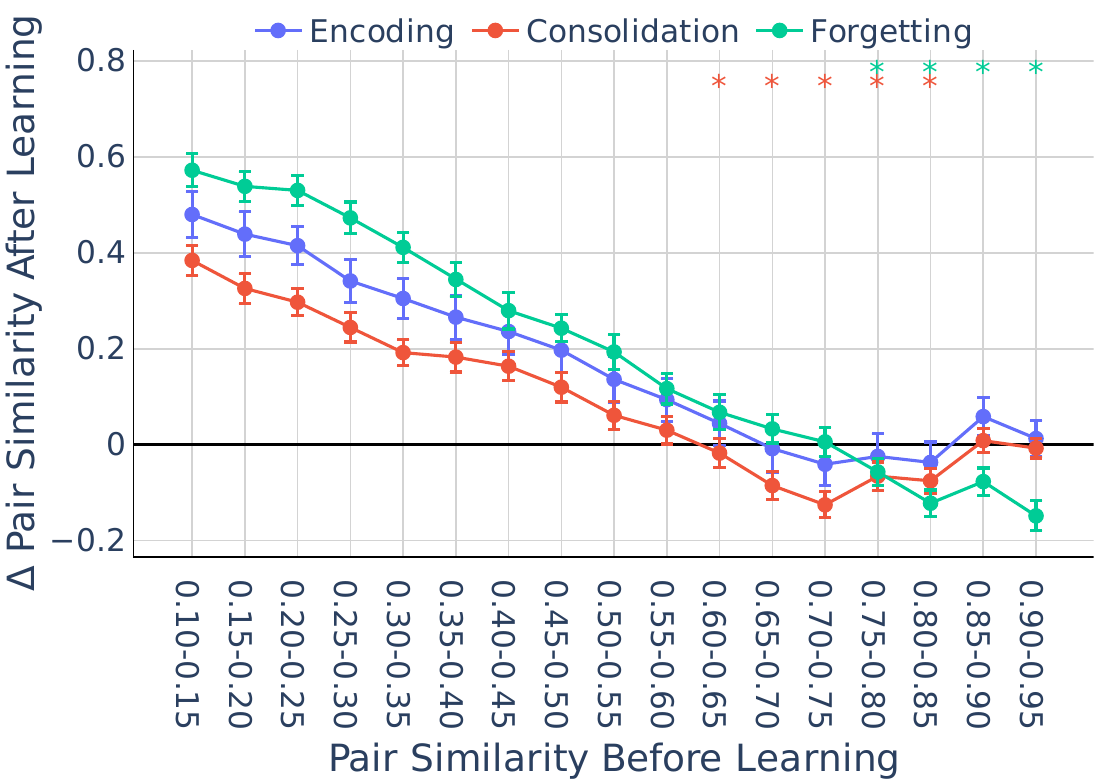}
        \caption{Representational change due to learning.}
        \label{fig:repr_100extendedp}
    \end{subfigure}
    \caption{
        Accuracy and representational changes during learning for an extended stimulus set comprising $100$ optimized token pairs in each of the $17$ similarity groups.
        (a) Models show three phases of learning: encoding, where accuracy steeply increases; consolidation, where accuracy stabilizes; and forgetting, where accuracy declines. The x-axis for each model is scaled by the length of its learning phase.
        (b) The U-shaped differentiation pattern, characteristic of the Non-Monotonic Plasticity Hypothesis, is observed only during consolidation (red).
        Asterisks ($\ast$) indicate groups that remain significant after Benjamini--Yekutieli correction for multiple comparisons across similarity groups and phases ($q<0.05$).
    }
    \label{fig:analysis_extendedp}
\end{figure*}
\clearpage
\section{Analysis for WordNet token pairs}
\label{appendix:sec_analysis_wordnet}

Our study intentionally selected token pairs selected for their pair similarity before learning, regardless of semantic meaning.
This design mirrors the use of synthetic stimuli in~\cite{wammes2022increasing}, which intentionally avoids meaningful real-world inputs and emphasizes the importance of sampling across the entire similarity spectrum--especially the mid-similarity range--to effectively test NMPH.
Because real-word tokens are unevenly distributed across this space, achieving precise control is otherwise difficult.
Accordingly, our primary aim in this work is not to study meaning, but to examine the structural dynamics of representational change in response to learning.

That said, in this section we briefly assess how representation dynamics evolve under more naturalistic conditions.
In our main analyses, we already filtered out tokens containing numbers, punctuation, or special characters.
Here, we further restricted token pairs to single-token WordNet words, which reduced the usable vocabulary to roughly $\approx2.4$k tokens out of $\approx28$k for Llama2-7b and $\approx4.8$k out of $\approx10$k for Llama3.2-1b.

\begin{figure*}[!h]
    \centering
    \includegraphics[width=0.48\linewidth]{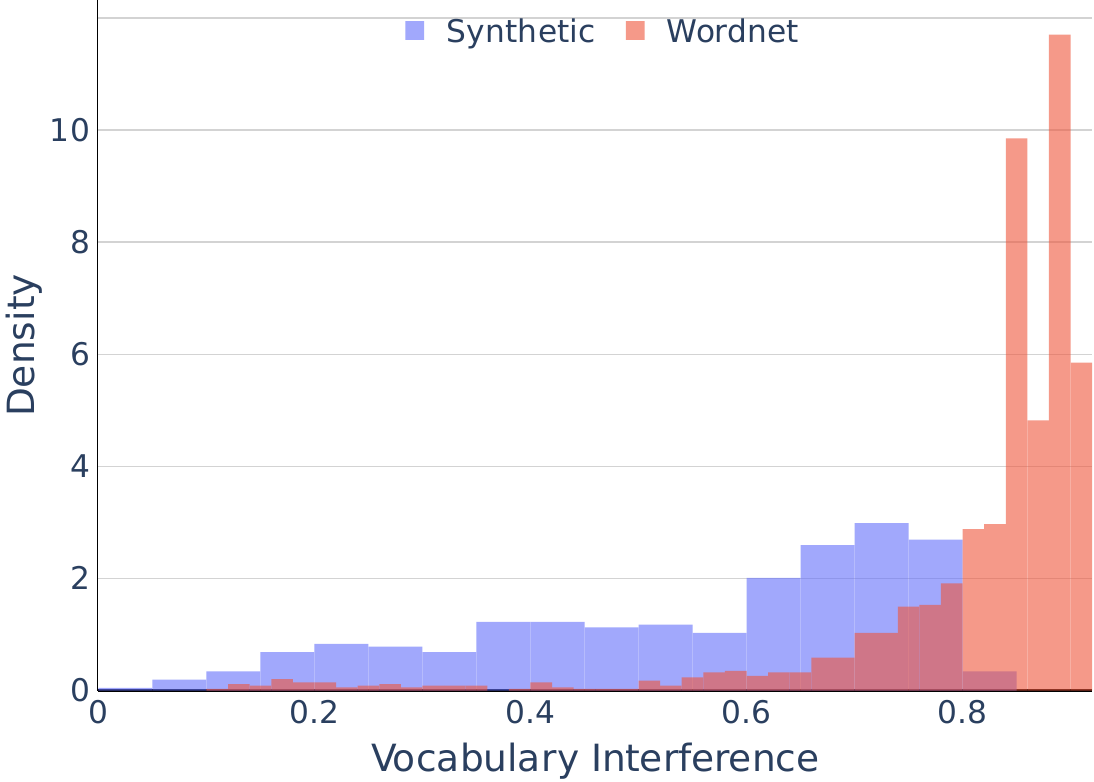}
    \caption{
        Distribution of vocabulary interference for previously-optimized synthetic token pairs versus WordNet token pairs.
        Original pairs (blue) span the full similarity spectrum, enabling controlled sampling across ranges, while WordNet pairs (red) cluster at high vocabulary interference values. 
        This skew highlights the difficulty of achieving balanced coverage with real-word tokens and motivates the use of optimized and more synthetic stimuli to test NMPH.
    }
    \label{fig:vocabinterf_distr_wordnet}
\end{figure*}

\begin{figure*}[!t]
    \centering
    \begin{subfigure}{0.48\textwidth}
        \centering
        \includegraphics[width=\linewidth]{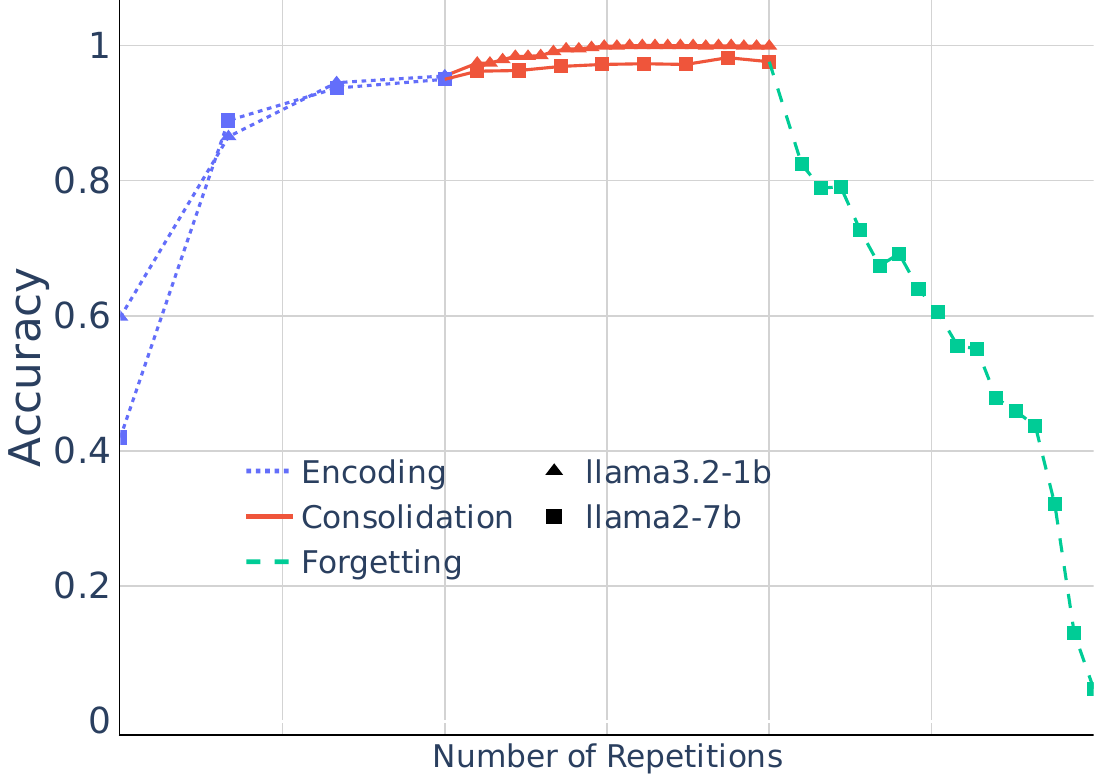}
        \caption{Accuracy across phases of learning.}
        \label{fig:acc_wornetp}
    \end{subfigure}
    \hfill
    \begin{subfigure}{0.48\textwidth}
        \centering
        \includegraphics[width=\linewidth]{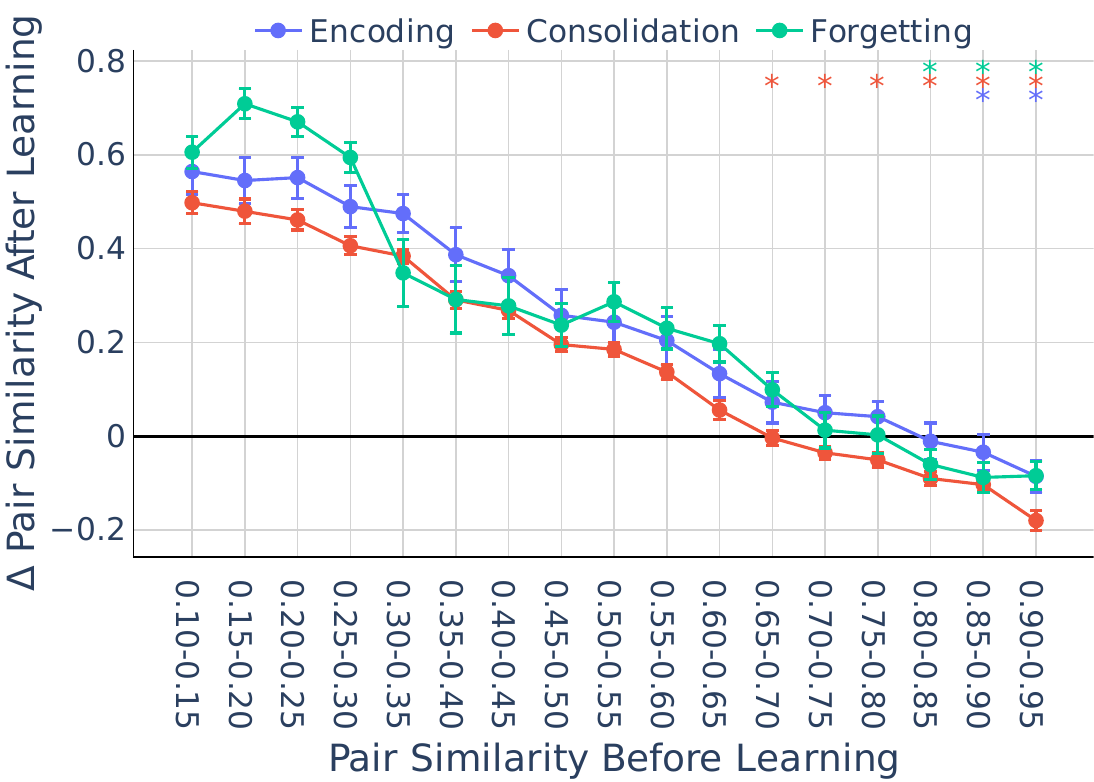}
        \caption{Representational change due to learning.}
        \label{fig:repr_wordnetp}
    \end{subfigure}
    \caption{
        Accuracy and representational changes during learning with a set of WordNet token pairs.
        (a) Across models, learning unfolds in three phases: Encoding, marked by a steep rise in accuracy; Consolidation, where accuracy stabilizes; and Forgetting, where accuracy declines. The x-axis for each model is scaled to the length of its learning phase.
        (b) In contrast to earlier results, the characteristic U-shaped differentiation pattern is diminished, giving way to a monotonically decreasing trend, consistent with the higher vocabulary interference observed among real-word token pairs.
        Asterisks ($\ast$) denote similarity groups that remain significant after Benjamini–Yekutieli correction for multiple comparisons across groups and phases ($q<0.05$).
    }
    \label{fig:analysis_wordnet}
\end{figure*}

We first examined how pairwise similarity and vocabulary interference were distributed within this constrained space, anticipating that the reduced vocabulary might bias pairs toward narrower interference ranges.
Indeed, sampled real-word token pairs show higher vocabulary interference than our synthetic token pairs (Figure \ref{fig:vocabinterf_distr_wordnet}), suggesting that they face stronger competition during prediction.
Our results (Figure~\ref{fig:analysis_wordnet}) confirmed this: similarity after learning decreased monotonically with respect to the similarity before learning, supporting the view that highly similar pairs are modulated by vocabulary interference.
Together, these findings suggest that global interference is a key factor modulating representational dynamics in naturalistic learning settings, and that NMPH emerges under specific conditions of global interference.

\clearpage
\subsection{WordNet token pair examples}
\label{appendix:subsec_wordnet_token_pair_examples}

\begin{table}[h!]
    \centering
    \begin{tabular}{l|lll}
    \textbf{Similarity Range} & \textbf{Pair 1} & \textbf{Pair 2} & \textbf{Pair 3} \\
    \hline
    0.1--0.15 & (mix, loaded) & (defined, standard) & (pub, any) \\
    0.15--0.2 & (identifier, astern) & (layout, eclipse) & (defined, slash) \\
    0.2--0.25 & (online, pop) & (suite, abb) & (pub, format) \\
    0.25--0.3 & (absolute, pus) & (round, pa) & (annotation, hum) \\
    0.3--0.35 & (series, math) & (black, roc) & (gas, bat) \\
    0.35--0.4 & (spec, cock) & (information, leg) & (argument, lear) \\
    0.4--0.45 & (contra, architecture) & (dictionary, ike) & (rooms, ho) \\
    0.45--0.5 & (gen, nil) & (factory, acre) & (shadow, nih) \\
    0.5--0.55 & (gen, raise) & (time, eb) & (zero, iga) \\
    0.55--0.6 & (dale, person) & (dawn, esp) & (irs, ante) \\
    0.6--0.65 & (any, essen) & (final, mission) & (gi, dim) \\
    0.65--0.7 & (cap, bind) & (mus, skim) & (dd, safe) \\
    0.7--0.75 & (bye, anas) & (izar, through) & (lined, click) \\
    0.75--0.8 & (replace, stock) & (unction, week) & (execution, frame) \\
    0.8--0.85 & (geometry, list) & (locale, embed) & (partition, brand) \\
    0.85--0.9 & (opacity, fragment) & (render, inflate) & (analysis, section) \\
    0.9--0.95 & (gable, board) & (volution, ship) & (slider, simple) \\
    \bottomrule
    \end{tabular}
    \caption{Wordnet token pairs examples for llama2-7b.}
    \label{tab:wordnet_pairs_llama2-7b}
\end{table}

\begin{table}[h!]
    \centering
    \begin{tabular}{l|lll}
    \textbf{Similarity Range} & \textbf{Pair 1} & \textbf{Pair 2} & \textbf{Pair 3} \\
    \toprule
    0.10--0.15 & (tour, rather) & (elect, subscribe) & (speaker, hear) \\
    0.15--0.20 & (inherit, soon) & (phone, six) & (internal, town) \\
    0.20--0.25 & (access, version) & (roman, doll) & (artist, then) \\
    0.25--0.30 & (import, traffic) & (flat, sin) & (license, raj) \\
    0.30--0.35 & (creator, solution) & (department, ne) & (package, ghost) \\
    0.35--0.40 & (use, company) & (declare, dead) & (linux, gu) \\
    0.40--0.45 & (sign, oracle) & (google, cro) & (district, bone) \\
    0.45--0.50 & (code, rabbit) & (sign, testing) & (public, edd) \\
    0.50--0.55 & (code, extended) & (code, radius) & (code, folder) \\
    0.55--0.60 & (far, match) & (sea, ledger) & (type, memory) \\
    0.60--0.65 & (wide, resize) & (sea, timing) & (dot, sector) \\
    0.65--0.70 & (express, window) & (sky, connection) & (mind, league) \\
    0.70--0.75 & (identifier, technical) & (mind, oracle) & (earth, corner) \\
    0.75--0.80 & (dream, burst) & (pixel, circle) & (earth, setter) \\
    0.80--0.85 & (beer, burst) & (moon, window) & (shirt, issue) \\
    0.85--0.90 & (ticker, check) & (poser, former) & (ticker, heartbeat) \\
    0.90--0.95 & (badge, piece) & (widget, pillar) & (spender, heading) \\
    \bottomrule
    \end{tabular}
    \caption{WordNet token pair examples for Llama3.2-1b.}
    \label{tab:wordnet_pairs_llama3.2-1b}
\end{table}
\clearpage
\section{Analysis of other layers of the models}
\label{appendix:sec_other_layers}

\subsection{Additional improvements to token pair search algorithm for obtaining earlier layer representations}
\label{appendix:subsec_cgc_forearlylayers}

While the procedure described in Section~\ref{appendix:subsec_cgc} identified suitable pairs across a range of similarities when we looked at hidden representations from the last layer of each LLM, some convergence issues arose when we explored representations in earlier layers. In particular, selecting tokens with the most negative gradients did not consistently decrease the loss over repeated iterations. We reasoned that this may be equivalent to taking step sizes that are too large in the gradient descent. To remedy this, we modified our procedure to add line search backtracking 
to impose a bound on the gradient, only selecting candidate tokens with gradients between $[0, -bound]$~\cite{nocedalLineSearchMethods2006}. If a given iteration does not decrease the loss a sufficient amount (under the Armijo condition, $\alpha=0.3$), the step is rejected. The gradient bound is then brought closer to 0 by a factor of $\beta=0.2$, until it reaches the maximum value of $1e-8$. The best candidate pair $[x_1, y_1]$ based on the smallest loss is kept across iterations.

If a given starting token $x_1$ does not converge after $100$ iterations, we add the best candidate pair to the similarity group that it falls into (if the group is not already full).

\subsection{Representational change using stimuli optimized for earlier layers}
\label{appendix:subsec_repr_dyn_other_layers}

\paragraph{Optimization setup.} We searched for stimulus pairs using representations from earlier layers in $3$ models: llama2-7b, mistral-7b, and gemma2-9b. We chose to evaluate $2$ layers each from early, middle and late positions in the model, for a total of $6$ layers. Early layers were always layers $1$ and $2$. The middle layers began at half the number of total layers (which varied between models), and the one after that. The late layers corresponded to layer indices $-3$ and $-2$, directly preceding the last layer that we analyzed in the main text.

We were able to find the full set of stimulus pairs ($12$ pairs per group) in the similarity interval $[0.1-0.8)$, but were less successful for the high similarity groups. The number of total stimulus pairs per group is given in Table \ref{tab:appendix_early_layer_stim}.

\begin{table}[h!]
\caption{Number of stimuli found in each similarity group for each learning phase and earlier layer, summed across the $3$ models.}
\vspace{1em}
\begin{tabular}{@{}llcccccc@{}}

\textbf{Phase} & \textbf{Layer} & \multicolumn{6}{c}{\textbf{Similarity group}}              \\  \toprule
               &                & 0.1-0.15 & ... & 0.75-0.8 & 0.8-0.85 & 0.85-0.9 & 0.9-0.95 \\  \cmidrule(l){3-8}
Encoding      & early & 528      &     & 528      & 398      & 197      & 95       \\
              & mid   & 540      &     & 540      & 400      & 202      & 74       \\
              & late  & 480      &     & 480      & 370      & 167      & 55       \\ \cmidrule(l){3-8}
Consolidation & early & 1440     &     & 1440     & 1230     & 693      & 395      \\
              & mid   & 1368     &     & 1368     & 1168     & 656      & 390      \\
              & late  & 1428     &     & 1428     & 1198     & 675      & 409      \\ \cmidrule(l){3-8}
Forgetting    & early & 696      &     & 696      & 616      & 292      & 50       \\
              & mid   & 756      &     & 756      & 676      & 324      & 76       \\
              & late  & 756      &     & 756      & 676      & 340      & 76       \\
\bottomrule
\end{tabular}
\label{tab:appendix_early_layer_stim}
\end{table}

\begin{figure}[!h]
    \centering
    \includegraphics[width=\linewidth]{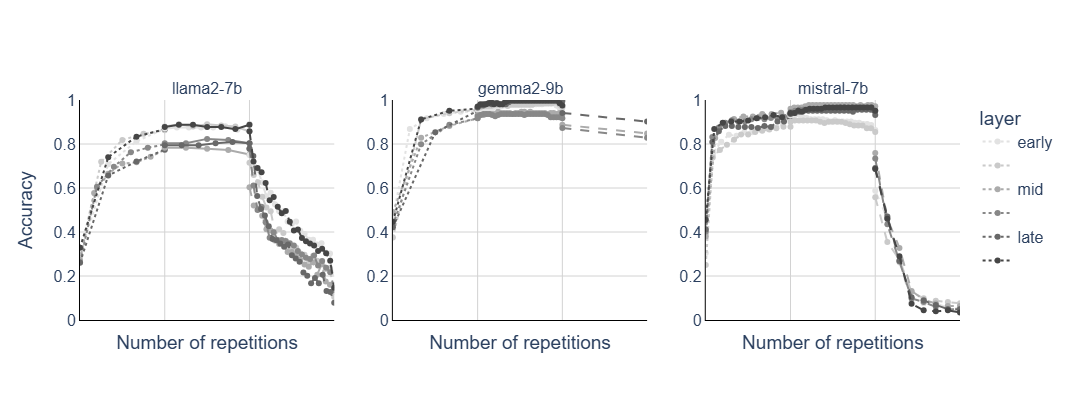}
    \caption{Stimuli optimized for representational similarity in other layers maintains similar accuracy on the associative learning task.}
    \label{fig:appendix_earlierlayers_acc}
\end{figure}

As expected, the accuracy on the task remained about the same when using stimuli optimized for similarity in earlier layers (Figure \ref{fig:appendix_earlierlayers_acc}).

\paragraph{Earlier layer results.} We then extracted hidden representations in two complementary ways.

\begin{figure}[!h]
    \centering
    
    \begin{subfigure}{\textwidth}
        \includegraphics[width=0.8\linewidth]{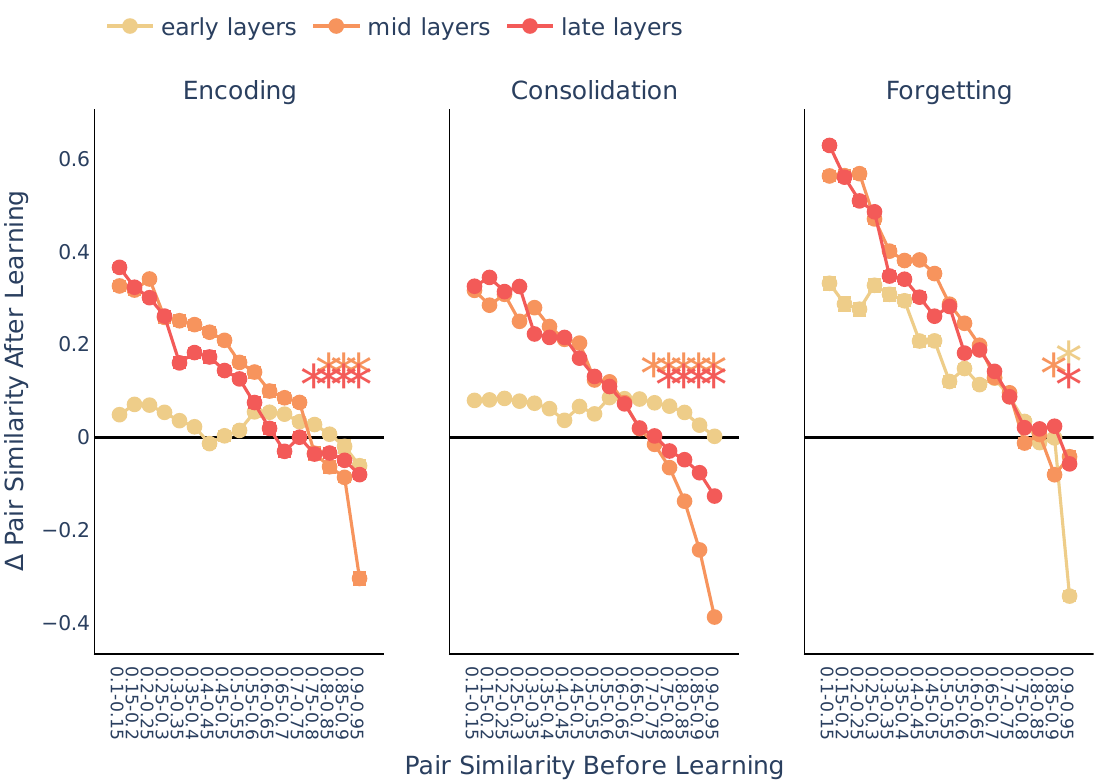}
        \caption{Token pairs optimized for earlier layers.}
        \label{fig:appendix_earlierlayers_repr-at-earlier}
    \end{subfigure}
    
    \begin{subfigure}{\textwidth}
        \includegraphics[width=0.8\linewidth]{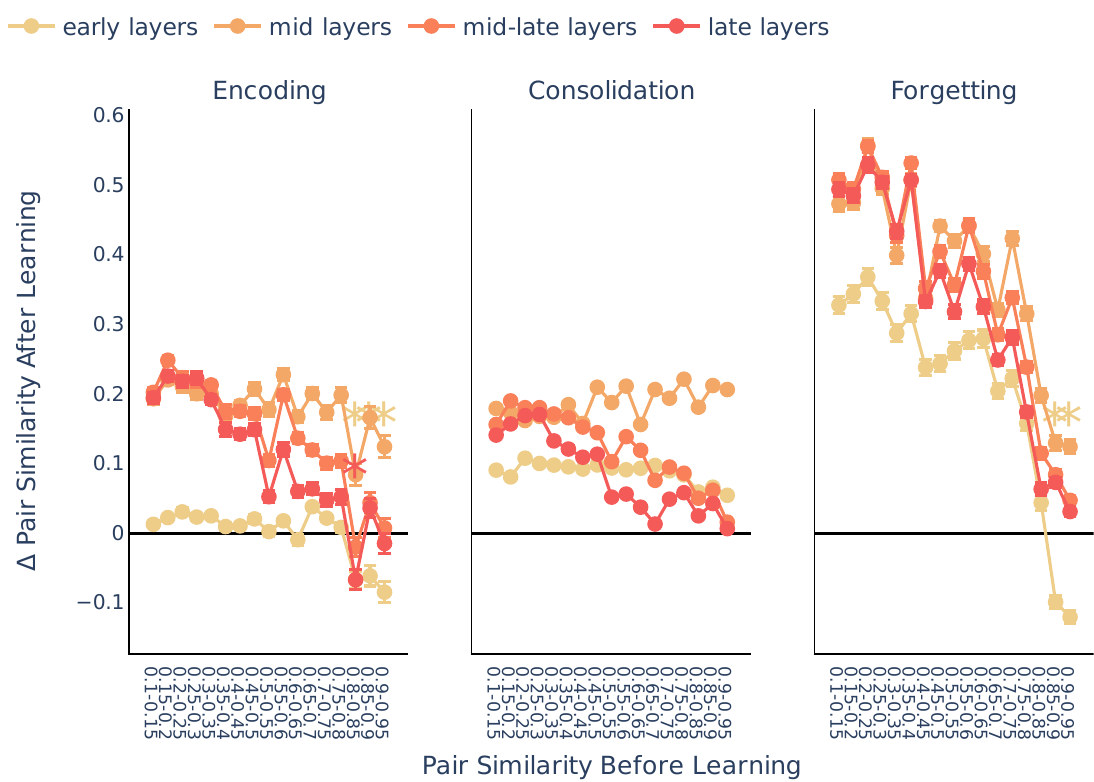}
        \caption{Token pairs optimized for the last layer.}
        \label{fig:appendix_earlierlayers_repr-at-last}
    \end{subfigure}
    
    \caption{
    (a) Representational change across model layers for token pairs optimized at early layers. Early layers (1-2) exhibit irregular and non-systematic changes in similarity, suggesting unstable representations. Intermediate and late layers show a more consistent monotonic decrease—particularly for highly similar pairs ($>$0.7 pair similarity before learning)--with intermediate layers showing stronger differentiation than late layers.
    (b) Representational change across model layers for token pairs optimized at the last hidden layer.
    During the consolidation phase, early to mid layers exhibit relatively flat or mildly monotonic decreases in similarity, reflecting representational integration.
    In contrast, mid-to-late layers begin to show clearer monotonic decreases, and the final layers display the emergence of a U-shaped, non-monotonic pattern.
    Although the minimum of the curve is not statistically significant, these results suggest that representations integrate at earlier stages and progressively develop non-monotonic structure with increasing model depth.
    }
    \label{fig:appendix_earlierlayers_repr}
\end{figure}

First, we analyzed token pairs that were optimized for token pair similarity in earlier layers (Figure~\ref{fig:appendix_earlierlayers_repr-at-earlier}). This allowed us to assess representational changes at intermediate depths of the model, relative to the pair similarity before learning for which the pairs were optimized.
In these analyses, intermediate and late layers exhibited a largely monotonic decrease in similarity, with pronounced differentiation for pairs with high pair similarity before learning ($>$0.7). Differentiation effects were stronger in mid layers than in late layers, whereas the earliest layers ($1$ and $2$) behaved more erratically and did not display a consistent trend.

Second, we evaluated the same set of token pairs that had been optimized for the last layer (as in Figure~\ref{fig:repr}) but measured their representational changes across earlier layers (Figure~\ref{fig:appendix_earlierlayers_repr-at-last}).
This analysis was designed to track how the non-monotonic pattern observed at the output layer emerges progressively across the model hierarchy.
During the consolidation phase, early to mid layers showed relatively flat or mildly monotonic decreasing trends, with similarity values remaining above zero and thus reflecting representational integration.
Mid-late layers began to show a clearer monotonic decrease in similarity.
In the final layers, the emergence of a non-monotonic, U-shaped pattern was visible, although the minimum of the curve did not correspond to statistically significant differentiation.
Taken together, these findings suggest that representations initially integrate across similarity levels and gradually develop the U-shaped structure as they propagate through the model depth.

\begin{figure}[!h]
    \centering
    
    \begin{subfigure}{0.45\textwidth}
        \includegraphics[width=0.8\linewidth]{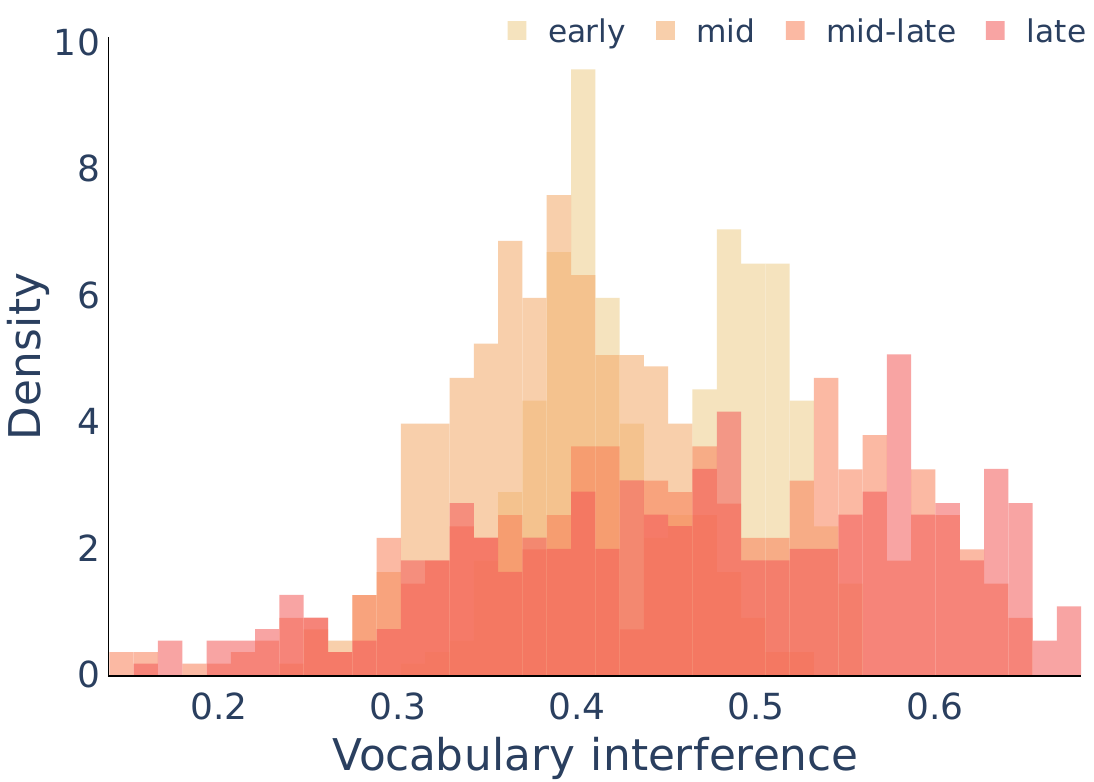}
        \caption{Llama2-7b}
        \label{fig:appendix_intermed_layers_vocab_interf_distr_llama2-7b}
    \end{subfigure}
    \hfill
    \begin{subfigure}{0.45\textwidth}
        \includegraphics[width=0.8\linewidth]{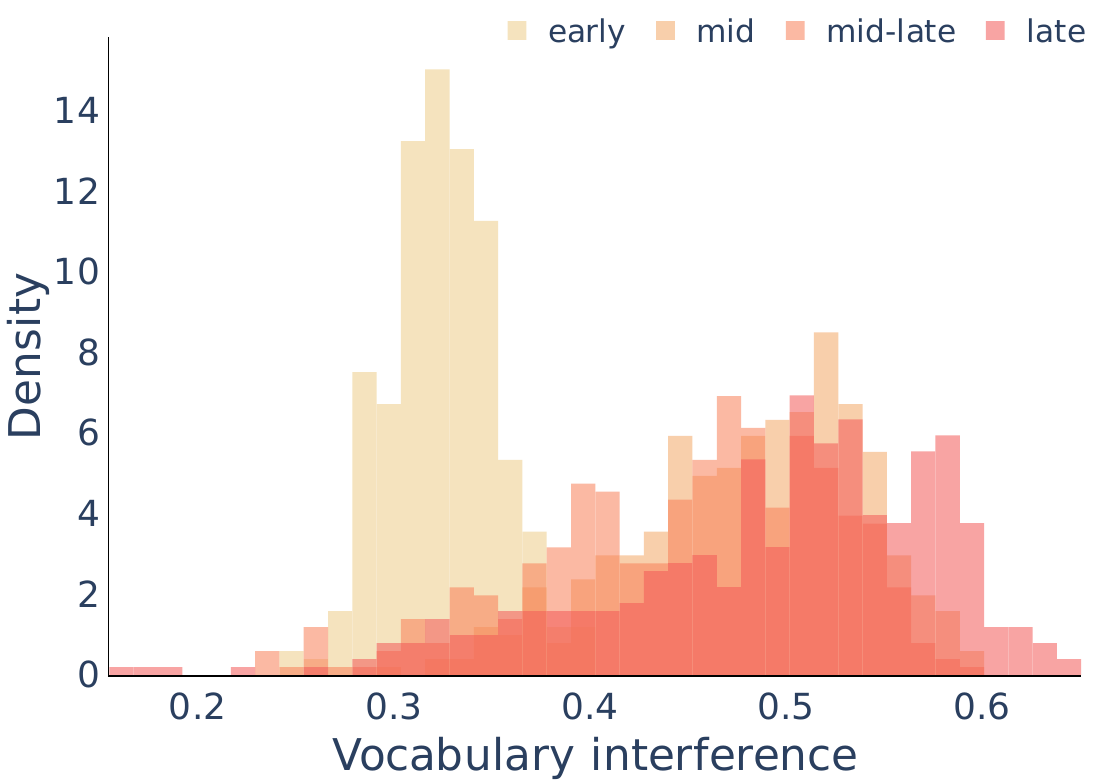}
        \caption{Llama2-7b}
        \label{fig:appendix_intermed_layers_vocab_interf_distr_mistral-7b}
    \end{subfigure}

    \begin{subfigure}{0.45\textwidth}
        \includegraphics[width=0.8\linewidth]{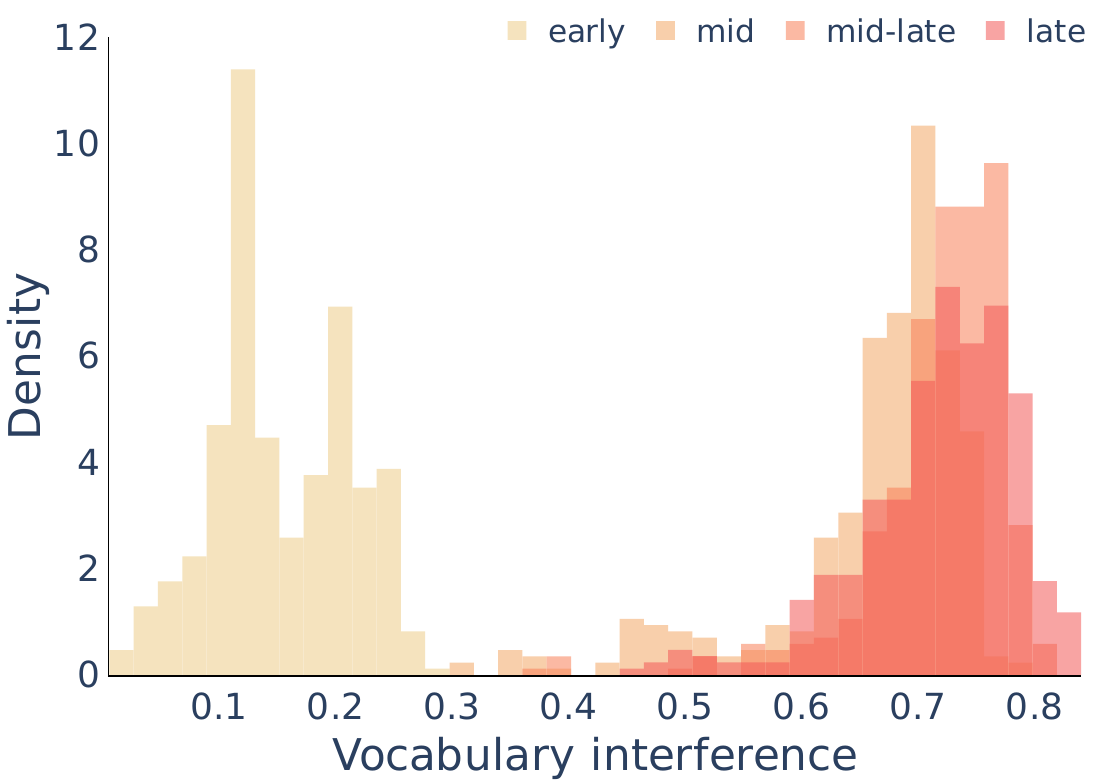}
        \caption{Gemma2-9b}
        \label{fig:appendix_intermed_layers_vocab_interf_distr_gemma2-9b}
    \end{subfigure}
    \hfill
    \begin{subfigure}{0.45\textwidth}
        \includegraphics[width=0.8\linewidth]{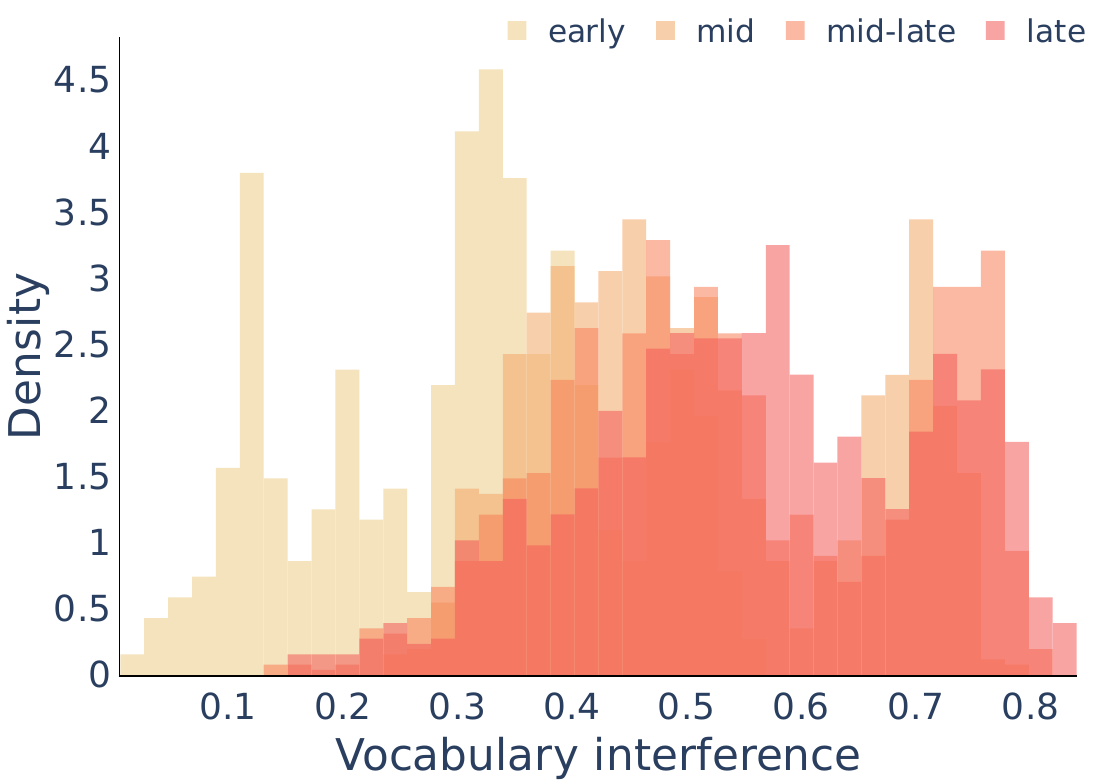}
        \caption{Combined models}
        \label{fig:appendix_intermed_layers_vocab_interf_distr_combined}
    \end{subfigure}
    
    \caption{
    Distribution of vocabulary interference across layers. Global interference increases with layer depth, indicating that deeper layers face stronger competition among possible token predictions. This trend provides a potential mechanism for the stronger differentiation observed in later layers, supporting the interpretation that global interference modulates representational change.
    }
    \label{fig:appendix_intedmed_layers_vocab_interf_distr}
\end{figure}

Finally, we examined the role of vocabulary interference as a potential driver of this effect.
We observed (Figure~\ref{fig:appendix_intedmed_layers_vocab_interf_distr}) a general increase in global interference with layer depth, such that deeper layers face stronger competition among possible token predictions.
This increasing interference provides a plausible mechanism for the stronger differentiation observed in later layers, supporting the interpretation that global interference modulates the representational change pattern.

\end{appendices}

\end{document}